\documentclass[a4paper,fleqn]{cas-sc}

\usepackage[authoryear]{natbib}
\usepackage{subfig}
\usepackage{float}
\usepackage{enumitem}
\usepackage[ruled,vlined]{algorithm2e}
\usepackage{amsmath}
\usepackage{graphicx} 
\usepackage{upgreek}
\usepackage{xcolor}
\usepackage{amssymb}
\usepackage{booktabs}
\usepackage{multirow}

\def\tsc#1{\csdef{#1}{\textsc{\lowercase{#1}}\xspace}}
\tsc{WGM}
\tsc{QE}

\makeatletter
\def\@printORCID{}
\def\@elsauthorsORCID{}
\def\@authorNotes{}

\makeatother

\SetKwInput{KwInput}{Input}
\SetKwInput{KwOutput}{Output}
\SetKwInput{KwInit}{Initialize}

\begin{document}
\let\WriteBookmarks\relax
\def\floatpagepagefraction{1}
\def\textpagefraction{.001}

\shorttitle{}    

\shortauthors{Li et~al.}  

\title [mode = title]{Few-Shot Learning for Dynamic Operations of Automated Electric Taxi Fleets under Evolving Charging Infrastructure: A Meta-Deep Reinforcement Learning Approach}


\author[1]{Xiaozhuang Li}
\author[2]{Xindi Tang}
\author[1]{Fang He}

\affiliation[1]{organization={Department of Industrial Engineering, Tsinghua University},
            city={Beijing},
            postcode={100084},
            country={P. R. China}}

\affiliation[2]{organization={School of Management Science and Engineering, Central University of Finance and Economics},
            city={Beijing},
            postcode={100081}, 
            country={P. R. China}}

\cortext[cor1]{Corresponding author. E-mail: \textcolor{blue}{fanghe@tsinghua.edu.cn}.}



\begin{abstract}
With the rapid expansion of electric vehicles (EVs) and charging infrastructure, the effective management of Autonomous Electric Taxi (AET) fleets faces a critical challenge in environments with dynamic and uncertain charging availability. While most existing research assumes a static charging network, this simplification creates a significant gap between theoretical models and real-world operations. To bridge this gap, we propose GAT-PEARL, a novel meta-reinforcement learning framework that learns an adaptive operational policy. Our approach integrates a graph attention network (GAT) to effectively extract robust spatial representations under infrastructure layouts and model the complex spatiotemporal relationships of the urban environment, and employs probabilistic embeddings for actor-critic reinforcement learning (PEARL) to enable rapid, inference-based adaptation to changes in charging network layouts without retraining. Through extensive simulations on real-world data in Chengdu, China, we demonstrate that GAT-PEARL significantly outperforms conventional reinforcement learning baselines, showing superior generalization to unseen infrastructure layouts and achieving higher overall operational efficiency in dynamic settings.
\end{abstract}



\maketitle

\section{Introduction}\label{sec_intro}


The transition to electric mobility is fundamentally reshaping the global automotive landscape, propelled by a concerted push toward decarbonization and clean transportation initiatives. According to the latest report from the International Energy Agency, global electric passenger vehicle sales exceeded 17 million units in 2024, accounting for over 20\% of all new car sales and representing a more than threefold increase relative to 2018 levels \citep{IEA_ElectricVehicles}. This robust growth trend has continued into 2025, with more than four million electric vehicles sold globally in the first quarter alone, marking a year-on-year increase of approximately 35\%. Policies such as purchase subsidies and planned bans on internal-combustion engine vehicles in leading markets, including China, the United Kingdom, and Norway, have further fueled this exponential expansion \citep{Wang2024}. Taken together, these collective efforts by governments and industries have significantly accelerated the mainstream adoption of electric vehicles. 

In response to this surging demand, the deployment of public charging infrastructure has undergone a qualitative and dynamic transformation. Public chargers have doubled since 2022 to reach more than five million worldwide. In 2024 alone, more than 1.3 million public charging points were added to the global stock, representing an increase of more than 30\% compared to the previous year \citep{IEA_ElectricVehicles}. Across major markets, public charging networks have expanded substantially in both scale and spatial coverage, gradually extending to key transportation hubs, commercial districts, residential areas, and highway corridors, thereby alleviating the range anxiety experienced by electric vehicle users; notably, the share of fast-charging stations has increased significantly, meeting EV users' growing demand for rapid charging.

Building on these developments, Autonomous Electric Taxi (AET) fleets are emerging as a promising frontier for smart cities. This trend is evidenced by recent large-scale commercial trial operations by leading technology providers, such as Waymo, Uber, Apollo Go, and Pony.ai, across major metropolitan areas globally (\citealp{Waymo2024LA,Uber2025,Luobo2025,Xiaoma}). Compared with typical human-driven ride-hailing fleets, centrally platform-managed autonomous fleets offer substantial potential advantages. On the one hand, autonomous vehicles can operate throughout the day without being limited by drivers' availability or personal driving patterns, which can improve vehicle utilization and service safety. On the other hand, centralized control supports detailed fleet scheduling and route planning, allowing operators to adjust decisions in real time as traffic and demand change, thereby reducing empty mileage and idling, improving operational efficiency, and easing congestion.

Despite these potential advantages, the operational efficiency of AET fleets remains fundamentally constrained by charging infrastructure, since charging decisions directly determine vehicle availability and the spatial extent of feasible service. This dependence makes fleet operations inherently coupled across space, time, and energy. Specifically, routing choices govern battery depletion and vehicle reachability, while charging choices shape the timing and location of future vehicle availability. Consequently, dispatching, rebalancing, and charging must be optimized jointly, making AET operations a large-scale dynamic optimization problem involving three interdependent components: (i) demand matching, which assigns requests to vehicles while accounting for service costs and state of charge (SOC); (ii) proactive rebalancing, which relocates idle vehicles toward anticipated demand under range limits and relocation costs; and (iii) strategic charging, which determines when, where, and how long vehicles should recharge given station availability, electricity prices, charging congestion, and demand forecasts.

\begin{figure}[!htbp]
    \centering
    \includegraphics[width=1.0\linewidth]{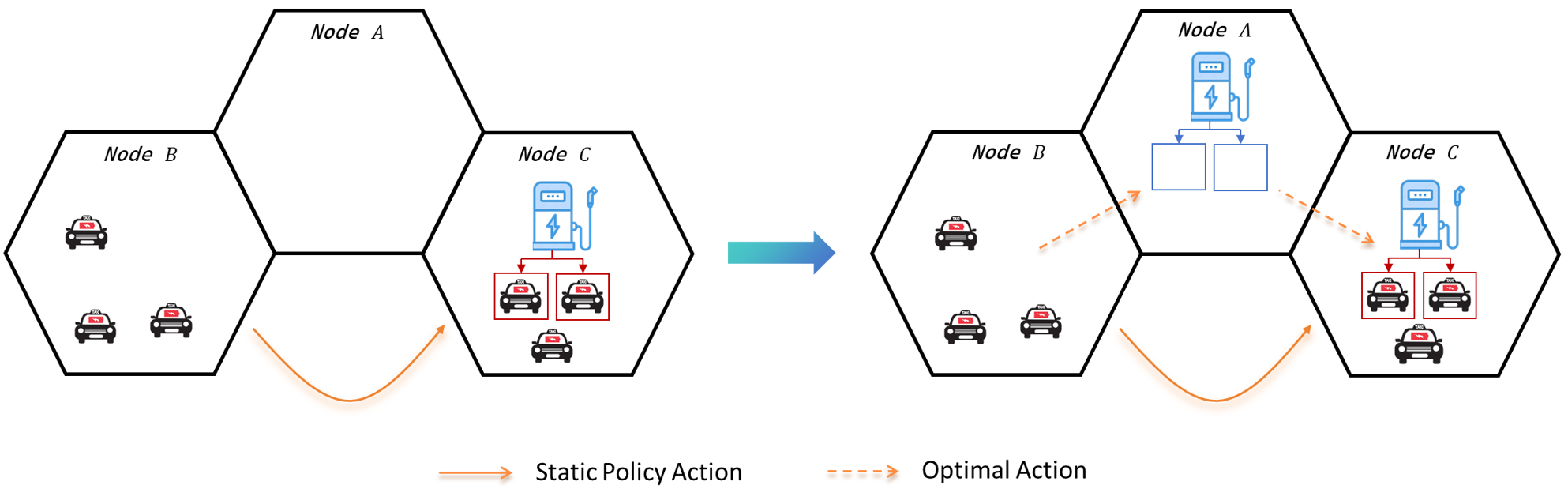}
    \caption{A toy example illustrating the failure of a static policy under evolving infrastructure.}
    \label{fig:toy_example}
\end{figure}

While significant progress has been made in jointly optimizing the aforementioned operational dimensions, most existing studies rely on a critical assumption that the network topology and operational rules of the charging infrastructure are \textbf{entirely static and given} \citep{ALKANJ2020ADP,klein2023electric}. While these static formulations provide a foundational framework for algorithmic development, they encounter significant limitations in maintaining operational effectiveness within the context of continuously expanding real-world infrastructures. Recent evidence indicates that charging infrastructures can undergo substantial structural expansion over short time horizons. For instance, Wuhan, China, deployed more than 70,000 new charging piles in 2023 alone and further expanded this capacity by constructing or renovating approximately 80,000 units in 2024 \citep{WuhanCharging2024}. We demonstrate the limitations of static policies through a toy network with three nodes, as depicted in Figure~\ref{fig:toy_example}. In the initial phase, charging infrastructure exists exclusively at the distant Node C. Consequently, the agent establishes a fixed policy that routes all vehicles that require to be charged from Node B to Node C. Subsequently, the infrastructure evolves with the deployment of a new charging station at the neighboring Node A. However, the static policy remains unresponsive and unsensitive to this structural modification. As indicated by the solid orange arrow, the system continues to dispatch vehicles to the congested Node C. This decision ignores the optimal local resources at Node A, which corresponds to the dashed orange arrow. Such behavior results in severe congestion at the legacy station while new facilities remain underutilized.

This example demonstrates the inherent limitations of fixed operational strategies. Static policies cannot adapt to the structural changes in the charging network that accompany urban development. This mismatch motivates a new formulation where the charging infrastructure is treated as a dynamic operating environment rather than a fixed input. Building on the above observation, we study AET fleet operations under evolving charging infrastructure. We view each charging network configuration as an operational context that may shift after deployments, whereas dispatching, rebalancing, and charging decisions are executed continuously at the operational timescale.

The resulting control problem is challenging for several reasons. First, the fleet operator must make real-time, city-scale decisions under coupled mobility–energy constraints. The high dimensional state space and the large combinatorial action space of a large-scale fleet exceed the computational capacity of standard model based approaches or monolithic reinforcement learning agents. Second, the charging network may evolve in a way that fundamentally changes the operational environment, rather than causing only minor parameter changes. Infrastructure changes affect not only charging decisions but also dispatching and relocation, because shifts in charging coverage can make previously feasible inter-regional movement patterns infeasible. This cascading effect makes policies trained under one layout unreliable under another. As a result, standard reinforcement learning methods often generalize poorly, and training a single policy across multiple layouts tends to produce conservative average behavior with degraded performance. Third, the conventional training approach is inefficient for managing evolving networks. Retraining an end-to-end model every time the charging network evolves is computationally expensive and time-consuming. Such a train-from-scratch paradigm cannot meet the real-time responsiveness required for large-scale AET fleet management, where timely and seamless adaptation to infrastructure shifts is critical.

Accordingly, this paper centers on the following research question: How can we design a robust control strategy that enables an AET fleet to sustain high operational efficiency as the charging network evolves? Motivated by aforementioned challenges, we propose GAT-PEARL, a hierarchical multi-agent meta-reinforcement learning (meta-RL) framework, which further integrates graph attention mechanisms and probabilistic context inference. Specifically, to capture the complex spatial heterogeneity and underlying topological structures of charging station layouts, we employ a graph attention network (GAT). By focusing on relative connectivity and neighborhood dependencies rather than absolute coordinates, GAT effectively extracts robust spatial representations that are invariant to geometric symmetries in infrastructure layouts. Furthermore, to enable rapid adaptation with high environment interaction efficiency, we adopt an inference-based meta-learning approach instead of conventional gradient-based approaches. We employ a probabilistic context encoder that decouples the structural constraints of the infrastructure from general operational decisions. The encoder summarizes recent trajectories into a latent task embedding, allowing the policy to condition its decisions on the inferred charging network configuration. Consequently, adaptation in deployment can be achieved without online parameter updates. This approach effectively eliminates the need for gradient backpropagation, thereby avoiding the optimization bias and instability that typically arise when gradient-based methods attempt to update parameters using only limited interaction data. These components are integrated within a hierarchical structure that decomposes city-scale fleet management across multiple decision levels: a top-layer \textit{Central Agent} operates on aggregated system states and issues high-level coordination signals (e.g., region-level targets and operating constraints); multiple \textit{Area Agents} govern tactical operations within individual regions based on local demand and fleet conditions; and a bottom-layer \textit{heuristic operational module} translates the agents' guidance into implementable vehicle-level movements, including order assignment, relocation, and charging, ensuring scalability and operational stability. 

The main contributions of this paper are summarized as follows:
\begin{itemize}
    \item First, to the best of our knowledge, this study is among the early efforts to explicitly characterize the dynamic operations of AET fleets within an evolving charging environment. Unlike traditional models that treat charging facilities as static and fully known inputs, we treat the infrastructure topology as a dynamic and uncertain core environmental variable. This shift allows for a more realistic representation of the stochastic and transitional nature of urban infrastructure development, effectively bridging a critical gap between existing theoretical models and real-world operational complexity.
    \item Second, we develop GAT-PEARL, a meta-reinforcement learning framework that utilizes inference-based context embedding to extract underlying infrastructure characteristics for few-shot adaptation to evolving network topologies. By integrating a hybrid gradient update scheme with adaptive hyperparameter optimization, the framework achieves stable and rapid convergence under sudden structural shifts. This methodological approach enables the fleet to generalize across diverse layouts, effectively eliminating the need for computationally expensive retraining from scratch.
    \item Third, we propose a scalable hierarchical control architecture for city-scale fleet operations. The system comprises a strategic Central Agent, decentralized Area Agents, and an executable heuristic layer, which together decouple macro-level strategy from local execution. To account for the spatial heterogeneity and underlying topological structures of charging layouts, we use GAT to encode inter-regional dependencies and spatial interactions in the state representation, which supports decision-making under coupled spatiotemporal and energy constraints. This combination of hierarchical decomposition and spatial awareness maintains computational efficiency and operational robustness as the fleet and charging infrastructure scale to metropolitan settings.
    \item Finally, through comprehensive empirical evaluation complemented by managerial interpretation, we demonstrate the practical utility of our framework. Using large-scale simulations calibrated with real-world datasets, we show that GAT-PEARL consistently outperforms strong reinforcement-learning benchmarks, delivering higher system-level performance while achieving faster adaptation and more stable convergence under infrastructure changes. Moreover, these advantages persist across a range of infrastructure and demand conditions, indicating robust generalization as charging networks expand or reconfigure. These results provide practical guidance for fleet operators on sustaining service quality and operational efficiency throughout charging network upgrades.
\end{itemize}

The remainder of this paper is organized as follows. Section~\ref{sec_lite} reviews the related literature. Section~\ref{sec_models} introduces the problem setting and the hierarchical GAT-PEARL framework, including MDP formulation and network architecture. Section~\ref{sec_algorithms} introduces the learning approach for the framework, including the meta-training and meta-validation procedure for Area Agents, parameter update for Central Agent, few-shot learning procedure, and the heuristic operation algorithm. Section~\ref{sec_numerical} describes the experimental setup and reports the main results and robustness analyses using simulations calibrated with real-world data. Section~\ref{sec_con} concludes the paper and discusses future research directions.

\section{Literature Review}\label{sec_lite}


This section reviews the literature relevant to AET operations. We first summarize model-based and approximate dynamic programming (ADP) approaches, and then review data-driven methods, particularly deep reinforcement learning, for city-scale real-time fleet control. We finally provide a brief review of meta-reinforcement learning as a general paradigm for rapid adaptation under distribution shifts, which will serve as methodological background for the learning framework developed in this paper.

\subsection{Model-Based and ADP Approaches for Fleet Operations}

Some operational approaches to AET fleet operations are based on mathematical modeling. These approaches formulate the problem as large-scale, static, or batch-based optimization problems. \citet{yang2020optimizing} modeled the relationship between the matching time interval and system efficiency, showing that batch matching pools requests over a time window before assignment and can significantly outperform instant matching strategies. \citet{YiSmart2021TRD} proposed an integrated decision-making framework for autonomous electric ride-hailing fleets that jointly models vehicle dispatching and charging decisions under operational and energy constraints, and evaluate performance impacts at system levels using scenario-based analysis. \citet{LaiLi2024TRB} formulated a joint optimization problem for planning charging and battery swapping networks for electrified ride-hailing fleets, capturing the platform's planning and operational decisions and analyzing how infrastructure configuration affects fleet operations and system efficiency. As these formulations scale, computational tractability becomes a primary bottleneck, and decomposition-based algorithms such as Benders decomposition and branch-and-price have been widely used to improve the tractability of large-scale mixed-integer formulations in fleet operations (\citealp{ALVO2021Benders,su2024branch,zigrand2024scalable}). Meanwhile, queuing theory complements these optimization frameworks by providing a robust analytical lens for modeling the stochastic arrival of passengers and vehicles. For example, \citet{Cheng2021} developed a queuing-based framework that forecasts demand and estimates driver idle times via a double-sided queuing model, and then performs batch dispatch by matching accumulated requests with available drivers periodically. \citet{GaoLi2024TRC} used a queuing-theoretic model to characterize the vehicle waiting action at charging stations and battery swapping stations, which is integrated into an economic analysis of electric autonomous mobility-on-demand (AMoD) systems.

Approximate Dynamic Programming (ADP) has emerged as a powerful methodology for solving large-scale, stochastic, and sequential decision problems inherent in fleet operations, bridging the gap between static optimization and fully model-free learning. In a pair of seminal studies, \citet{Godfrey2002a,Godfrey2002b} examined a stochastic dynamic resource allocation problem and proposed an adaptive dynamic programming approach that employs nonlinear value function approximations to estimate the future value of resources. \citet{ALKANJ2020ADP} integrated dispatching decisions, including order assignment, vehicle repositioning, and parking, with charging decisions for an autonomous electric ride‑hailing fleet into a dynamic decision model and uses ADP to develop high-quality operational dispatch strategies. \citet{MAHYARI2025ADP} introduced an ADP policy for managing the charging of EV fleets at a charging depot equipped with diverse multi-connector chargers. \citet{Hu2025ADP} considered stochastic customer demand, AET energy consumption, and charging station resources, and developed an ADP model to optimize the joint operations of AETs and mobile charging vehicles. 

\subsection{Deep Reinforcement Learning for Fleet Operations}

Some researchers employ data-driven approaches, particularly deep reinforcement learning (DRL), as it provides a robust framework for handling the dynamics and randomness encountered in operating large-scale AET fleets. The operation of a fleet is often modeled as a Markov Decision Process (MDP), where states typically include vehicle location, battery charge level, passengers' demand, and other information. Actions involve vehicle dispatching, charging decisions, and similar operations. The reward function comprehensively evaluates metrics such as operational efficiency and passenger satisfaction. Unlike static optimization, which generates operation plans for a fixed set of inputs, DRL learns an adaptive policy that maps real-time system states to optimal actions. There has been extensive research exploring this direction. \citet{Shi2020RL} abstracted the operation of an EV ride‑hailing fleet under a single policy controller, using reinforcement learning to handle dispatching and charging decisions with the objectives of lowering passenger waiting time and operational and electricity costs. \citet{TANG2020RL} proposed an advisor-student reinforcement learning framework to solve the online operations problem of AET fleet, which combines a centralized reinforcement-learning controller with decentralized optimization-based execution units, enabling taxis to intelligently perform demand matching, relocation, and charging actions. \citet{LIU2022RL} proposed a single-agent DRL approach for the vehicle dispatching problem called deep dispatching, by reallocating vacant vehicles to regions with a large demand gap in advance. \citet{yan2023online} developed an online reinforcement learning approach that simultaneously optimizes order dispatching and charging decisions for an e-hailing EV fleet, demonstrating improved operational efficiency over myopic policies.

While single-agent frameworks provide a powerful centralized perspective, their scalability is often challenged by the combinatorial explosion of the state and action spaces as the fleet size grows. This has motivated a shift towards Multi-Agent Reinforcement Learning (MARL). For example, \citet{Li2019MARL} treated each driver as an individual agent and employs mean‑field MARL to address order dispatching in real time. \citet{zhu2025learning} proposed a learning-informed optimization framework for shared autonomous electric vehicle systems to jointly optimize fleet operations and charging under uncertainty. They used MARL to learn a grid-level balancing-charging strategy that guides an optimization-based network flow model for real-time assignment. \citet{wang2025merci} proposed a MARL framework that enhances on-demand electric taxi operations by coordinating rebalancing, charging, and order handling among agents. \citet{zhao2025MARL} proposed two value-function-free MARL approaches for large-scale ride-sharing platforms, which eliminated critic estimation by leveraging group-level rewards to mitigate training instability and estimation bias. 

\subsection{Meta-Reinforcement Learning: Foundations and Optimization Applications}

While DRL has achieved superhuman performance in complex, static environments like Atari games and Go, these agents often suffer from poor data efficiency and limited generalization when faced with new, unseen tasks (\citealp{mnih2015meta}). This limitation motivated the development of Meta-Reinforcement Learning (Meta-RL), which aims to create agents that can ``learn to learn''. As detailed in the comprehensive tutorial by \citet{Beck2025}, the core idea of Meta-RL is to train an agent not on a single task, but on a distribution of related tasks. By doing so, the agent learns a high-level strategy or an efficient learning algorithm that allows it to adapt rapidly to a novel task with very few interactions. A landmark algorithm in this domain is Model-Agnostic Meta-Learning (MAML) (\citealp{finn2017model}). MAML learns an initial set of model parameters that is not optimal for any single task but is for fast fine-tuning, enabling good performance on a new task after only one or a few gradient updates. The effectiveness of MAML has been demonstrated across diverse domains, including continuous-control robotics tasks and competitive multi-agent game environments (\citealp{finn2017model,al2017continuous}). Beyond gradient-based adaptation, probabilistic latent-variable approaches provide an alternative mechanism for fast task inference. Probabilistic Embeddings for Actor-Critic Reinforcement Learning (PEARL) (\citealp{rakelly2019efficient}) meta-trains an off-policy agent to infer a latent task variable from a context buffer and condition the policy on this inferred embedding, thereby disentangling task inference from control and enabling sample-efficient adaptation to new tasks. The effectiveness of PEARL has also been demonstrated in robotic manipulation benchmarks and real world industrial insertion tasks (\citep{yu2020meta,schoettler2020meta}).

The powerful adaptability of Meta-RL has shown significant potential for complex sequential decision-making and optimization. In traffic signal control, \citet{Zang2020meta} proposed MetaLight, which uses MAML to learn a generalized initialization for rapid adaptation to new traffic scenarios. \citet{huang2021modellight_icmlws} further integrated an ensemble of traffic models in ModelLight to improve adaptation efficiency under novel traffic patterns or intersection configurations. \citet{Kim2023meta} enabled the control objective to adjust to changing environments by switching the reward function based on real-time saturation levels. In urban mobility system control, \citet{gammelli2022meta} formulated the AMoD control as a cross-city transfer problem and combined graph neural networks with meta-RL to enable rapid adaptation in new cities with only a small amount of additional interaction. For more general combinatorial optimization problems, related studies further pursue transferable solution methods that generalize across problem scales and data distributions. For instance, \citet{zhou2023towards} developed a more robust omni-generalizable neural solver for vehicle routing problems, and \citet{chen2023efficient} proposed an efficient meta neural heuristic for multi-objective combinatorial optimization. In multi-agent settings, \citet{Huang2025meta} further demonstrated the potential of using a policy set and a selection mechanism to improve coordination in previously unseen scenarios through Multi-Personality Multi-Agent Meta-Reinforcement Learning (MPMA-MRL).

\subsection{Research Gap}

Most existing studies assume that the charging infrastructure topology and operational rules are fully static and specified in advance, which conflicts with the dynamic evolution of urban charging networks driven by continuous expansion and maintenance activities. This type of infrastructure-induced structural shift differs fundamentally from demand randomness, as it reshapes the underlying constraint structure and feasible region, causing policies optimized or trained for a fixed layout to suffer substantial performance decline, or even become infeasible, under previously unseen network configurations.

Mainstream deep reinforcement learning methods often exhibit low sample efficiency and weak generalization in such out-of-distribution settings. When the charging network changes, retraining that relies on extensive interactions is typically computationally expensive and time-consuming, and therefore cannot meet the real-time responsiveness and operational stability required for city-scale fleet management. Although meta-reinforcement learning has been explored to a limited extent in transportation applications such as traffic signal control and adaptation to demand fluctuations, it has not been used to characterize fleet operational decision changes induced by evolving charging infrastructure. Without explicit structural representation and constraint-aware adaptation, generic meta-RL may yield invalid actions, unstable adaptation, and excessive data requirements, struggling to extract transferable knowledge across topologically distinct infrastructure layouts. Consequently, a key open gap remains: how to develop a meta-adaptive operational framework that can learn from only a small amount of new experience and rapidly adapt to previously unseen charging layouts without incurring the high cost of full retraining.

\section{Models}\label{sec_models}


This section presents the modeling framework for our proposed hierarchical multi-agent meta-RL approach. We first formalize the joint passenger matching, fleet dispatching and charging problem by characterizing the interactions among fleet supply, passenger demand, and the urban environment under coupled mobility and energy constraints. Since evolving charging infrastructure alters system dynamics and network topology, we model the environment as a contextual Markov decision process (CMDP) where each infrastructure layout defines an operational context that influences the state transitions. Building on this foundation, we introduce a hierarchical control architecture consisting of a top-layer Central Agent and region-level Area Agents. For each layer, we define the state space, action space, state transition, reward design and terminal signal, and then present the corresponding neural network architectures for forward propagation and action generation. Finally, we explain the execution pipeline of the proposed hierarchical GAT-PEARL framework.

\subsection{Problem Statement}\label{subsec_problem_statement}
This study investigates the joint vehicle dispatching, repositioning, and charging problem for an AET fleet. The central challenge is to design an operational policy that remains effective under dynamically evolving charging infrastructure, where station availability and capacity vary over time. To formalize this challenge, our model is built upon three primary components: the physical layout of the charging infrastructure, the operational characteristics of the vehicle supply, and the stochastic nature of passenger demand.

\begin{figure}[!htbp]
\centering
\includegraphics[width=0.25\linewidth]{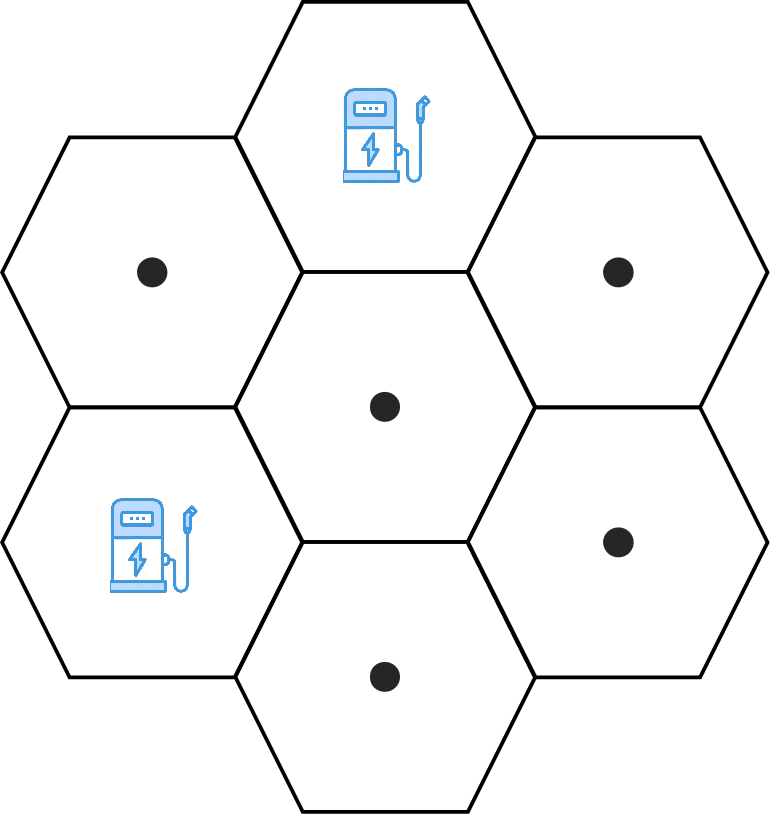}
\caption{Hexagonal regions map.}
\label{fig:map}
\end{figure}

The operational area is discretized into a graph $\mathcal{G} = (\mathcal{J}, \mathcal{E})$, where $\mathcal{J}$ denotes a set of regular hexagonal regions that partition the city (Figure~\ref{fig:map}), and $\mathcal{E}$ represents adjacency relationships between neighboring regions. Each region is represented by its center, and without loss of generality, we assume that all travel activities originate from and terminate at the centers of these hexagonal regions. Owing to the regular hexagonal grid structure, we assume homogeneous travel characteristics between adjacent regions. Specifically, the travel time $\tau_{j,\bar{j}}$ and energy consumption $\varepsilon_{j,\bar{j}}$ are assumed to be constant for any pair of neighboring regions $(j, \bar{j}) \in \mathcal{E}$.

From the supply perspective, we consider a homogeneous AET fleet, denoted by $\mathcal{K}$, operating over graph $\mathcal{G}$. All vehicles share identical passenger capacity and cruising speed. Each vehicle $k \in \mathcal{K}$ has an individual service start time. Once it comes online, it remains available for service for the rest of the simulation horizon. We assume that each order serves either a single passenger or a passenger group, and ride-sharing is not considered. The platform assigns each order to a single vehicle, which serves one origin destination trip at a time. At the time of assignment, an order can only be matched with an idle vehicle located in the same hexagonal region $j \in \mathcal{J}$. Consistent with real-world charging behavior, the charging process of each vehicle $k$ follows a state-dependent charging function to capture the nonlinear characteristics of lithium-ion batteries, where charging rates vary with the vehicle's current battrey SOC $\xi_k$.

From the demand perspective, we consider passenger requests $\mathcal{R}$ that arrive on demand and must be served within a short time after order placement. Each request $r \in \mathcal{R}$ is characterized by a tuple of necessary information $\boldsymbol{\chi}_r = (j_r^{\mathrm{o}}, j_r^{\mathrm{d}}, [b_r, a_r], t_r^{\max}, \varrho_r, \kappa_r)$, which specifies the origin and destination of the order, the pickup time window, the maximum additional waiting time tolerated after the pickup window, the revenue obtained by serving request, and the penalty incurred when a request is abandoned. Passengers are assumed to remain at their specified origin throughout the pickup window and do not leave the system unless they are served. Once an order is matched to a vehicle, neither the passenger nor the vehicle can cancel the trip, and a profit $\varrho_r$ is accrued upon successful service completion. No waiting cost is incurred if a request is matched within its pickup time window $[b_r, a_r]$. If the request remains unmatched after $a_r$, a waiting cost is incurred during the interval $[a_r, a_r + t_r^{\max}]$ to capture increasing passenger dissatisfaction. If the request is still not matched by $a_r + t_r^{\max}$, it is considered abandoned, and an abandonment penalty $\kappa_r$ is incurred to reflect the associated loss in service quality and revenue.

Within this spatial framework, the primary source of nonstationarity arises from the charging infrastructure, whose spatial layouts vary across different urban development scenarios. We represent each charging infrastructure layout as a task and denoted it by $ C \in \mathcal{C}$, where $\mathcal{C}$ is the set of admissible layouts. A layout $C$ is represented as a mapping $C: \mathcal{J} \to \{0,1\}$, and $\mathcal{C}(j)=1$ indicates that the  charging service is available in region $j$. Across episodes, the operating environment is assumed to experience varying layouts drawn from a task distribution $p(C)$, which captures the infrastructure changes encountered in practice.

\begin{figure}[!htbp]
\centering
\includegraphics[width=0.8\linewidth]{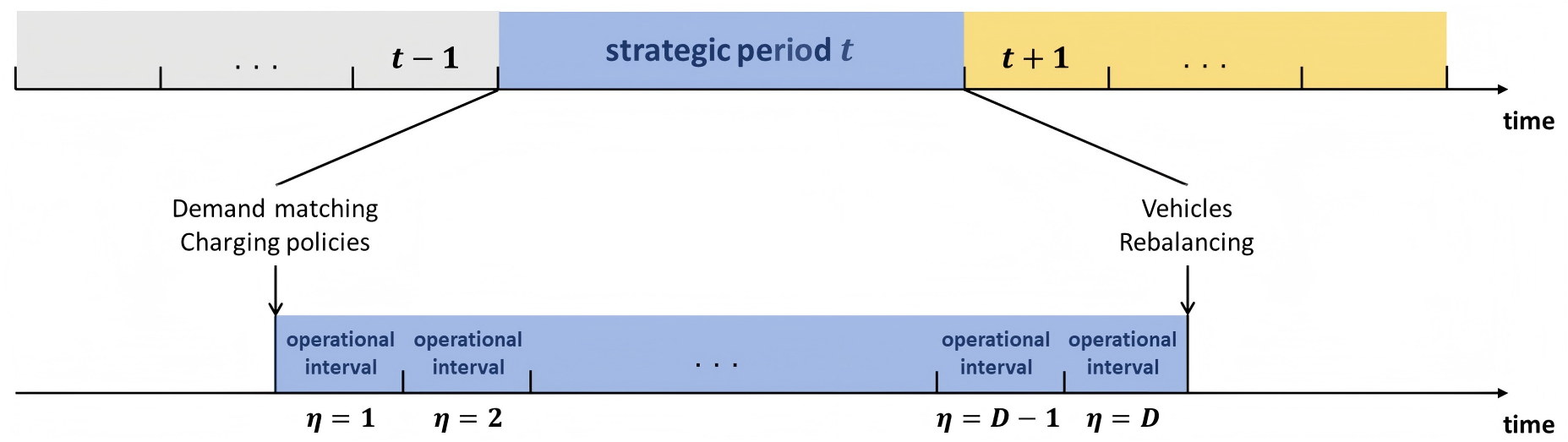}
\caption{Temporal hierarchy of strategic periods and operational intervals.}
\label{fig:PeriodsvsIntervals}
\end{figure}

To manage the computational complexity of large-scale fleet operations, we adopt a two-tier temporal structure that separates strategic decision-making from real-time execution, as illustrated in Figure~\ref{fig:PeriodsvsIntervals}. The planning horizon is divided into $T$ strategic \textit{periods}, indexed by $t \in \{1, \ldots, T\}$. Each strategic period $t$ is further subdivided into $D$ operational \textit{intervals}, indexed by $\eta \in \{1, \ldots, D\}$ within the period. Passenger requests arrive continuously and require timely processing within operational intervals, while strategic decisions and policy updates are made only at the period level and need not be updated at a high frequency. This temporal structure improves computational tractability and decision stability by decoupling high-level planning from fine grained execution.

\begin{figure}[!htbp]
\centering
\includegraphics[width=0.3\linewidth]{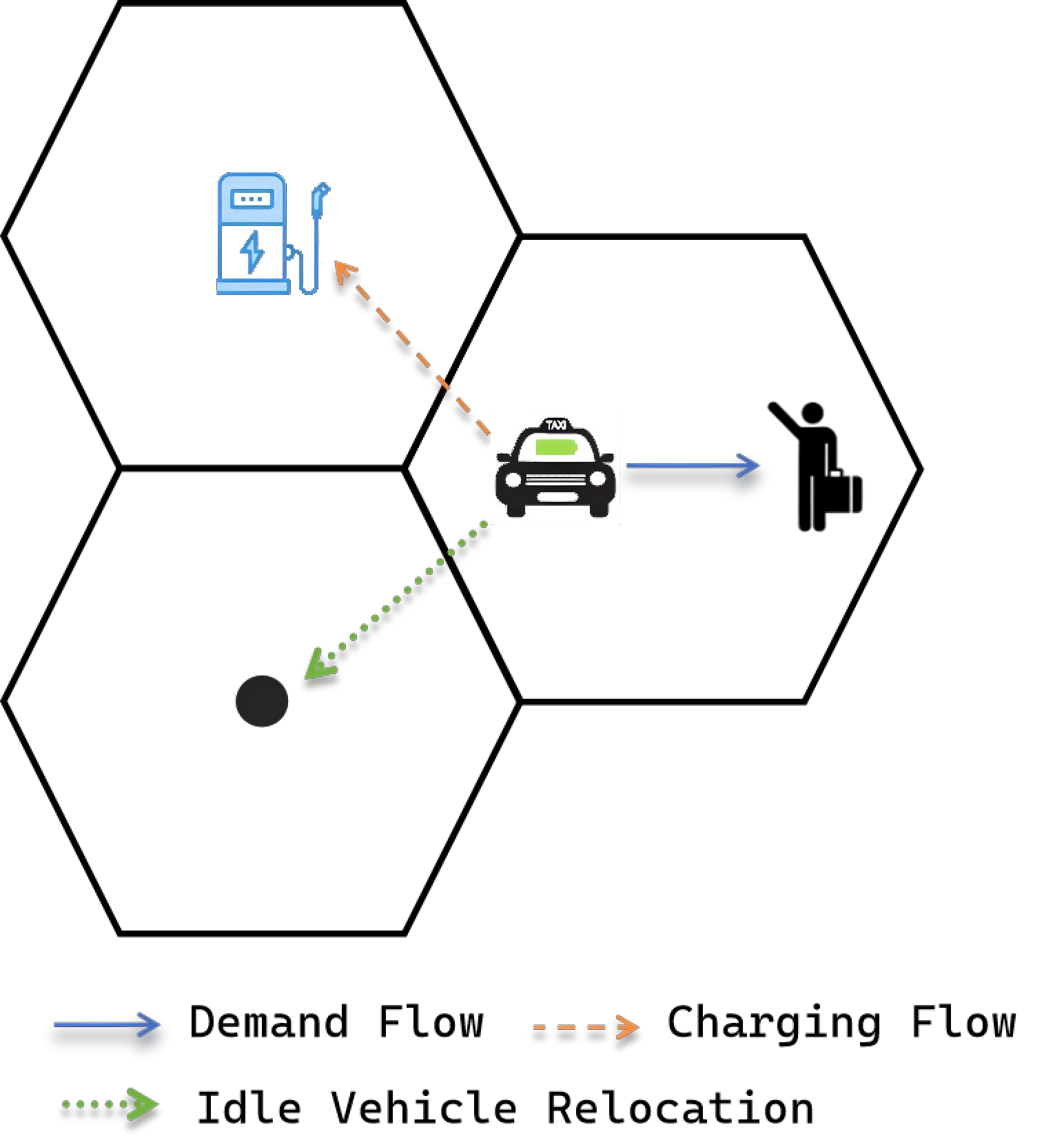}
\caption{Categorization of AET movements.}
\label{fig:actions}
\end{figure}

Building on this temporal structure, we introduce a hierarchical control architecture to address imbalances between supply and demand. At the beginning of each strategic period $t$, the Central Agent and the decentralized Area Agents generate high-level tactical policies that govern three core fleet operations: passenger demand matching, vehicle repositioning, and charging, as shown in Figure~\ref{fig:actions}. All operations are assumed to occur at the centers of the corresponding hexagonal regions. Within each operational interval $\eta$ of period $t$, a heuristic dispatching algorithm executes concrete vehicle assignments in real time, guided by the tactical policies produced at the period level.

\subsection{System Dynamics and Hierarchical Formulation}\label{subsec_Hierarchical_Control_Architecture}

This subsection establishes the mathematical framework of the operation problem. We first present the system dynamics and mathematical formulation. We then formalize the operational environment, where the infrastructure layout serves as a latent context variable governing the system transition dynamics. Building on this contextual formulation, we present a hierarchical control architecture that decomposes the city scale management problem into two interacting decision levels.

\subsubsection{Problem Formulation}

We formulate the AET fleet operation problem as a stochastic dynamic resource allocation problem (SDRAP) over a finite planning horizon indexed by $t \in \{1,\ldots,T\}$. At each period $t$, the system is characterized by a global state $\mathbf{S}_t$, including the spatial distribution of idle vehicles across regions, the realized passenger demand, and the aggregate energy status of the fleet. The platform selects a composite action $\mathbf{A}_t = \pi(\mathbf{S}_t)$ that allocates available vehicles to matching, repositioning, and charging decisions, where $\pi \in \Pi$, and $\Pi$ denotes the set of admissible policies mapping system states to actions.

The platform seeks to maximize the cumulative expected profit, defined as total revenue from serving passenger requests minus operational costs:
\begin{equation}
    \max_{\pi \in \Pi} \; \mathcal{V}(\pi)
    = \mathbb{E}_{\{\Theta_t\}_{t=1}^{T}} \left[
        \sum_{t=1}^{T}
        \left(
            \sum_{r \in \mathcal{R}_t^{+}} \varrho_r
            - g(\mathbf{S}_t, \mathbf{A}_t)
        \right)
    \right],
\end{equation}
where $\mathcal{R}_t^{+} \subseteq \mathcal{R}$ denotes the set of served requests at period $t$, $\varrho_r$ is the revenue obtained by serving request $r$, and $g(\mathbf{S}_t, \mathbf{A}_t)$ represents the operational cost incurred under system state $\mathbf{S}_t$ and action $\mathbf{A}_t$. The expectation is taken with respect to the exogenous stochastic process $\Theta_t$, defined as $\Theta_t = (\mathbf{O}_t, \mathbf{c})$. Specifically, $\mathbf{O}_t$ represents the time-varying demand process (e.g., random arrival of requests). At the start of each episode, a charging station layout $C$ is sampled from the task distribution $p(C)$ and remains fixed throughout the episode. This sampled layout induces a time-invariant infrastructure context $\mathbf{c}=\Gamma(C)$.

The composite action $\mathbf{A}_t$ is formalized as a collection of decision flow variables, comprising $n_{j,\bar{j},t}^{m}$ for passenger matching, $n_{j,\bar{j},t}^{r}$ for fleet repositioning, and $n_{j,\bar{j},t}^{c}$ for charging dispatch. These variables determine the allocation of limited vehicle resources and must strictly adhere to physical and operational limitations. Consequently, the feasible action space $\Omega_t(\mathbf{c})$ is characterized by the intersection of three primary classes of constraints.

First, the flow conservation constraint ensures that the aggregate outflow from any region $j$ does not exceed the available available idle supply in that region. This guarantees that the dispatch decisions are physically executable given the current state:
\begin{equation}
    \sum_{\bar{j} \in \mathcal{J}} \left( 
        n_{j,\bar{j},t}^{m} 
        + n_{j,\bar{j},t}^{r} 
        + n_{j,\bar{j},t}^{c} 
    \right)
    \le |\mathcal{K}_t^{j,0}|, 
    \quad \forall j \in \mathcal{J}.
\end{equation}
where $\mathcal{K}_t^{j,0}$ denotes the set of idle vehicles located in region $j$ at the beginning of period $t$.

Second, the demand feasibility constraint restricts the volume of matched vehicles based on the realized passenger demand. Specifically, the total number of vehicles dispatched to serve requests in region $j$ is upper bounded by the request count observed originating from that region:
\begin{equation}
\sum_{i \in \mathcal{J}} n_{i,j,t}^{m}
\le |\mathcal{R}_t^{j}|,
\quad \forall j \in \mathcal{J},
\end{equation}
where $|\mathcal{R}_t^{j}|$ denotes the number of requests with origin region $j$ at period $t$. This quantity is consistent with the aggregated OD demand matrix $\mathbf{Q}_t$, and can be recovered as the row-sum $|\mathcal{R}_t^{j}|=\sum_{k \in \mathcal{J}} Q_t(j,k)$.

Third, the charging capacity constraint enforces the physical limits of the energy infrastructure. The total number of vehicles dispatched to charge in region $\bar{j}$ is bounded by the effective station capacity. Crucially, unlike the previous constraints which depend on dynamic state variables, this bound is strictly determined by the infrastructure capacity:
\begin{equation}
\sum_{j \in \mathcal{J}} n_{j,\bar{j},t}^{c}
\le \bar{C}_{\bar{j}}(C),
\quad \forall \bar{j} \in \mathcal{J}.
\end{equation}
Here, $\bar{C}_{\bar{j}}(C)$ denotes the charging capacity in region $\bar{j}$ under layout $C$. This constraint explicitly couples the feasible action space with the latent infrastructure topology, ensuring that no charging flows are directed to regions where stations are absent or disabled as specified by $C$.

The formulation above characterizes the operational challenge as a constrained optimization process driven by distinct uncertainty components. A critical observation from the constraint analysis is that the infrastructure context $\mathbf{c}$ plays a fundamental role in defining the physical feasibility of charging actions. Unlike the demand process $\mathbf{O}_t$ which induces parametric fluctuations in state transitions, the infrastructure layout imposes topological constraints that explicitly shape the feasible action space $\Omega_t(\mathbf{c})$. This unique characteristic, where the system topology remains constant within an \textit{episode} but varies across tasks, implies that a standard Markov decision process is insufficient to capture the environmental heterogeneity. Here, an episode is defined as a complete operational sequence spanning from the initial time step to the terminal time step of the MDP simulation. To rigorously address this context dependent variation in the operational environment, we formulate the problem as a CMDP in Section \ref{sec:CMDP}.

\subsubsection{Infrastructure-dependent Dynamics and Contextual MDP}
\label{sec:CMDP}

Given the feasible action space defined by the operational constraints, the system state evolves dynamically over time as vehicles are dispatched, occupied, and subsequently released back into service. The state evolution is driven jointly by the selected actions and the environmental stochasticity. Among all state components, the spatial distribution of idle vehicles constitutes the core endogenous state variable that directly couples current decisions with future feasibility. Accordingly, we explicitly characterize the evolution of the idle vehicle distribution. Let $\mathbf{x}_t \in \mathbb{N}^{|\mathcal{J}|}$ denote the vector of idle vehicle counts across regions at the beginning of period $t$. At the next period, the number of idle vehicles is determined by the net outflow of vehicles dispatched at the current period and the inflow of vehicles that complete assigned engagements:
\begin{equation}
    \mathbf{x}_{t+1}
    = \mathbf{x}_t
    - \mathbf{o}_t
    + \sum_{\delta \in \mathbb{Z}_+} \boldsymbol{\Phi}_{\delta}(\mathbf{A}_{t-\delta}),
\end{equation}
where $\mathbf{o}_t \in \mathbb{N}^{|\mathcal{J}|}$ denotes the vector of total vehicle outflows from each region at period $t$, aggregated from matching, repositioning, and charging decisions, and $\boldsymbol{\Phi}_{\delta}(\mathbf{A}_{t-\delta})$ represents the vector of vehicles that reenter the idle fleet at period $t+1$ after completing vehicle engagements with completion delay $\delta$. Here, $\delta$ denotes the engagement completion delay measured in operational intervals, reflecting the time required for passenger service, repositioning, and charging.

While the equation above describes the system dynamics, the transition probability distribution $P(\mathbf{S}_{t+1} | \mathbf{S}_t, \mathbf{A}_t)$ is heavily influenced by the nature of uncertainty. In our problem, the two components of uncertainty, $\mathbf{O}_t$ and $\mathbf{c}$, exert fundamentally different influences on the system dynamics. We distinguish these as quantitative parametric shifts and qualitative topological shifts:

\noindent (i) Demand stochasticity $\mathbf{O}_t$: Variations in the demand process $\mathbf{O}_t$ induce smooth perturbations in the transition probabilities. For instance, if the demand density in a region increases, the probability of a vehicle being matched may shift numerically (e.g., from 0.3 to 0.5). Standard reinforcement learning algorithms can effectively adapt to such continuous numerical fluctuations through gradient based updates.

\noindent (ii) Infrastructure context $\mathbf{c}$: In contrast, changes in the infrastructure context $\mathbf{c}$ induce topological shifts in the system dynamics. Consider a charging action directed at a specific region. If the charging station exists in context $\mathbf{c}$, the system transitions to a charged state with high probability. However, if the station is removed in a different context $\mathbf{c}'$, the probability of successfully transitioning to a charged state abruptly drops to zero, resulting in disjoint supports for the transition probability distributions. This implies that a policy optimal for one layout may be physically infeasible for another.

To mathematically formalize this structural heterogeneity, we model the problem as a CMDP. At the start of each episode, a charging station layout $C \in \mathcal{C}$ is sampled to induce an infrastructure context $\mathbf{c}$. The system dynamics are thus context dependent, denoted as $P(\mathbf{S}_{t+1} | \mathbf{S}_t, \mathbf{A}_t, \mathbf{c})$. Unlike standard MDPs where dynamics are fixed, the CMDP framework explicitly acknowledges that the physical laws governing state transitions (specifically, the feasibility of charging and energy recovery) are modulated by the latent context. This formulation necessitates a meta-learning approach capable of identifying the underlying context to adapt the control policy to the specific topological constraints of the current environment.

The CMDP formulation established above provides a mathematical framework for capturing the interplay between fleet dynamics and evolving infrastructure. By explicitly treating the infrastructure layout as a latent context variable $\mathbf{c}$, we transform the problem from learning a single policy for a static environment into learning a generalizable meta-policy capable of adapting to diverse topological structures. However, directly solving this high dimensional CMDP presents significant computational challenges. The combinatorial nature of the joint action space, coupled with the need for simultaneous spatial coordination and context inference, renders monolithic control approaches intractable. To overcome these complexity barriers and achieve scalable execution, we propose to decompose the global control problem into a hierarchical architecture, as detailed in Section~\ref{subsec_Hierarchical_MDP_Reformulation}.

\subsubsection{Hierarchical MDP Reformulation}\label{subsec_Hierarchical_MDP_Reformulation}

\begin{figure}[!htbp]
\centering
\includegraphics[width=1\linewidth]{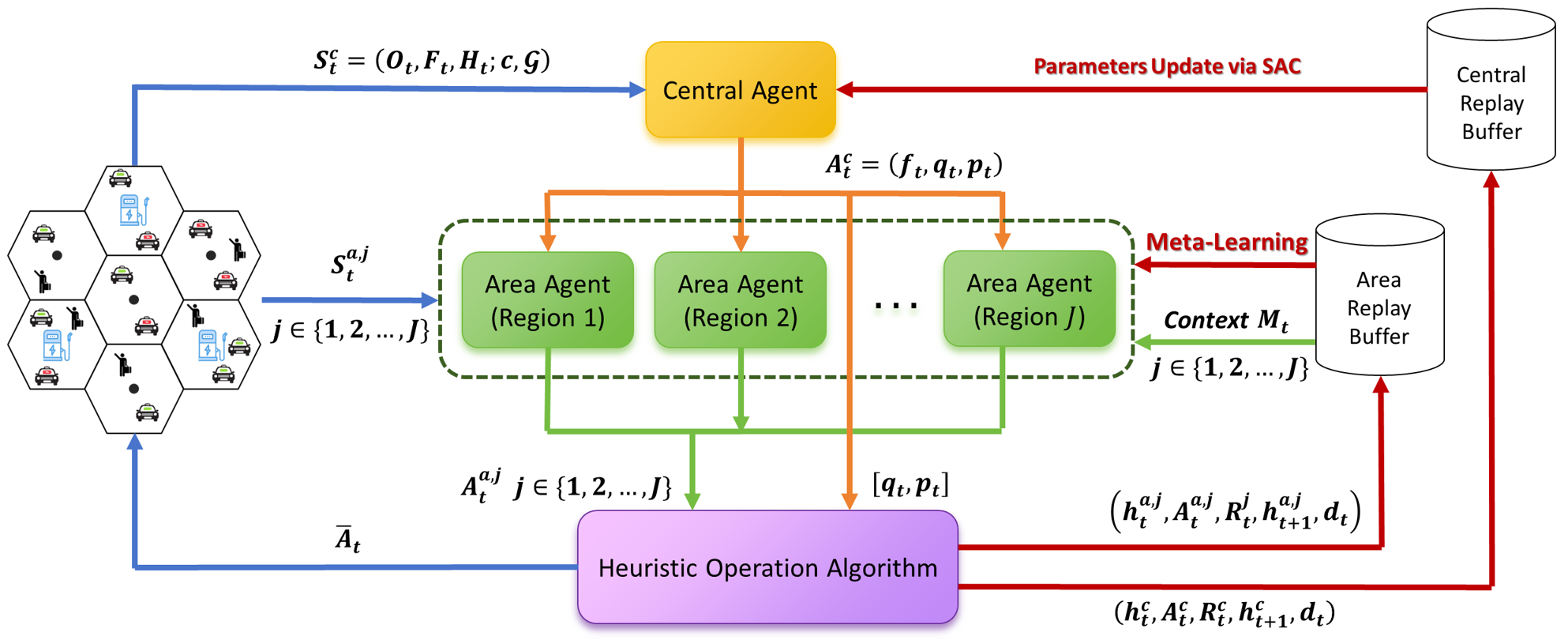}
\caption{The hierarchical multi-agent meta-reinforcement learning architecture.}
\label{fig:overall}
\end{figure}

The CMDP framework formulated above provides a rigorous description of the AET fleet operation problem, explicitly capturing the structural uncertainty induced by infrastructure evolution. However, solving this global CMDP with a unified policy is computationally intractable for city-scale applications. The high dimensional state space, the combinatorial explosion of the joint action space, and the complex dependency on the latent context $\mathbf{c}$ render direct optimization impractical. To address these computational challenges while preserving the capability to adapt to structural changes, we reformulate the original global CMDP using a hierarchical decision architecture. This framework decomposes the control problem across two hierarchically coordinated levels: a global strategic level and a local tactical level, shown in Figure~\ref{fig:overall}.

At the upper level, we define a Central Agent that functions as a global coordinator. This agent operates on a global MDP defined over aggregated system states. Its primary objective is to manage the strong spatial coupling across different regions by coordinating the distribution of vehicle resources, regulating energy infrastructure usage, and shaping economic incentives. By abstracting away the granular details of local infrastructure topology, the Central Agent focuses on global system efficiency, generating high-level guidance signals to coordinate the lower layer. These signals specifically include target net flow ratios to guide vehicle rebalancing, dynamic charging quotas to mitigate station congestion, and region specific cost multipliers to steer fleet movements via soft economic incentives. This multidimensional signaling mechanism allows the Central Agent to align local tactical decisions with system-wide operational goals effectively.

At the lower level, we deploy multiple Area Agents, each governing the tactical operations within a specific region. Unlike the Central Agent, these Area Agents interact directly with the system dynamics described in section~\ref{sec:CMDP}. We therefore model each Area Agent as a local CMDP whose transitions and feasible charging actions are modulated by an episode-level meta-context $\mathbf{c}$ induced by layout $C$. Conditioned on this meta-context, each Area Agent executes matching, repositioning, and charging decisions within its region. This decomposition confines the challenge of structural adaptation to the local level: the Area Agents infer the meta-context from local interaction data and adjust their policies accordingly, while the Central Agent provides global coordination signals to align these context-adaptive local policies with system-wide operational objectives.

The interaction between the two levels is facilitated through an explicit coordination mechanism. The Central Agent does not dictate atomic vehicle movements; instead, it modifies the reward structures or constraint boundaries for the Area Agents. In response, the Area Agents optimize their local actions to satisfy regional demand and fleet conditions, subject to the strategic guidance provided by the Central Agent. This hierarchical reformulation effectively decouples the global resource allocation problem from the local context adaptation problem, enabling scalable learning and robust control in evolving environments.

\subsection{Central Agent}

\begin{figure}[!htbp]
\centering
\includegraphics[width=\linewidth]{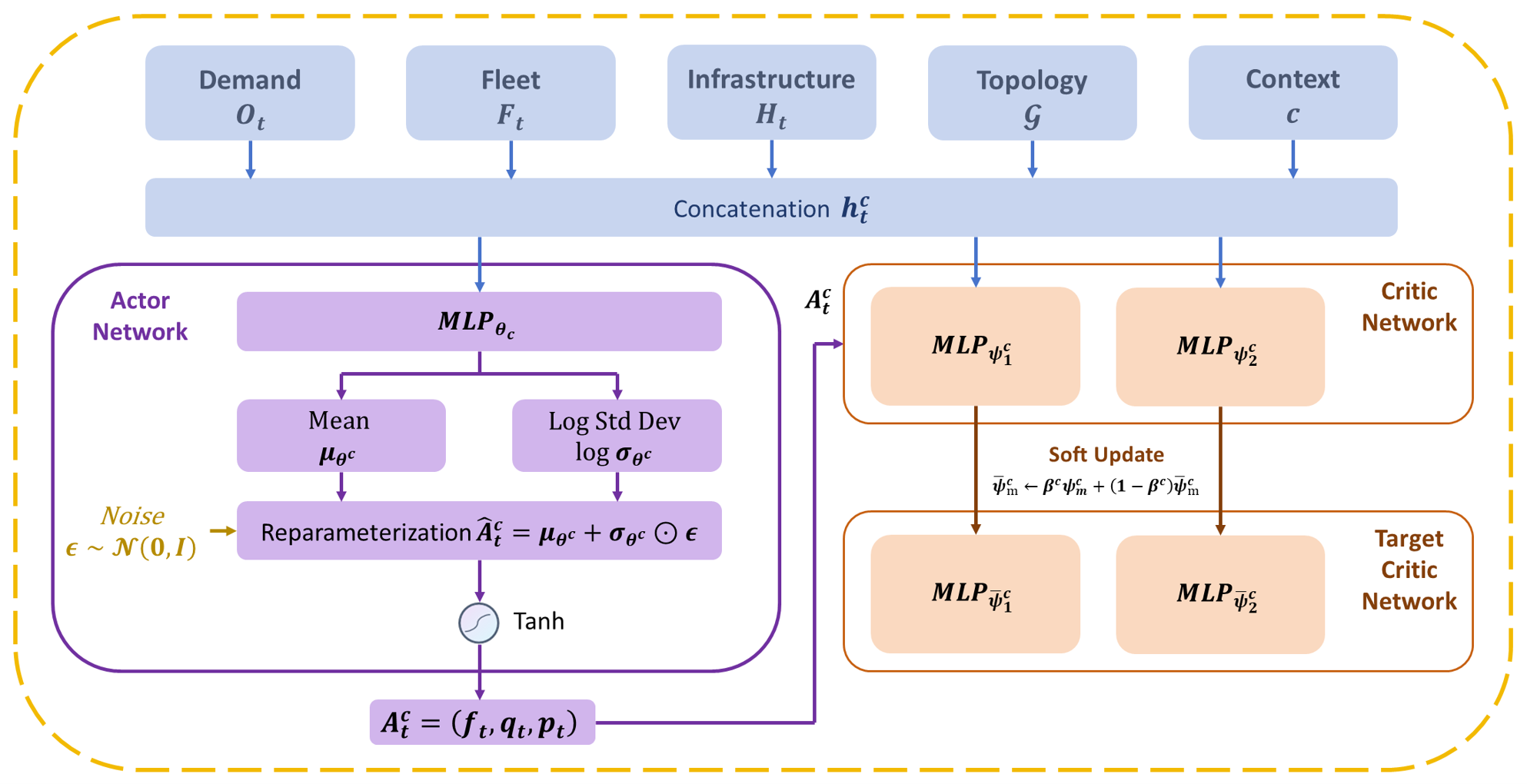}
\caption{Architecture of the hierarchical Central Agent framework.}
\label{fig:centralagent}
\end{figure}

Building on the hierarchical MDP reformulation, we specify the decision making structure of the Central Agent. As illustrated in Figure~\ref{fig:centralagent}, the Central Agent acts as the strategic coordinator of the AET fleet and regulates the spatiotemporal allocation of vehicle resources at the system level. Rather than executing vehicle-level actions, it produces high-level control signals that guide decentralized regional operations.

\subsubsection{MDP formulation}

We model the Central Agent as a global Markov Decision Process (MDP), denoted by $(\mathbf{S}_t^{c}, \mathbf{A}_t^{c}, \mathbf{P}_t^{c}, R_t^{c}, d_t)$, where the state aggregates information across the network on demand, fleet availability, charging infrastructure, and spatial topology. The Central Agent learns a strategic policy that maps aggregated system states to strategic control actions. In the following, we detail the specific design of the Central Agent's state space, action space, transition mechanism, reward function and terminal signal.

\noindent \textbf{State space} $\mathbf{S}_t^{c}$: The global state $\mathbf{S}_t^{c}$ provides a system level representation of supply demand conditions and infrastructure status at period $t$. It is defined as a composite tuple $\mathbf{S}_t^{c} = (\mathbf{O}_t, \mathbf{F}_t, \mathbf{H}_t; \mathbf{c}, \mathcal{G})$, which captures the dynamic operational status and the static environmental context:
\begin{enumerate}
    \item \textit{Demand dynamics} $\mathbf{O}_t$: This component aligns with the exogenous demand process defined in the SDRAP formulation. It captures both realized and anticipated passenger demand. In particular, it includes the origin destination matrix $\mathbf{Q}_t \in \mathbb{R}^{|\mathcal{J}| \times |\mathcal{J}|}$ of unserved requests, where each entry is defined by the OD aggregation of the underlying request set $Q_t(i,j):= \left|\left\{ r \in \mathcal{R}_t : j_r^{\mathrm{o}} = i,\; j_r^{\mathrm{d}} = j \right\}\right|, \quad \forall i,j \in \mathcal{J}$. To mitigate myopic behavior, $\mathbf{O}_t$ also incorporates a demand forecast tensor $\hat{\mathbf{Q}}_{t:t+\Delta_t}$ that summarizes expected order volumes over the subsequent $\Delta_t$ periods.
    \item \textit{Fleet supply distribution} $\mathbf{F}_t$: This component characterizes vehicle availability across regions. For each region $j \in \mathcal{J}$, it includes (i) the count of vehicles by operational status (idle, charging, relocating, and serving), (ii) a forecast of incoming vehicles that are currently en route and expected to arrive in future periods, and (iii) the aggregate energy condition of the regional fleet, represented by the average SOC.
    \item \textit{Charging status} $\mathbf{H}_t$: While the infrastructure layout is static within an episode, its operational status evolves dynamically. This component records real-time occupancy information and effective remaining capacity, which is critical for managing congestion and coordinating charging decisions.
    \item \textit{Static context and topology} $(\mathbf{c}, \mathcal{G})$: This component encodes the time-invariant structures. $\mathbf{c}$ represents the infrastructure context (station layout and the count of charging piles) as defined in the CMDP formulation, while $\mathcal{G}$ encodes the spatial structure of the transportation network, including travel distances and energy consumption between different regions.
\end{enumerate}

\noindent \textbf{Action space} $\mathbf{A}_t^{c}$: The Central Agent influences the behavior of decentralized Area Agents through a continuous action vector $\mathbf{A}_t^{c}$. Rather than prescribing actions at the level of individual vehicles, these signals shape regional decision making by imposing targets and incentives at the system level that guide local optimization and discourage uncoordinated behavior. Formally, the action vector is defined as $\mathbf{A}_t^{c}=(\mathbf{f}_t, \mathbf{q}_t, \mathbf{p}_t)$, comprising the following components:
\begin{itemize}
    \item \textit{Net flow ratio} $\mathbf{f}_t$: The vector $\mathbf{f}_t \in [-f_{\max}, f_{\max}]^{|\mathcal{J}|}$ specifies a normalized target net vehicle flow ratio across the whole map. Positive values encourage net inflow of vehicles into the region, while negative values promote net outflow. These targets are subject to a global flow balance condition, $\sum_{j \in \mathcal{J}} f_t^{j} \approx 0$, which ensures that the Central Agent issues redistribution signals that are globally neutral on average. The actual relocation magnitudes are determined endogenously by the Area Agents under local feasibility constraints and fleet availability, so explicit weighting by regional fleet size is not required at the signaling level. The condition is enforced as a soft constraint to allow flexibility at the decentralized execution layer.

    \item \textit{Charging quota} $\mathbf{q}_t$: The vector $\mathbf{q}_t \in [q_{\min}, 1]^{|\mathcal{J}|}$ specifies the maximum fraction of charging capacity utilized in each region. By adjusting these quotas dynamically, the Central Agent can reserve charging resources in anticipation of future demand and mitigate congestion at heavily utilized stations.
    
    \item \textit{Dynamic cost multiplier} $\mathbf{p}_t$: The vector $\mathbf{p}_t \in [-p_{\max}, p_{\max}]^{|\mathcal{J}|}$ specifies region-specific adjustment factors applied to the effective compensation of passenger orders. By increasing or decreasing the marginal reward associated with serving requests in different regions, these multipliers act as soft incentives that steer Area Agents toward demand hotspots and improve spatial alignment between vehicle supply and passenger demand.
\end{itemize}

\noindent \textbf{State transition}: The evolution of the Central Agent state is governed jointly by its current state $\mathbf{S}_t^{c}$, the issued control action $\mathbf{A}_t^{c}$, and the exogenous environmental components $\Theta_t$. After the Central Agent selects $\mathbf{A}_t^{c}$, the system evolves through the decentralized execution of Area Agents and heuristic operation module. The global transition dynamics can be viewed as an aggregation of local transitions induced by the lower level policies. Formally, this is represented by the transition kernel $\mathbf{P}_t^{c}(\mathbf{S}_{t+1}^{c} \mid \mathbf{S}_t^{c}, \mathbf{A}_t^{c}; \Theta_t)$. The Central Agent does not directly affect the exogenous processes; instead, it influences system evolution indirectly by shaping the incentives and soft constraints under which the Area Agents make local decisions. At the end of period $t$, the next global state $\mathbf{S}_{t+1}^{c}$ is constructed by aggregating the resulting vehicle distributions, realized demand, and infrastructure status.

\noindent \textbf{Reward design}: To ensure that the strategic policy learned by the Central Agent is both effective and implementable, we adopt a regularized reward formulation that balances global operational performance with execution feasibility. The reward at period $t$ is defined as
\begin{equation}
    R_t^{c}(\mathbf{S}_t^{c}, \mathbf{A}_t^{c}) 
    = \sum_{j \in \mathcal{J}} R_t^{j} 
    - \lambda \left\| \mathbf{A}_t^{c} - \bar{\mathbf{A}}_t \right\|_2^2
\end{equation}
where $R_t^{j}$ denotes the realized operational reward aggregated over region $j$ at period $t$. In particular, $R_t^{j}$ is constructed from the underlying order-level revenue terms $\varrho_r$ and operational costs, and is explicitly adjusted by the Central Agent's incentive signal $p_t^j$; see Eq.~(\ref{eq:area_reward}) for its detailed definition. The second term introduces a quadratic penalty weighted by $\lambda$, measuring the deviation between the strategic control signals issued by the Central Agent, $\mathbf{A}_t^{c}$, and the aggregate actions actually realized by the system, $\bar{\mathbf{A}}_t$. This regularization serves to bridge the gap between high-level commands and low-level execution, discouraging the Central Agent from issuing overly aggressive or infeasible commands that violate the operational constraints faced by the Area Agents.

\noindent \textbf{Termination signal}: Finally, the tuple includes a binary termination signal $d_t$ which marks the conclusion of the finite planning horizon. It is formally defined as:
\begin{equation}
    d_t =
    \begin{cases}
    1, & \text{if } t = T, \\
    0, & \text{otherwise}.
    \end{cases}
\label{eq:central_termination}
\end{equation}
When $d_t=1$, the episode terminates, and no further state transitions occur. In the context of value estimation, this signal ensures that the bootstrapping of future rewards is correctly truncated at the end of the horizon.

\subsubsection{Network architecture}

The Central Agent employs a Soft Actor-Critic (SAC) framework to learn a stochastic policy that jointly maximizes expected cumulative rewards and policy entropy. At each decision period $t$, to construct a vectorized representation amenable to neural processing, all observable components of the global state $\mathbf{S}_t^{c}$ are directly concatenated into a single feature vector. Specifically, the input vector $\mathbf{h}_t^c$ integrates the dynamic operational information (demand $\mathbf{O}_t$, fleet $\mathbf{F}_t$, infrastructure status $\mathbf{H}_t$) with the static structural attributes (infrastructure context $\mathbf{c}$ and network topology $\mathcal{G}$):
\begin{equation}
    \mathbf{h}_t^c
    =
    \mathrm{Concat}\!\left(
    \mathbf{O}_t,\,
    \mathbf{F}_t,\,
    \mathbf{H}_t,\,
    \mathbf{c},\,
    \mathcal{G}
    \right),
    \qquad
    \mathbf{h}_t^c \in \mathbb{R}^{d_{\mathrm{in}}},
\end{equation}
where $d_{\mathrm{in}}$ denotes the dimension of the concatenated feature vector. The inclusion of the infrastructure context $\mathbf{c}$ is critical, as it enables the policy network to distinguish between different topological configurations and adjust its strategic signals accordingly. The resulting feature vector $\mathbf{h}_t^c$ serves as the shared input to both the actor and critic networks of the Central Agent.

The actor network, parameterized by $\theta^{c}$, is designed to approximate a stochastic action distribution conditioned on the global state representation. Given the feature vector $\mathbf{h}_t^{c}$, the actor network maps the input through a multilayer perceptron (MLP) to output the parameters of a diagonal Gaussian distribution, namely the mean vector $\boldsymbol{\mu}_{\theta^{c}}(\mathbf{h}_t^{c})$ and the logarithmic standard deviation vector $\log \boldsymbol{\sigma}_{\theta^{c}}(\mathbf{h}_t^{c})$:
\begin{equation}
    \boldsymbol{\mu}_{\theta^{c}}(\mathbf{h}_t^{c}), \;
    \log \boldsymbol{\sigma}_{\theta^{c}}(\mathbf{h}_t^{c})
    =
    \mathrm{MLP}_{\theta^{c}}\!\left(\mathbf{h}_t^{c}\right).
\end{equation}
This parameterization defines a state dependent stochastic policy prior to action squashing. Using the logarithm of the standard deviation ensures numerical stability during optimization and guarantees that the resulting standard deviation $\boldsymbol{\sigma}_{\theta^{c}}(\mathbf{h}_t^{c})$ remains strictly positive after exponentiation.

To allow gradients to propagate through the stochastic action sampling process, we employ the reparameterization trick. Instead of sampling the action directly from the Gaussian policy, we first draw an auxiliary noise vector $\boldsymbol{\epsilon}$ from a standard normal distribution and construct a raw action sample as an affine transformation:
\begin{equation}
\hat{\mathbf{A}}_t^{c}
=
\boldsymbol{\mu}_{\theta^{c}}(\mathbf{h}_t^{c})
+
\boldsymbol{\sigma}_{\theta^{c}}(\mathbf{h}_t^{c}) \odot \boldsymbol{\epsilon},
\qquad
\boldsymbol{\epsilon} \sim \mathcal{N}(\mathbf{0}, \mathbf{I}).
\end{equation}
This formulation separates the stochasticity of the policy from its deterministic parameters, making $\hat{\mathbf{A}}_t^{c}$ a differentiable function of $\theta^{c}$. As a result, policy gradients can be computed efficiently using standard backpropagation, which is essential for stable actor optimization.

Since the physical action space of the fleet control problem is bounded, the raw Gaussian sample $\hat{\mathbf{A}}_t^{c}$ is subsequently transformed using a hyperbolic tangent function:
\begin{equation}
\mathbf{A}_t^{c} = \tanh\!\left(\hat{\mathbf{A}}_t^{c}\right).
\end{equation}
This squashing operation maps the unbounded real valued output to a compact interval $[-1, 1]$. To generate the executable control signals defined in the MDP formulation, the squashed action $\mathbf{A}_t^{c}$ is linearly rescaled element-wise to the specific physical ranges of the decision variables. Specifically, the output dimensions corresponding to the net flow ratio $\mathbf{f}_t$, charging quota $\mathbf{q}_t$, and cost multiplier $\mathbf{p}_t$ are mapped to $[-f_{\max}, f_{\max}]$, $[q_{\min}, 1]$, and $[-p_{\max}, p_{\max}]$, respectively. This rescaling is applied only at the interface with the environment. During forward propagation and backpropagation within the policy network, optimization is performed on the normalized action $\mathbf{A}_t^{c}$. This design decouples network training from problem specific action magnitudes, preserving stable gradient flow while ensuring that executed actions remain feasible under system level constraints.

\subsection{Area Agent}
\begin{figure}[!htbp]
\centering
\includegraphics[width=\linewidth]{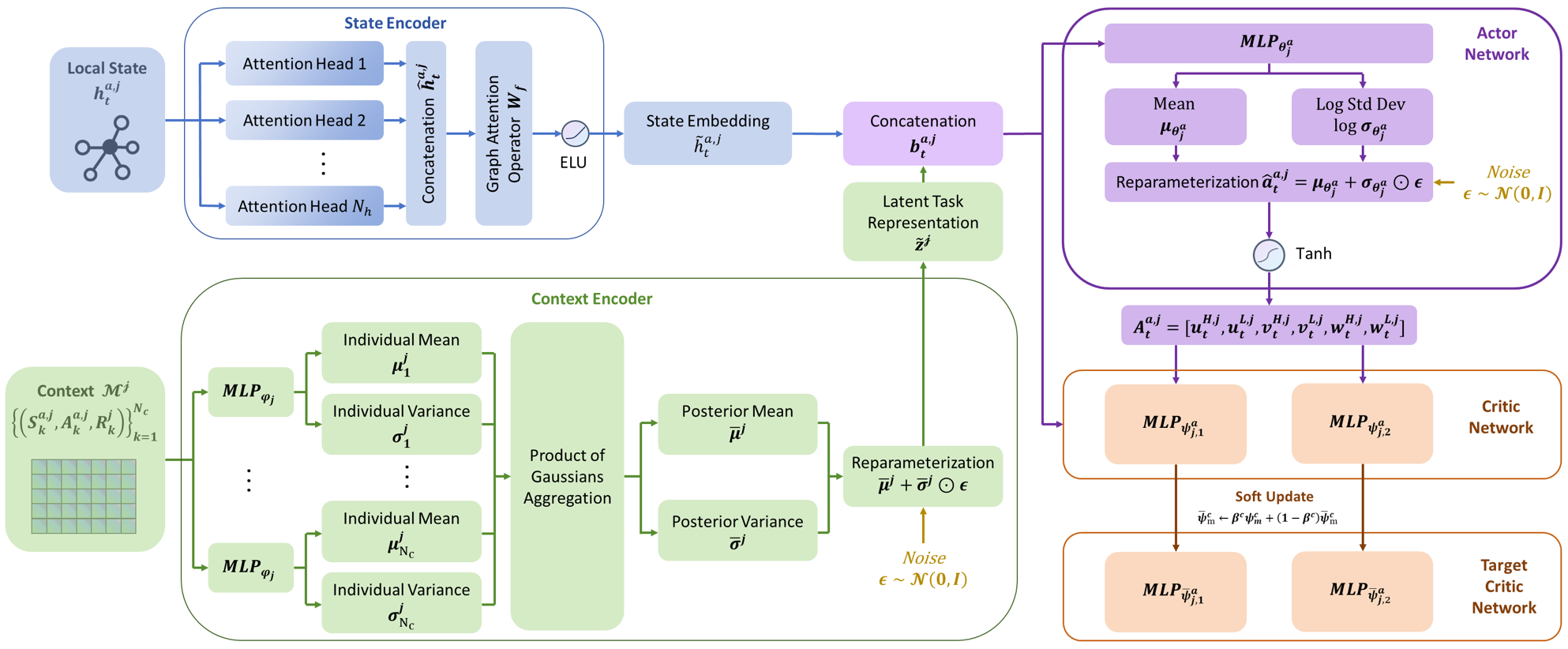}
\caption{Architecture of the Area Agents.}
\label{fig:areaagent}
\end{figure}

To complement the strategic coordination performed by the Central Agent, we introduce a set of decentralized Area Agents that are responsible for regional tactical decision making within the hierarchical framework, as illustrated in Fig. \ref{fig:areaagent}. Specifically, one Area Agent is assigned to each spatial region $j \in \mathcal{J}$ to manage localized operational decisions. By decomposing the global control problem into a collection of region specific decision processes, this design avoids reliance on a single centralized controller and improves both scalability and responsiveness.

\subsubsection{CMDP formulation}\label{subsubsec_CMDP}

All Area Agents share an identical policy architecture but are trained as independent learners with separate parameters. This design ensures structural consistency across regions while allowing each agent to specialize its policy to local demand patterns, fleet conditions, and the specific infrastructure context $\mathbf{c}$. Each Area Agent associated with region $j$ is formulated as a local CMDP, defined by the tuple $(\mathbf{S}_t^{a,j}, \mathbf{A}_t^{a,j}, \mathbf{P}_t^{a}, R_t^{j}, d_t)$. This CMDP governs sequential decision making based on local observations and the strategic control signals issued by the Central Agent.

\noindent \textbf{State space}:
The state vector $\mathbf{S}_t^{a,j}$ is designed to provide a comprehensive regional view of the system, integrating local operational data with global strategic signals to mitigate partial observability. It consists of three primary components:
\begin{enumerate}
    \item \textit{Local operational state}. This component captures the immediate supply and demand conditions within region $j$. It includes the regional fleet distribution (categorized by operational status: idle, charging, relocating, serving), the energy condition represented by the average SOC, and the local realization of the demand process (outstanding requests originating from or destined for region $j$). Additionally, it incorporates a deterministic inflow forecast vector derived from the travel time of vehicles dispatched in earlier periods.
    \item \textit{Strategic and structural context}. To promote coordination and adaptation, the agent observes two types of external signals. First, it receives the \textit{strategic control signals} from the Central Agent: the net flow ratio target $f_t^j$, the charging quota $q_t^j$, and the cost multiplier $p_t^j$. Second, it perceives the \textit{structural context} derived from $(\mathbf{c}, \mathcal{G})$, including the local charging station layout (availability and capacity) and the topological connectivity with neighboring regions. This structural information is critical for the agent to infer the latent infrastructure context and adapt its policy via the meta-learning module.
    \item \textit{Temporal encoding}. To capture the strong temporal regularities in urban mobility demand, the discrete decision period is embedded into a continuous representation. We utilize a learnable embedding matrix $\mathbf{E} \in \mathbb{R}^{T \times h_e}$, where the row vector $\mathbf{e}_t$ corresponds to period $t$. Here, $h_e$ denotes the embedding dimension. This temporal encoding enables the policy network to represent periodic demand patterns and dependencies over long time horizons.

\end{enumerate}

\noindent \textbf{Action space}:
The action $\mathbf{A}_t^{a,j}$ represents a high-level tactical directive rather than a direct vehicle-level control. It is a continuous vector that parameterizes the low-level heuristic operation algorithm within region $j$. Through this interface, the Area Agent regulates the allocation of available vehicles across competing tasks under the guidance of the Central Agent.

To balance service quality and energy sustainability, the action space explicitly differentiates vehicles by energy status. Specifically, the action is composed of six destination indexed sub vectors, each in $\mathbb{R}^{|\mathcal{J}|}$:
\begin{equation}
\mathbf{A}_t^{a,j}
=
\big[
\mathbf{u}_t^{\mathrm{H},j}, \mathbf{u}_t^{\mathrm{L},j},
\mathbf{v}_t^{\mathrm{H},j}, \mathbf{v}_t^{\mathrm{L},j},
\mathbf{w}_t^{\mathrm{H},j}, \mathbf{w}_t^{\mathrm{L},j}
\big],
\qquad
\mathbf{u}_t^{\cdot,j},\mathbf{v}_t^{\cdot,j},\mathbf{w}_t^{\cdot,j}\in\mathbb{R}^{|\mathcal{J}|}.
\end{equation}
Here, $\mathbf{u}$, $\mathbf{v}$, and $\mathbf{w}$ correspond to flow thresholds for passenger matching, vehicle repositioning, and charging decisions, respectively. The superscripts $\mathrm{H}$ and $\mathrm{L}$ indicate whether the thresholds apply to vehicles with high or low SOC, as determined by a predefined energy threshold. These action components act as soft capacity limits for the heuristic dispatcher. For example, the $k$-th element of $\mathbf{u}_t^{\mathrm{H},j}$ specifies the maximum allowable allocation of high-SOC vehicles from region $j$ to serve passenger requests destined for region $k$. In this way, the Area Agent shapes spatial vehicle flows based on learned value assessments, while delegating the granular assignment to the execution layer.

\noindent \textbf{State transition}:
The local environment of Area Agent $j$ evolves according to a context-dependent transition kernel
\begin{equation}
\mathbf{P}^{a}\!\left(
\mathbf{S}_{t+1}^{a,j}
\mid
\mathbf{S}_t^{a,j},
\mathbf{A}_t^{a,j},
\mathbf{c}
\right),
\end{equation}
where $\mathbf{c}=\Gamma(C)$ denotes the episode-level infrastructure context. The stochasticity in the transition is primarily driven by demand realizations and charging completion processes over operational intervals.

At each period $t$, Area Agent $j$ outputs a tactical action $\mathbf{A}_t^{a,j}$, which serves as a set of soft flow thresholds for the local heuristic operation module. Given these thresholds, the heuristic module assigns individual vehicles to passenger service, repositioning, and charging tasks during the subsequent operational intervals. The next state $\mathbf{S}_{t+1}^{a,j}$ is observed after the execution phase concludes. In this way, the Area Agent influences system evolution indirectly through heuristic execution.

\noindent \textbf{Reward design}:
The reward function provides the primary learning signal for regional operational efficiency. For each Area Agent $j$, the reward $R_t^{j}$ is defined as the net profit generated within the region, explicitly adjusted by the Central Agent's economic incentives. It is formulated as:
\begin{equation}
\label{eq:area_reward}
R_t^{j}
=
    \sum_{r \in \mathcal{R}_t^{j,+}} \varrho_r \cdot (1 + p_t^j)
    -
    \left(
    \sum_{k \in \mathcal{K}_t^{r,j}} c_{v}^{r}
    +
    \sum_{k \in \mathcal{K}_t^{c,j}} c_{v}^{c}
    \right).
\end{equation}
Here, $\mathcal{R}_t^{j,+}$ denotes the set of served requests originating in region $j$ during period $t$, yielding revenue $\varrho_r$ for request $r$. The term $(1 + p_t^j)$ incorporates the \textit{dynamic cost multiplier} issued by the Central Agent, effectively altering the marginal utility of serving passengers in region $j$. This mechanism aligns the local agent's greedy profit maximization with global strategic goals (e.g., incentivizing service in high demand areas). The cost terms $c_v^{r}$ represents the operational cost incurred by vehicle $k$ due to empty repositioning movements and $c_v^{c}$ denotes the cost associated with energy consumption incurred by vehicle $k$ during charging activities.

\noindent \textbf{Termination signal}: The Area Agents operate under the same finite planning horizon $T$ as the Central Agent. Therefore, the termination signal $d_t$ shares the same definition as Eq.~(\ref{eq:central_termination}), enforcing a unified termination condition for both levels of the hierarchy.

\subsubsection{Network architecture}

We design the Area Agent based on the SAC framework, augmented with a probabilistic context encoder and a graph attention mechanism. The overall architecture consists of four modules: a state encoder, a context encoder, a policy network, and dual critic networks.

\noindent \textbf{State encoder with graph attention.}
Building on the state space definition introduced in Section~\ref{subsubsec_CMDP}, we employ GAT to encode regional state features. The raw state $\mathbf{S}_t^{a,j}$ of Area Agent $i$ integrates local operational conditions, strategic signals, and temporal embeddings. However, treating these states in isolation ignores the strong spatial coupling inherent in fleet dynamics. Furthermore, relying on absolute spatial coordinates can hinder generalization across different infrastructure layouts. To address this, we utilize GAT to extract a topological representation that captures the complex spatial heterogeneity and underlying topological structures of charging station layouts. This approach offers two key advantages. First, it focuses on relative connectivity and neighborhood dependencies rather than absolute coordinates of regions, so that each region is characterized by its topological role in the graph network and the influence propagated from adjacent regions through attention. Second, it extracts robust spatial representations that are invariant to geometric symmetries in infrastructure layouts, thereby improving generalization across different scenarios.

\begin{figure}[!htbp]
    \centering
    \includegraphics[width=0.4\linewidth]{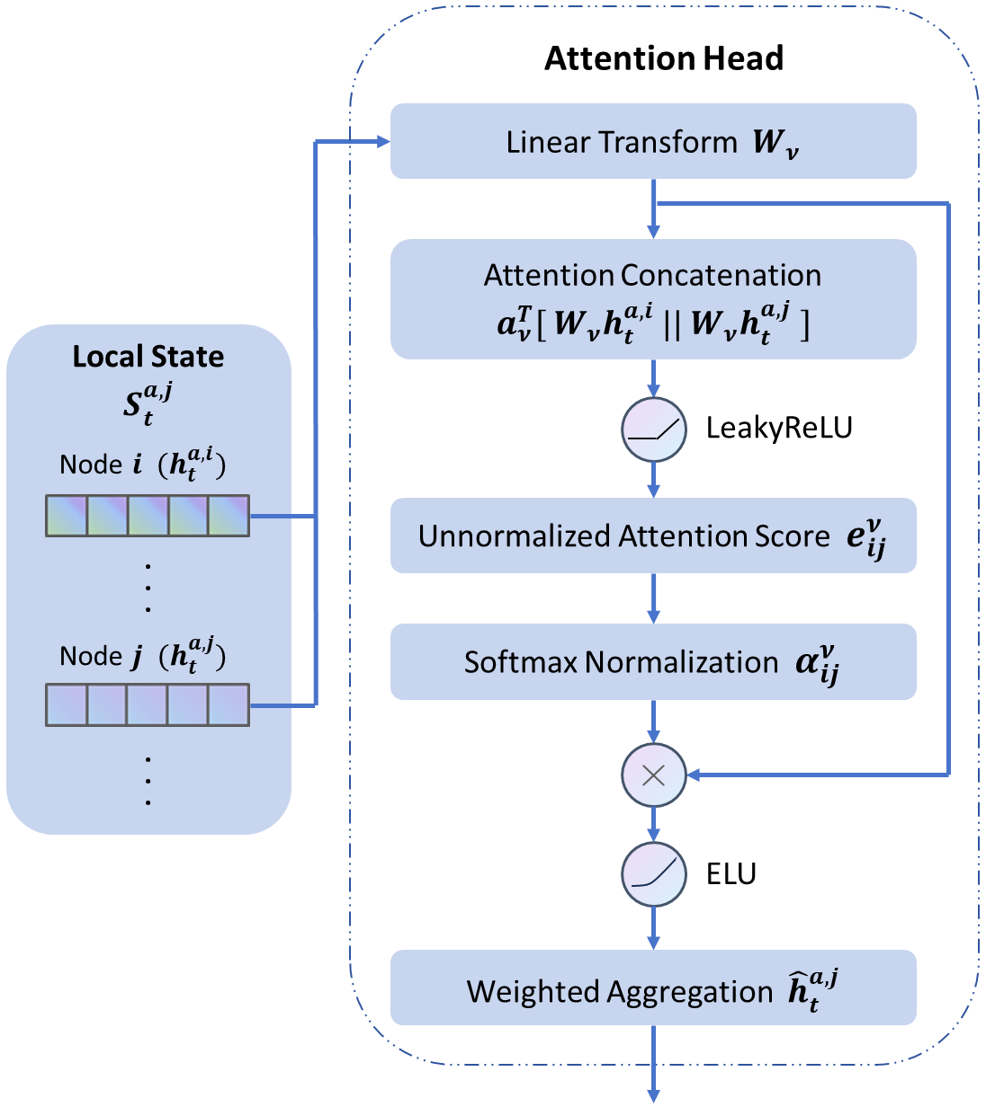}     
    \caption{The internal processing pipeline of a single-attention head.}
    \label{fig:gat_structure} 
\end{figure}

We first construct the initial node feature vector $\mathbf{h}_{t}^{a,i}$ for each region $i$ by flattening and concatenating the three components of the CMDP state tuple $\mathbf{S}_t^{a,i}$ (local operational state, strategic context, and temporal encoding). Using the underlying network topology $\mathcal{G}$ to define the neighborhood set $\mathcal{N}_i$ for each region, the GAT computes attention coefficients that quantify the relevance of neighboring region $j$ to region $i$. To improve training stability and representation expressiveness, a multi-head attention mechanism is adopted in the first layer. The detailed information flow within a single-head attention head is depicted in Figure~\ref{fig:gat_structure}. For each attention head $\nu \in \{1,\dots,N_h\}$, the node features are first projected into a lower-dimensional space using a learnable weight matrix $\mathbf{W}_\nu \in \mathbb{R}^{F_{\mathrm{in}} \times F'}$. Here, $F_{\mathrm{in}}$ is the input feature dimension of each node, and $F'$ is the per-head output dimension after linear projection. The unnormalized attention score between regions $i$ and $j$ is then computed as:
\begin{equation}
e_{ij}^\nu
    =
    \text{LeakyReLU}
    \left(
    \mathbf{a}_\nu^\top
    \left[
    \mathbf{W}_\nu \mathbf{h}_{t}^{a,i} \;\Vert\; \mathbf{W}_\nu \mathbf{h}_{t}^{a,j}
    \right]
    \right),
\end{equation}
where $\mathbf{a}_\nu \in \mathbb{R}^{2F'}$ is a learnable attention vector, and $\Vert$ denotes vector concatenation.

These scores are normalized over the neighborhood $\mathcal{N}_i$ using a softmax function,
\begin{equation}
\alpha_{ij}^\nu
=
\frac{\exp(e_{ij}^\nu)}
{\sum_{j' \in \mathcal{N}_i} \exp(e_{ij'}^\nu)},
\end{equation}
yielding adaptive weights $\alpha_{ij}^\nu$ that emphasize information from task relevant neighbors (e.g., adjacent regions with high demand or available charging stations) while suppressing less informative connections.

The features from all $N_h$ attention heads are aggregated via concatenation to preserve diverse relational patterns. This produces a high-dimensional intermediate representation:
\begin{equation}
\hat{\mathbf{h}}_{t}^{a,i}
    =
    \bigg\Vert_{\nu=1}^{N_h}
    \text{ELU}
    \left(
    \sum_{j \in \mathcal{N}_i}
    \alpha_{ij}^\nu \mathbf{W}_\nu \mathbf{h}_{t}^{a,j}
    \right).
\end{equation}

In the final layer, the concatenated representation $\hat{\mathbf{h}}_{t}^{a,i}$ is processed by a single-head graph attention operator to produce the compact spatial embedding. A new set of attention coefficients $\bar{\alpha}_{ij}$ and a learnable projection matrix $\mathbf{W}^{\mathrm{f}}$ are applied to perform an attentive fusion:
\begin{equation}
    \tilde{\mathbf{h}}_{t}^{a,i}
    =
    \text{ELU}
    \left(
    \sum_{j \in \mathcal{N}_i}
    \bar{\alpha}_{ij}
    \mathbf{W}^{\mathrm{f}}
    \hat{\mathbf{h}}_{t}^{a,j}
    \right).
\end{equation}
The resulting vector $\tilde{\mathbf{h}}_{t}^{a,i}$ serves as the spatially contextualized state embedding for region $i$. By effectively aggregating information from the local neighborhood $\mathcal{N}_i$, this embedding allows the Area Agent to make decentralized decisions that are implicitly coordinated with the conditions of surrounding regions, thereby mitigating the limitations of partial observability.

\noindent \textbf{Probabilistic context encoder.}
While GAT extracts explicit spatial features from the current observation, it does not capture the unobserved variations in system dynamics caused by the evolving charging infrastructure. Different charging station layouts define a family of related but dynamically distinct tasks, where each task imposes unique feasibility constraints on dispatching and charging decisions. To enable generalization across these heterogeneous layouts without retraining, we augment the architecture with a probabilistic context encoder parameterized by $\phi$. Unlike standard approaches that rely solely on instantaneous state inputs, this module infers a latent task representation $\tilde{\mathbf{z}}^j$ from the agent's interaction history. This inference-based design is motivated by the need to capture implicit operational constraints, such as effective station availability and regional congestion dynamics, which are not immediately visible in the static topology map but can be inferred from transition experiences.

We define the context set $\mathcal{M}^{j}$ for the $j$-th Area Agent as an unordered collection of $N_c$ context tuples sampled from the agent's context buffer under the current infrastructure context. Specifically, the context element is given by
\begin{equation}
\mathcal{M}^{j}
=
\left\{
m_k^{j}
\right\}_{k=1}^{N_c},
\qquad
m_k^{j}
=
\left(
\mathbf{S}_{k}^{a,j},
\mathbf{A}_{k}^{a,j},
R_{k}^{j}
\right),
\end{equation}
where $\mathbf{S}_{k}^{a,j}$, $\mathbf{A}_{k}^{a,j}$, and $R_{k}^{j}$ denote the state, action, and realized reward of the $j$-th Area Agent corresponding to the $k$-th context sample retrieved from the buffer. Note that $\mathbf{S}_{k}^{a,j}$ here includes the full state vector defined in the CMDP formulation, allowing the encoder to correlate operational outcomes with both local conditions and strategic signals.

To obtain a permutation-invariant representation, we model the latent context as a random variable $\mathbf{Z}^j$ whose posterior distribution is conditioned on the set $\mathcal{M}^{j}$. We factorize this posterior as a product of independent factors corresponding to each context tuple:
\begin{equation}
\mathcal{Q}_{\phi_j}\!\left(\mathbf{Z}^j \mid \mathcal{M}^{j}\right)
\propto
\prod_{k=1}^{N_c}
\Psi_{\phi_j}\!\left(\mathbf{Z}^j \mid m_k^{j}\right),
\end{equation}
where each factor $\Psi_{\phi_j}(\mathbf{Z}^j \mid m_k^{j})$ represents the local probabilistic belief derived from a single context item.

The encoding procedure consists of three stages. First, each context item $m_k^{j}$ is processed independently by a multilayer perceptron $MLP_{\phi_j}(\cdot)$. This network maps the tuple to the parameters of a Gaussian distribution associated with the $k$-th individual experience:
\begin{equation}
\Psi_{\phi_j}\!\left(\mathbf{Z}^j \mid m_k^{j}\right)
=
\mathcal{N}\!\left(
\boldsymbol{\mu}_{k}^j,\,
\mathrm{diag}\big((\boldsymbol{\sigma}_{k}^j)^2\big)
\right),
\qquad
(\boldsymbol{\mu}_{k}^j,\boldsymbol{\sigma}_{k}^j)
=
MLP_{\phi_j}\!\left(
\mathbf{S}_{k}^{a,j},
\mathbf{A}_{k}^{a,j},
R_{k}^{j}
\right),
\end{equation}
where the standard deviation $\boldsymbol{\sigma}_{k}^j$ is constrained to be positive via a Softplus activation function.

Second, since the context set $\mathcal{M}^{j}$ is unordered, the aggregation must be permutation invariant to ensure consistent inference regardless of the sequence of collected samples. We form the aggregated variational posterior $\mathcal{Q}_{\phi_j}(\cdot \mid \mathcal{M}^{j})$ by multiplying the individual Gaussian factors. Because the product of Gaussian PDFs is proportional to another Gaussian PDF, the parameters of the resulting posterior distribution $\mathcal{N}(\bar{\boldsymbol{\mu}}^j, \text{diag}((\bar{\boldsymbol{\sigma}}^j)^2))$ can be computed analytically using the Product-of-Gaussian rule:
\begin{equation}
(\bar{\boldsymbol{\sigma}}^j)^2
=
\left(
\sum_{k=1}^{N_c}
\frac{1}{(\boldsymbol{\sigma}_{k}^j)^2}
\right)^{-1},
\qquad
\bar{\boldsymbol{\mu}}^j
=
(\bar{\boldsymbol{\sigma}}^j)^2
\odot
\left(
\sum_{k=1}^{N_c}
\frac{\boldsymbol{\mu}_{k}^j}{(\boldsymbol{\sigma}_{k}^j)^2}
\right).
\label{eq:pog_no_prior}
\end{equation}

Finally, the latent task embedding $\tilde{\mathbf{z}}^j$ is sampled using the reparameterization trick:
\begin{equation}
\tilde{\mathbf{z}}^j
=
\bar{\boldsymbol{\mu}}^j
+
\bar{\boldsymbol{\sigma}}^j \odot \boldsymbol{\epsilon},
\qquad
\boldsymbol{\epsilon} \sim \mathcal{N}(\mathbf{0}, \mathbf{I}).
\end{equation}
This formulation preserves differentiability, allowing gradients to propagate through the stochastic sampling process during meta-training. During training, we additionally regularize $\mathcal{Q}_{\phi_j}\!\left(\mathbf{Z}^j \mid \mathcal{M}^{j}\right)$ toward a standard normal reference to stabilize the latent space.

The objective of the context encoder is to approximate the true posterior over tasks, thereby mapping observed operational outcomes to a latent variable that encapsulates the implicit constraints induced by the charging infrastructure layout. This probabilistic inference architecture offers two fundamental advantages. First, it achieves task disentanglement. By separating task-specific structural constraints (captured by $\tilde{\mathbf{z}}^j$) from general control dynamics (learned by the policy), the network can learn transferable dispatching behaviors that are modulated by the inferred context. Second, it enables rapid online adaptation. After meta-training, the encoder functions as a learned inference rule that allows the agent to adapt to previously unseen charging layouts without requiring gradient updates. By simply constructing a context set $\mathcal{M}^{j}$ from the context buffer, the agent immediately infers the corresponding embedding $\tilde{\mathbf{z}}^j$, adjusting its policy to operate effectively in the new environment.

\noindent \textbf{Actor-Critic networks.}
The decision making process of each Area Agent is governed by an actor network $\theta_j^{a}$ that conditions its decisions on both the spatially contextualized system state and the inferred task context. Specifically, the actor receives two complementary inputs: the spatial embedding $\tilde{\mathbf{h}}_{t}^{a,j}$ produced by GAT (capturing explicit topological relations) and the latent task representation $\tilde{\mathbf{z}}^j$ inferred by the context encoder (capturing implicit infrastructure constraints). These inputs are fused to form a unified policy input representation:
\begin{equation}
\mathbf{b}_t^{a,j}
    =
    \left[
    \tilde{\mathbf{h}}_{t}^{a,j}
    \;\parallel\;
    \tilde{\mathbf{z}}^j
    \right].
\end{equation}
This joint embedding is critical for meta adaptation, as it allows the policy to modulate its region-level control logic based on the underlying charging infrastructure layout while retaining transferable behaviors across different environments.

The fused representation $\mathbf{b}_t^{a,j}$ is processed by a multilayer perceptron to parameterize a stochastic policy with bounded outputs. Concretely, the actor network parameterizes a diagonal Gaussian policy by outputting the mean vector and log standard deviation vector,
\begin{equation}
\boldsymbol{\mu}_{\theta_j^{a}}(\mathbf{b}_t^{a,j}), \;
    \log \boldsymbol{\sigma}_{\theta_j^{a}}(\mathbf{b}_t^{a,j})
    =
    \mathrm{MLP}_{\theta_j^{a}}(\mathbf{b}_t^{a,j}).
\end{equation}
Actions are generated using the reparameterization trick to preserve differentiability. A raw action sample is first constructed as
\begin{equation}
\hat{\mathbf{a}}_t^{a,j}
    =
    \boldsymbol{\mu}_{\theta_j^{a}}(\mathbf{b}_t^{a,j})
    +
    \boldsymbol{\sigma}_{\theta_j^{a}}(\mathbf{b}_t^{a,j})
    \odot \boldsymbol{\epsilon},
    \qquad
    \boldsymbol{\epsilon} \sim \mathcal{N}(\mathbf{0}, \mathbf{I}),
\end{equation}
and subsequently squashed by a hyperbolic tangent transformation to ensure feasibility:
\begin{equation}
    \mathbf{A}_t^{a,j} = \tanh(\hat{\mathbf{a}}_t^{a,j}).
\end{equation}
The resulting bounded action vector directly corresponds to the operational threshold parameters defined in the Area Agent CMDP formulation, controlling order matching, repositioning, and charging thresholds under different SOC phases.

Action evaluation is performed by a pair of critic networks $Q_{\psi_{j,1}^{a}}$ and $Q_{\psi_{j,2}^{a}}$, which estimate the expected soft value of the task conditioned policy. Each critic takes as input the complete context state pair and the selected action:
\begin{equation}
    Q_{\psi_{j,m}^{a}}
    \!\left(
    \mathbf{h}_{t}^{a,j},
    \mathbf{A}_t^{a,j},
    \tilde{\mathbf{z}}^j
    \right)
    =
    \mathrm{MLP}_{\psi_{j,m}^{a}}
    \!\left(
    \left[
    \mathbf{h}_{t}^{a,j}
    \;\parallel\;
    \mathbf{A}_t^{a,j}
    \;\parallel\;
    \tilde{\mathbf{z}}^j
    \right]
    \right),
    \qquad m \in \{1,2\}.
\end{equation}
By explicitly conditioning the value estimation on the inferred task representation $\tilde{\mathbf{z}}^j$, the critics can accurately assess the long term return of an action within the specific constraints of the current infrastructure configuration. During training, target networks and the minimum Q-value strategy are employed, consistent with SAC, to stabilize learning and mitigate overestimation bias.

\subsection{Execution Pipeline}\label{subsec_pipeline}

\begin{figure}[!htbp]
    \centering
    \includegraphics[width=0.9\linewidth]{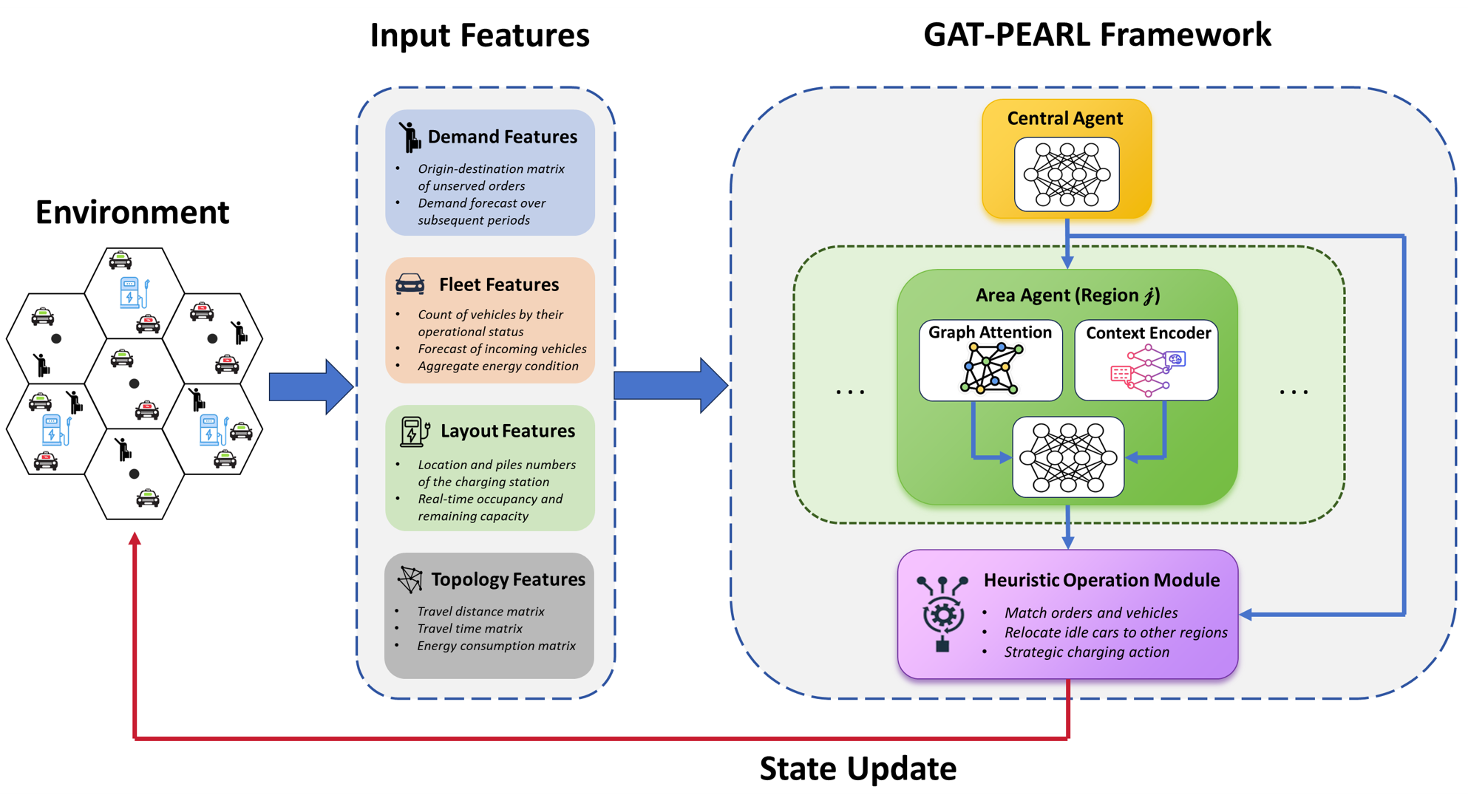}
    \caption{The execution pipeline of the hierarchical GAT-PEARL framework.}
    \label{fig:pipeline}
\end{figure}

This subsection summarizes the end-to-end forward propagation and execution pipeline of the proposed hierarchical GAT-PEARL framework. Figure~\ref{fig:pipeline} shows how environment observations are transformed into strategic and tactical control signals, how these signals drive real-time operations through the heuristic module, and how the system state is updated for the next decision period. After training, in deployment, a single forward pass through this pipeline suffices to produce an operational policy that adapts to a new infrastructure layout.

At the beginning of each period, the environment provides a global snapshot of system conditions over the spatial network. We organize the observation into four feature groups: demand features that describe outstanding and anticipated requests, fleet features that summarize regional vehicle availability and energy conditions, layout features that reflect charging station availability and real-time capacity, and topology features that capture travel time and energy relationships across regions.

The Central Agent processes the aggregated global features to produce high-level coordination signals. These signals do not specify vehicle-level moves. Instead, they provide system-wide guidance that regulates rebalancing intensity, charging usage targets, and economic incentives for serving different regions. The Central Agent then broadcasts these signals to all Area Agents. 

Each Area Agent combines three types of inputs: its local operational observations, the strategic signals from the Central Agent, and the local structural information induced by the current infrastructure layout and network connectivity, together with a time index encoding. The Area Agent then performs two coupled inference steps. First, it applies a graph attention encoder to incorporate neighborhood information and capture spatial interactions. Second, it applies a context encoder to infer a latent representation of the current infrastructure-induced dynamics from recent interaction history. Conditioned on both spatial encoding and inferred context, the Area Agent outputs a tactical control vector that parameterizes regional matching, repositioning, and charging preferences. 

The heuristic operation module serves as the execution layer. It takes the strategic guidance from the Central Agent and the tactical parameters from all Area Agents, and then performs real-time decisions within operational intervals. Specifically, it matches orders to feasible vehicles, reallocates idle vehicles across regions when necessary, dispatches vehicles to charging stations subject to infrastructure feasibility and capacity, and applies the incentive signals when computing realized regional profits. After interval-level execution finishes, the environment updates the fleet distribution, charging status, and demand backlog, producing the next-period observations and rewards.

\section{Algorithms} \label{sec_algorithms}

This section outlines the algorithmic pipeline of our GAT-PEARL framework, covering how policies are trained, validated, adapted, and executed under evolving charging infrastructure. We first present the Area Agent meta-training procedure over a task distribution of charging layouts, which learns both a transferable initialization and a context-conditioned inference mechanism. We then incorporate an online meta-validation step to monitor generalization on validation tasks during training. After convergence, we describe the few-shot adaptation procedure on the held-out test set, followed by a deterministic heuristic execution module that converts high-level agent directives into feasible dispatching actions subject to operational constraints.

\subsection{Meta-Training Algorithm}\label{subsec_Area_Agent_Meta}

\begin{figure}[!htbp]
    \centering
    \includegraphics[width=1\linewidth]{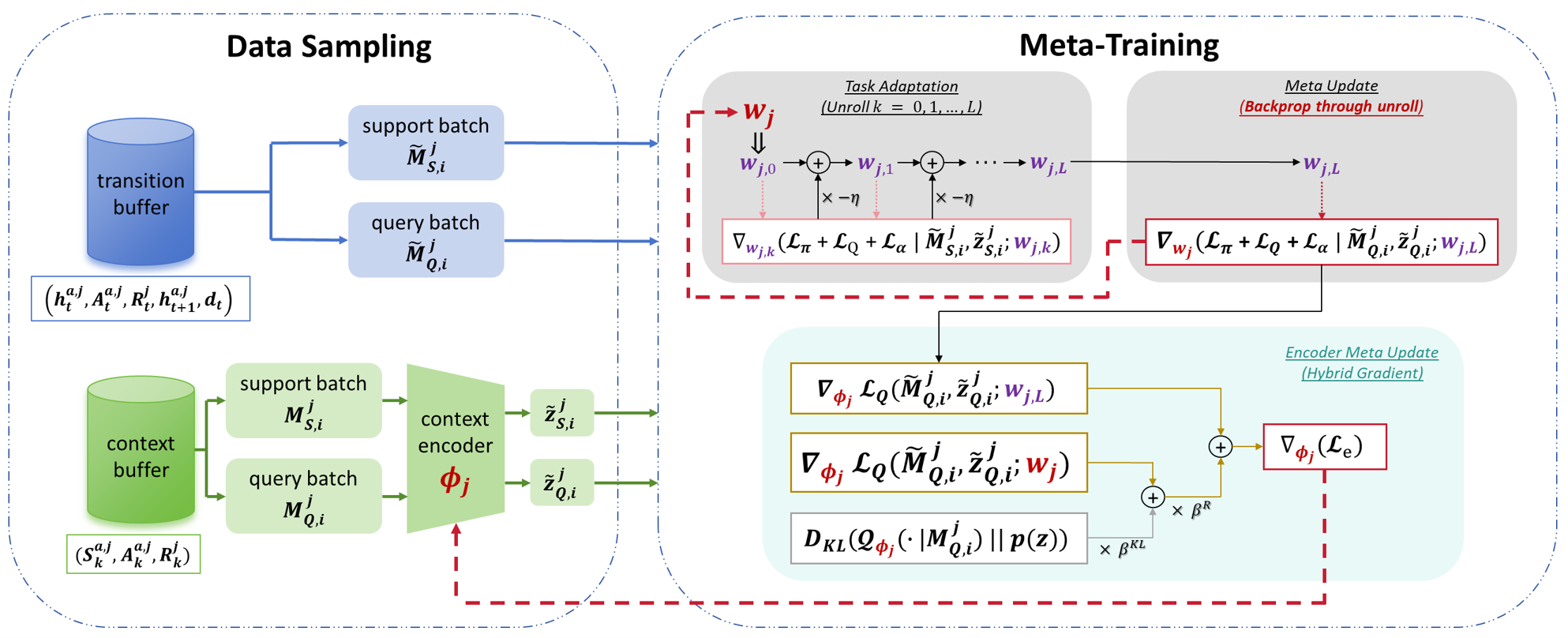}
    \caption{The process of meta-training.}
    \label{fig:meta-training}
\end{figure}

The primary objective of the Area Agent training is to acquire a generalizable control policy capable of adapting to heterogeneous infrastructure layouts $C$ without extensive retraining. We formulate this objective as a meta-learning problem over a task distribution $p(C)$, where each task corresponds to a distinct charging station layout. The training process is governed by a set of meta-parameters $\boldsymbol{\Lambda} = (\phi_j, \theta_j^a, \psi_j^a, \alpha^a)$, which fulfill two complementary roles. The context encoder parameters $\phi_j$ constitute a shared inference mechanism designed to extract latent structural information from the interaction history. Conversely, the actor-critic parameters $(\theta_j^a, \psi_j^a, \alpha^a)$ serve as a global initialization shared across all tasks. Unlike standard reinforcement learning, which directly optimizes the policy for immediate deployment, our approach optimizes these initial parameters such that a few steps of gradient descent produce a task-specific policy with maximal performance. This is achieved through a meta-training framework augmented with a hybrid gradient strategy. By aggregating gradients from both the adaptation process and the representation learning process, this strategy effectively balances the plasticity required for fast adaptation with the stability needed for consistent representation learning. In this section, we first define the component objectives, then detail the meta-training procedure for policy initialization, and finally describe the hybrid update rule for the context encoder. The meta-training process is shown in Figure~\ref{fig:meta-training}.

\subsubsection{Component Loss Functions}

For each Area Agent $j$, the training involves data sampled from two distinct replay buffers: a \textit{transition buffer} storing full trajectories for actor-critic updates, and a \textit{context buffer} storing interaction tuples for latent inference.

In each training step, we first sample a context batch $\mathcal{M}^{j} = \left\{(\mathbf{S}_{k}^{a,j},\mathbf{A}_{k}^{a,j},R_{k}^{j})\right\}_{k=1}^{N_c}$ from the context buffer containing context tuples to infer the latent task embedding. To enable gradient propagation through the sampling process, we apply the reparameterization trick, yielding a sampled latent variable $\tilde{\mathbf{z}}^j \sim \mathcal{Q}_{\phi_j}(\mathbf{Z}^j \mid \mathcal{M}^{j})$. Subsequently, we sample a transition batch $\tilde{\mathcal{M}}^{j}$ from the transition buffer containing tuples $(\mathbf{h}_{t}^{a,j}, \mathbf{A}_t^{a,j}, R_t^j, \mathbf{h}_{t+1}^{a,j}, d_t)$ to evaluate the actor and critic losses. The specific objective functions are defined as follows:

\noindent \textbf{Critic loss.} Value estimation is performed using two critic networks parameterized by $\psi_{j,1}^{a}$ and $\psi_{j,2}^{a}$ to mitigate overestimation bias. Given the transition batch $\tilde{\mathcal{M}}^{j}$ and the sampled latent context $\tilde{\mathbf{z}}^j$, each critic minimizes the soft Bellman residual:
\begin{equation}
    \mathcal{L}_Q(\tilde{\mathcal{M}}^{j}, \tilde{\mathbf{z}}^j; \psi_j^a) = \mathbb{E}_{(\mathbf{h}_{t}^{a,j}, \mathbf{A}_t^{a,j}, R_{k}^{j}, \mathbf{h}_{t+1}^{a,j}) \sim \tilde{\mathcal{M}}^{j}} \left[ \frac{1}{2} \left( Q_{\psi_{j,m}^{a}}(\mathbf{h}_{t}^{a,j}, \mathbf{A}_t^{a,j}, \tilde{\mathbf{z}}^j) - y_t^j \right)^2 \right], \quad m \in \{1,2\},
\end{equation}
where the target $y_t^j$ incorporates entropy regularization. The next-state action $\tilde{\mathbf{A}}_{t+1}^{a,j}$ used in the target calculation is sampled from the current policy $\pi_{\theta_j^a}(\cdot|\mathbf{h}_{t+1}^{a,j}, \tilde{\mathbf{z}}^j)$. The target value is computed as:
\begin{equation}
    y_t^j = R_t^j+ \gamma^a (1-d_t) \left( \min_{m=1,2} Q_{\bar{\psi}_{j,m}^{a}}(\mathbf{h}_{t+1}^{a,j}, \tilde{\mathbf{A}}_{t+1}^{a,j}, \tilde{\mathbf{z}}^j) - \alpha_j^a \log \pi_{\theta_j^a}(\tilde{\mathbf{A}}_{t+1}^{a,j}|\mathbf{h}_{t+1}^{a,j}, \tilde{\mathbf{z}}^j) \right).
\end{equation}
Crucially, $\tilde{\mathbf{z}}^j$ is treated as a fixed input during the critic update to decouple representation learning from value estimation errors.

\noindent \textbf{Actor loss.} The actor network $\theta_j^a$ is optimized to maximize the expected soft value conditioned on the inferred context. We employ the reparameterization trick to sample actions, defined as $\tilde{\mathbf{A}}_t^{a,j} = \tanh(\mu_{\theta_j^a}(\cdot) + \sigma_{\theta_j^a}(\cdot) \odot \epsilon)$, where $\epsilon \sim \mathcal{N}(0, I)$. To prevent the policy from exploiting the latent space dynamics, $\tilde{\mathbf{z}}^j$ is detached from the computation graph during this update. The loss is defined as:
\begin{equation}
    \mathcal{L}_\pi(\tilde{\mathcal{M}}^{j}, \tilde{\mathbf{z}}^j; \theta_j^a) = \mathbb{E}_{\mathbf{h}_{t}^{a,j} \sim \tilde{\mathcal{M}}^{j}} \left[ \alpha_j^a \log \pi_{\theta_j^a}(\tilde{\mathbf{A}}_t^{a,j}|\mathbf{h}_{t}^{a,j}, \tilde{\mathbf{z}}^j) - \min_{m=1,2} Q_{\psi_{j,m}^{a}}(\mathbf{h}_{t}^{a,j}, \tilde{\mathbf{A}}_t^{a,j}, \tilde{\mathbf{z}}^j) \right].
\end{equation}

\noindent \textbf{Entropy coefficient loss.} The temperature parameter $\alpha_j^a$ is automatically tuned to maintain the policy entropy close to a target level $\bar{\mathcal{H}}^{a}$, ensuring a consistent balance between exploration and exploitation. The objective is to minimize:
\begin{equation}
    \mathcal{L}_\alpha(\tilde{\mathcal{M}}^{j}, \tilde{\mathbf{z}}^j; \alpha_j^a) = \mathbb{E}_{\mathbf{h}_{t}^{a,j} \sim \tilde{\mathcal{M}}^{j}} \left[ -\alpha_j^a \left( \log \pi_{\theta_j^a}(\tilde{\mathbf{A}}_t^{a,j}|\mathbf{h}_{t}^{a,j}, \tilde{\mathbf{z}}^j) + \bar{\mathcal{H}}^{a} \right) \right].
\end{equation}

\subsubsection{Meta Update for Actor-Critic}

In each training iteration, we sample a batch of tasks $\{C_i\}$, where each task corresponds to a distinct infrastructure layout. For every sampled task $i$, the agents first interact with the specific environment using the proposed framework to generate operational trajectories. The resulting transitions and context data are accumulated into the corresponding transitions buffer and context buffer of agent $j$. Subsequently, we draw four distinct batches for training procedure: a support context batch $\mathcal{M}^{j}_{S,i}$, a support transition batch $\tilde{\mathcal{M}}^{j}_{S,i}$, a query context batch $\mathcal{M}^{j}_{Q,i}$, and a query transition batch $\tilde{\mathcal{M}}^{j}_{Q,i}$. We update the actor and critic initialization parameters $\mathbf{w}_j = (\theta_j^a, \psi_j^a, \alpha^a)$ for each Area Agent $j$ such that applying a few steps of gradient descent starting from these values leads to rapid convergence on any specific task. This is achieved through a task adaptation phase and a meta update phase.

During the task adaptation phase, the primary objective is to construct a differentiable optimization trajectory tailored to the specific infrastructure layout. To establish this dependency, we instantiate a differentiable clone of the actor-critic parameters, initialized as $\mathbf{w}_{j,0} = \mathbf{w}_j$, which we refer to as the \textit{fast agent}. Crucially, the context encoder parameters $\phi_j$ are held fixed throughout this phase, serving strictly as a static inference module. Over the course of $L$ adaptation steps, we explicitly unroll the computational graph by performing sequential gradient updates on the parameter clone. Conditioned on the latent embedding $\tilde{\mathbf{z}}^j_{S,i}$ inferred from the support context batch, the process iteratively refines the policy. For each step $k=0, \dots, L-1$, the process proceeds as follows. First, the frozen encoder infers the task-specific latent embedding $\tilde{\mathbf{z}}^j_{S,i}$ from the support context batch. Subsequently, we evaluate the independent component objectives, specifically the critic loss $\mathcal{L}_Q$, actor loss $\mathcal{L}_\pi$, and entropy loss $\mathcal{L}_\alpha$, on the support transition batch. The parameter state $\mathbf{w}_{j,k}$ is then updated by applying gradient descent with component specific step sizes:
\begin{equation}
\begin{aligned}
\theta_{j,k+1}^{a}
&= \theta_{j,k}^{a}
- \eta_{\pi}^a\,\nabla_{\theta}\,
\mathcal{L}_{\pi}\!\left(\tilde{\mathcal{M}}^{j}_{S,i}, \tilde{\mathbf{z}}^{j}_{S,i}; \theta_{j,k}^{a}\right),\\
\psi_{j,k+1}^{a}
&= \psi_{j,k}^{a}
- \eta_{Q}^a\,\nabla_{\psi}\,
\mathcal{L}_{Q}\!\left(\tilde{\mathcal{M}}^{j}_{S,i}, \tilde{\mathbf{z}}^{j}_{S,i}; \psi_{j,k}^{a}\right),\\
\alpha_{j,k+1}^{a}
&= \alpha_{j,k}^{a}
- \eta_{\alpha}^a\,\nabla_{\alpha}\,
\mathcal{L}_{\alpha}\!\left(\tilde{\mathcal{M}}^{j}_{S,i}, \tilde{\mathbf{z}}^{j}_{S,i}; \alpha_{j,k}^{a}\right).
\end{aligned}
\end{equation}
Throughout this iterative process, the computational graph retains the complete history of operations connecting the initial parameters $\mathbf{w}_j$ to the current state. This mechanism effectively encodes the final adapted parameters $\mathbf{w}_{j,L}$ as a complex, differentiable function of the initialization $\mathbf{w}_j$, thereby enabling the preservation of higher order gradient information required for the subsequent meta update.

In the meta update phase, we optimize the meta-parameters $\mathbf{w}_j$ (representing the initial actor and critic configurations) to minimize the error on the query set after adaptation. First, a query-specific latent embedding $\tilde{\mathbf{z}}^j_{Q,i}$ is inferred from the query context batch. We then evaluate the generalization performance of the final adapted parameters $\mathbf{w}_{j,L}$ by computing the actor loss $\mathcal{L}_\pi$ and critic loss $\mathcal{L}_Q$ on the query transition batch $\tilde{\mathcal{M}}^{j}_{Q,i}$. Since the entire adaptation process is explicitly differentiated, the adapted parameter state $\mathbf{w}_{j,L}$ functions as a differentiable transformation of the initialization $\mathbf{w}_{j,0}$. Consequently, we can compute the meta-gradient with respect to the initialization by backpropagating the query loss through the unrolled computational graph of the $L$ adaptation steps. This effectively transfers the supervision signal from the final adapted model back to the initial configuration. The meta-parameters are then updated as follows:
\begin{equation}
\begin{aligned}
\theta_{j}^{a}
&\leftarrow \theta_{j}^{a}
- \eta_{\pi}^a\,
\nabla_{\theta_{j}^{a}}\,
\mathbb{E}_{C_i \sim p(C)}\!\Big[
\mathcal{L}_{\pi}\!\left(
\tilde{\mathcal{M}}^{j}_{Q,i},\, \tilde{\mathbf{z}}^{j}_{Q,i};\, \theta_{j,L}^{a}
\right)
\Big],\\
\psi_{j}^{a}
&\leftarrow \psi_{j}^{a}
- \eta_{Q}^a\,
\nabla_{\psi_{j}^{a}}\,
\mathbb{E}_{C_i \sim p(C)}\!\Big[
\mathcal{L}_{Q}\!\left(
\tilde{\mathcal{M}}^{j}_{Q,i},\, \tilde{\mathbf{z}}^{j}_{Q,i};\, \psi_{j,L}^{a}
\right)
\Big],\\
\alpha_{j}^{a}
&\leftarrow \alpha_{j}^{a}
- \eta_{\alpha}^a\,
\nabla_{\alpha_{j}^{a}}\,
\mathbb{E}_{C_i \sim p(C)}\!\Big[
\mathcal{L}_{\alpha}\!\left(
\tilde{\mathcal{M}}^{j}_{Q,i},\, \tilde{\mathbf{z}}^{j}_{Q,i};\, \alpha_{j,L}^{a}
\right)
\Big].
\end{aligned}
\end{equation}

In summary, the task adaptation and meta update phases effectively harmonize inference-based context recognition with gradient-based parameter adaptation. By explicitly optimizing the initialization for rapid fine tuning, the framework ensures that the agent can not only identify the underlying infrastructure topology via the latent context but also swiftly adjust its control policy to achieve optimal performance in previously unseen environments.

\subsubsection{Meta Update for Context Encoder using Hybrid Gradient}

While the policy parameters are updated via standard meta-gradients, the context encoder $\phi_j$ requires a more robust update strategy to balance representation quality with control efficacy. To achieve this, we compute gradients for $\phi_j$ by aggregating signals from two distinct computational pathways: the adaptation path (traversing the unrolled optimization of the fast agent) and the representation path (directly derived from the base agent).

The total gradient applied to the encoder is formulated as a weighted combination of the adaptation objective $\mathcal{L}_{\rho}$ and the auxiliary representation objective $\mathcal{L}_{r}$:
\begin{equation}
\begin{aligned}
\nabla_{\phi_j}\,\mathcal{L}_{e}
=\;
\underbrace{
\nabla_{\phi_j}\,
\mathcal{L}_{Q}\!\left(\mathbf{w}_{j,L}\right)
}_{\text{Adaptation Gradients (via Fast Agent)}}
+
\beta^{\mathrm{R}}\,
\underbrace{
\Bigg(
\nabla_{\phi_j}\,
\mathcal{L}_{Q}\!\left(\mathbf{w}_{j}\right)
+
\beta^{\mathrm{KL}}\,
\nabla_{\phi_j}\,
D_{\mathrm{KL}}\!\left(
\mathcal{Q}_{\phi_j}(\cdot \mid \mathcal{M}^{j}_{Q,i})
\,\big\|\,
p(\mathbf{z})
\right)
\Bigg)
}_{\text{Representation Gradients (via Base Agent)}} .
\end{aligned}
\end{equation}
Here, $\mathbf{w}_{j,L}$ denotes the adapted parameters of the fast agent after the task adaptation unrolling, whereas $\mathbf{w}_{j}$ denotes the primitive initialization of the base agent. Although these terms share mathematical similarities, they enforce fundamentally different behaviors on the latent space:

\begin{itemize}

    \item \textbf{Adaptation gradients $\mathcal{L}_{\rho}$:} This component backpropagates through the unrolled computational graph of the task adaptation phase. It treats the latent embedding $\tilde{\mathbf{z}}^j$ as a preconditioner for optimization, explicitly instructing the encoder to generate embeddings that modify the loss landscape such that gradient descent leads to maximal performance gain. This mechanism promotes plasticity, ensuring that the generated context effectively accelerates the adaptation process.
    \item \textbf{Representation gradients $\mathcal{L}_{r}$:} This component provides a direct and stable learning signal for the encoder by optimizing it with respect to the base agent, without backpropagating through the unrolled adaptation dynamics. It encourages the inferred latent embedding $\tilde{\mathbf{z}}^j$ to be predictively useful for value estimation under the pre-adaptation critic, thereby improving the identifiability and consistency of the task representation.
\end{itemize}

Furthermore, within these gradient components, the critic loss and KL divergence play complementary roles. The critic based terms $\mathcal{L}_{Q}$ provide the informational drive, pushing the encoder to extract discriminative features relevant for value prediction. Conversely, the KL divergence terms act as a structural regularizer, constraining the latent space to a smooth manifold close to the prior $p(\mathbf{z})$ to ensure generalization. 

By integrating these diverse signals, the final update rule is given by: 
\begin{equation} 
\phi_j \leftarrow \phi_j - \eta_{e} \nabla_{\phi_j} \mathcal{L}_{e}, 
\end{equation} 
ensures that the learned latent space is both physically grounded (via the base agent signals) and adaptation friendly (via the fast agent signals), effectively resolving the trade off between representation stability and adaptive plasticity. Algorithm \ref{alg:meta_learning} shows the corresponding pseudocode for the Area Agent's meta-training process.

\begin{algorithm}[!t]
\caption{Meta-Training for Area Agent $j$}
\label{alg:meta_learning}
\DontPrintSemicolon
\SetAlgoLined

\KwIn{
Task distribution $p(\mathcal{C})$; 
learning rate $\eta_{\pi}^{a},\,\eta_{Q}^{a},\,\eta_{\alpha}^{a}, \eta_{e}$; 
representation weight $\beta^{\mathrm{R}}$; KL penalty weight $\beta^{KL}$.
}
\KwOutput{Updated Context encoder parameters $\phi_j$ and actor-critic parameters $\mathbf{w}_j = (\theta_j^a, \psi_j^a, \alpha_j^a)$.}

\KwInit{
Context encoder $\phi_j$ and actor-critic parameters $\mathbf{w}_j$;
transition buffer and context buffer.
}

\BlankLine

\While{not converged}{
    Sample a batch of tasks $\{C_i\} \sim p(C)$.\;

    Initialize cumulative gradients $\nabla_{\boldsymbol{\Lambda}} \mathcal{L}_{e} \leftarrow 0$.\;

    \ForEach{task $C_i$}{
        \tcp{Sample batches for meta-learning}
        Sample support context batch $\mathcal{M}^{j}_{S,i}$ and support transition batch $\tilde{\mathcal{M}}^{j}_{S,i}$.\;
        Sample query context batch $\mathcal{M}^{j}_{Q,i}$ and query transition batch $\tilde{\mathcal{M}}^{j}_{Q,i}$.\;

        \tcp{Infer support latent embedding}
        $\tilde{\mathbf{z}}^{j}_{S,i} \sim \mathcal{Q}_{\phi_j}(\cdot \mid \mathcal{M}^{j}_{S,i})$.\;

        \tcp{Task Adaptation (Fast Agent)}
        Initialize fast agent parameters: $\mathbf{w}_{j,0} \leftarrow \mathbf{w}_j$.\;

        \For{$k = 0$ \KwTo $L-1$}{
            Compute losses on support transitions:\;
            $\theta_{j,k+1}^{a}
            = \theta_{j,k}^{a}
            - \eta_{\pi}^a\,\nabla_{\theta}\,
            \mathcal{L}_{\pi}\!\left(\tilde{\mathcal{M}}^{j}_{S,i}, \tilde{\mathbf{z}}^{j}_{S,i}; \theta_{j,k}^{a}\right),$;\;
            $\psi_{j,k+1}^{a}
            = \psi_{j,k}^{a}
            - \eta_{Q}^a\,\nabla_{\psi}\,
            \mathcal{L}_{Q}\!\left(\tilde{\mathcal{M}}^{j}_{S,i}, \tilde{\mathbf{z}}^{j}_{S,i}; \psi_{j,k}^{a}\right),$;\;
            $\alpha_{j,k+1}^{a}
            = \alpha_{j,k}^{a}
            - \eta_{\alpha}^a\,\nabla_{\alpha}\,
            \mathcal{L}_{\alpha}\!\left(\tilde{\mathcal{M}}^{j}_{S,i}, \tilde{\mathbf{z}}^{j}_{S,i}; \alpha_{j,k}^{a}\right)$.\;

            Update fast agent (unroll the computation graph):\;
            $\theta_{j,k+1}^{a}
            = \theta_{j,k}^{a}
            - \eta_{\pi}^a\,\nabla_{\theta}\,
            \mathcal{L}_{\pi}\!\left(\tilde{\mathcal{M}}^{j}_{S,i}, \tilde{\mathbf{z}}^{j}_{S,i}; \theta_{j,k}^{a}\right)$;\;
            $\psi_{j,k+1}^{a}
            = \psi_{j,k}^{a}
            - \eta_{Q}^a\,\nabla_{\psi}\,
            \mathcal{L}_{Q}\!\left(\tilde{\mathcal{M}}^{j}_{S,i}, \tilde{\mathbf{z}}^{j}_{S,i}; \psi_{j,k}^{a}\right)$;\;
            $\alpha_{j,k+1}^{a}
            = \alpha_{j,k}^{a}
            - \eta_{\alpha}^a\,\nabla_{\alpha}\,
            \mathcal{L}_{\alpha}\!\left(\tilde{\mathcal{M}}^{j}_{S,i}, \tilde{\mathbf{z}}^{j}_{S,i}; \alpha_{j,k}^{a}\right)$.\;
            }
        Denote adapted fast agent parameters as $\mathbf{w}_{j,L}$.\;

        \tcp{Meta Update}
        Infer query latent embedding:\;
        $\tilde{\mathbf{z}}^{j}_{Q,i} \sim \mathcal{Q}_{\phi_j}(\cdot \mid \mathcal{M}^{j}_{Q,i})$.\;

        \tcc{(A) Meta update for Actor-Critic}
        $\theta_{j}^{a}
        \leftarrow \theta_{j}^{a}
        - \eta_{\pi}^a\,
        \nabla_{\theta_{j}^{a}}\,
        \mathbb{E}_{C_i \sim p(C)}\!\Big[
        \mathcal{L}_{\pi}\!\left(
        \tilde{\mathcal{M}}^{j}_{Q,i},\, \tilde{\mathbf{z}}^{j}_{Q,i};\, \theta_{j,L}^{a}
        \right)
        \Big]$;\;
        $\psi_{j}^{a}
        \leftarrow \psi_{j}^{a}
        - \eta_{Q}^a\,
        \nabla_{\psi_{j}^{a}}\,
        \mathbb{E}_{C_i \sim p(C)}\!\Big[
        \mathcal{L}_{Q}\!\left(
        \tilde{\mathcal{M}}^{j}_{Q,i},\, \tilde{\mathbf{z}}^{j}_{Q,i};\, \psi_{j,L}^{a}
        \right)
        \Big]$.\;

        \tcc{(B) Meta update for Encoder using Hybrid gradient}
        $\nabla_{\phi_j}\,\mathcal{L}_{e}
        =\;
        \nabla_{\phi_j}\,
        \mathcal{L}_{Q}\!\left(\mathbf{w}_{j,L}\right)
        +
        \beta^{\mathrm{R}}\,
        \Bigg(
        \nabla_{\phi_j}\,
        \mathcal{L}_{Q}\!\left(\mathbf{w}_{j}\right)
        +
        \beta^{\mathrm{KL}}\,
        \nabla_{\phi_j}\,
        D_{\mathrm{KL}}\!\left(
        \mathcal{Q}_{\phi_j}(\cdot \mid \mathcal{M}^{j}_{Q,i})
        \,\big\|\,
        p(\mathbf{z})
        \right)
        \Bigg) $.
    }
}
\end{algorithm}

\subsubsection{Meta-Validation}

Meta-validation evaluates the generalization capability of the meta-trained agents on infrastructure layouts that are strictly disjoint from those used during meta-training. Specifically, the validation task set contains layouts whose intersection with the training task set is empty. At the end of each meta-training epoch, we sample a batch of validation tasks $\{C_i\}$ from the validation set and freeze all network parameters (including the actor, critic, and context encoder). The agents are then evaluated on each unseen task without any gradient-based updates, ensuring that the measured performance reflects pure generalization rather than additional learning.

Since the context encoder requires task-specific evidence to infer the latent embedding, meta-validation begins with a lightweight interaction phase to populate the context buffer of the new task. Concretely, for each validation layout, the agent interacts with the environment for several steps using the meta-trained initialization and stores the resulting context batch $\mathcal{M}_{v}^j$ into the task-specific context buffer. The fixed encoder then produces a deterministic task embedding by taking the posterior mean:
\begin{equation}
    \mathbf{z}^{j}_{v} \leftarrow \mathbb{E}\!\left[\mathcal{Q}_{\phi_j}\!\left(\mathbf{Z}^j \mid \mathcal{M}_{v}^j\right)\right],
\end{equation}
rather than sampling from the posterior. This deterministic choice eliminates evaluation noise introduced by posterior variance and isolates the quality of the inferred task representation.

To further ensure a deterministic evaluation, both the actor and critic also operate in mean mode. That is, the actor outputs the mean action $\mu_{\theta}(\cdot)$ without stochastic sampling, and the critic is queried consistently with these mean actions. For each validation task, we perform three forward passes under the same buffered context and evaluate the resulting performance. This procedure provides a stable estimate of meta-generalization and serves as a diagnostic signal for hyperparameter selection during meta-training.

\subsection{Central Agent Training}

\begin{figure}[!htbp]
    \centering
    \includegraphics[width=0.7\linewidth]{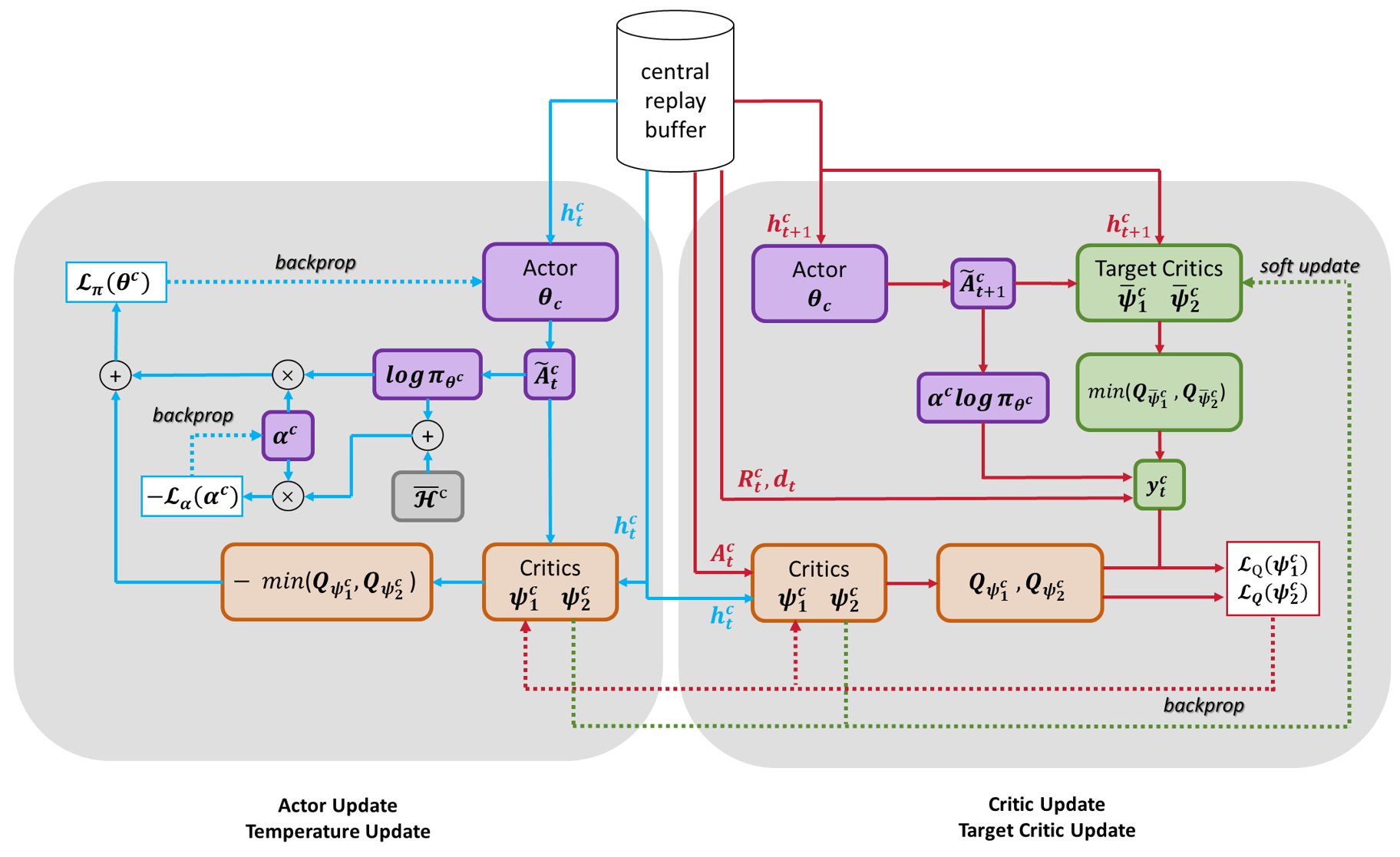} 
    \caption{The training process of central agent.}
    \label{fig:training_central_agent}
\end{figure}

The Central Agent operates as the strategic coordinator, optimizing a global objective through a continuous control policy. We employ the SAC algorithm, a maximum entropy reinforcement learning framework that optimizes a stochastic policy for both substantial expected return and high entropy. This dual objective encourages robust exploration and prevents the policy from collapsing into suboptimal deterministic behaviors. Figure~\ref{fig:training_central_agent}illustrates the training process of the central agent.

To estimate the value of the current policy, we utilize a dual critic mechanism parameterized by $\psi_1^{c}$ and $\psi_2^{c}$. At each time step $t$, both critics process the Central Agent feature representation $\mathbf{h}_t^{c}$ and the normalized action $\mathbf{A}_t^{c}$ to output scalar soft action values:
\begin{equation}
    Q_{\psi_m^{c}}(\mathbf{h}_t^{c}, \mathbf{A}_t^{c}) \in \mathbb{R}, \quad m \in \{1,2\}.
\end{equation}
For a transition tuple $(\mathbf{h}_t^{c}, \mathbf{A}_t^{c}, R_t^{c}, \mathbf{h}_{t+1}^{c}, d_t)$ sampled from the central agent's replay buffer, we construct a soft Bellman target. This involves sampling a next-state action $\tilde{\mathbf{A}}_{t+1}^{c} \sim \pi_{\theta^{c}}(\cdot \mid \mathbf{h}_{t+1}^{c})$ from the current policy and computing the target value as:
\begin{equation}
    y_t^c = R_t^{c} + \gamma^c (1 - d_t) \left( \min_{m \in \{1,2\}} Q_{\bar{\psi}_m^{c}}(\mathbf{h}_{t+1}^{c}, \tilde{\mathbf{A}}_{t+1}^{c}) - \alpha^c \log \pi_{\theta^{c}}(\tilde{\mathbf{A}}_{t+1}^{c} \mid \mathbf{h}_{t+1}^{c}) \right).
\end{equation}
Here, the target incorporates an entropy bonus weighted by the temperature parameter $\alpha^c$, and $\bar{\psi}_m^{c}$ denotes the parameters of the target critic networks. The critic networks are updated by minimizing the mean squared Bellman error:
\begin{equation}
    \mathcal{L}_Q(\psi_m^{c}) = \mathbb{E}_{(\mathbf{h}_t^{c}, \mathbf{A}_t^{c}, R_t^{c}, \mathbf{h}_{t+1}^{c}, d_t)} \left[ \left( Q_{\psi_m^{c}}(\mathbf{h}_t^{c}, \mathbf{A}_t^{c}) - y_t^c \right)^2 \right], \quad m \in \{1,2\}.
\end{equation}
The utilization of two critics alongside the minimum operator in the target calculation serves to mitigate the overestimation bias inherent in value estimation.

The actor network, parameterized by $\theta^{c}$, aims to maximize the expected soft value. We apply the reparameterization trick to sample actions $\tilde{\mathbf{A}}_t^{c} \sim \pi_{\theta^{c}}(\cdot \mid \mathbf{h}_t^{c})$. The policy objective is defined to minimize the following loss (equivalent to maximizing the soft value):
\begin{equation}
    \mathcal{L}_{\pi}(\theta^{c}) = \mathbb{E}_{\mathbf{h}_t^{c}} \left[ \alpha^c \log \pi_{\theta^{c}}(\tilde{\mathbf{A}}_t^{c} \mid \mathbf{h}_t^{c}) - \min_{m \in \{1,2\}} Q_{\psi_m^{c}}(\mathbf{h}_t^{c}, \tilde{\mathbf{A}}_t^{c}) \right].
\end{equation}

To ensure the agent maintains an appropriate magnitude of exploration throughout training, we employ an automated entropy adjustment mechanism. The temperature $\alpha^c$ is treated as a learnable variable and optimized to align the policy entropy with a predefined target level $\bar{\mathcal{H}}^{c}$. The objective for $\alpha^c$ is given by:
\begin{equation}
    \mathcal{L}_{\alpha}(\alpha^c) = \mathbb{E}_{\mathbf{h}_t^{c}} \left[ -\alpha^c \left( \log \pi_{\theta^{c}}(\tilde{\mathbf{A}}_t^{c} \mid \mathbf{h}_t^{c}) + \bar{\mathcal{H}}^{c} \right) \right].
\end{equation}

Finally, to ensure stability in temporal difference learning, the target critic parameters $\bar{\psi}_m^{c}$ are effectively updated using Polyak averaging. This mechanism slowly tracks the learned networks:
\begin{equation}
    \bar{\psi}_m^{c} \leftarrow \beta^{c} \psi_m^{c} + (1-\beta^{c})\bar{\psi}_m^{c}, \quad m \in \{1,2\},
\end{equation}
where $\beta^{c} \in (0,1)$ represents the soft update rate. The complete optimization procedure for the Central Agent is summarized in Algorithm \ref{alg:central_agent_sac}.

\begin{algorithm}[!t]
\caption{Central Agent Training}
\label{alg:central_agent_sac}
\DontPrintSemicolon
\SetAlgoLined

\KwIn{
A random batch of historical transitions $\{(\mathbf{h}_t^{c}, \mathbf{A}_t^{c}, R_t^{c}, \mathbf{h}_{t+1}^{c}, d_t)\}_{i=1}^{B^{c}}$;
current parameters $\{\theta^{c}, \psi_{1,2}^{c}, \bar{\psi}_{1,2}^{c}, \alpha^{c}\}$;
hyperparameters batch size $B^{c}$; discount $\gamma^{c}$; polyak $\beta^{c}$; target entropy $\bar{\mathcal{H}}^{c}$.
}
\KwOut{Updated parameters $\{\theta^{c}, \psi_{1,2}^{c}, \bar{\psi}_{1,2}^{c}, \alpha^{c}\}$.}

\BlankLine
\textbf{Critic update:}\;
\Indp
Sample next action from current policy:
$\tilde{\mathbf{A}}_{t+1}^{c} \sim \pi_{\theta^{c}}(\cdot \mid \mathbf{h}_{t+1}^{c})$\;

Compute soft Bellman target:\;
$
y_t^{c} \leftarrow
R_t^{c}
+
\gamma^{c}(1-d_t)
\Big(
\min_{m\in\{1,2\}} Q_{\bar{\psi}_m^{c}}(\mathbf{h}_{t+1}^{c},\tilde{\mathbf{A}}_{t+1}^{c})
-
\alpha^{c}\log \pi_{\theta^{c}}(\tilde{\mathbf{A}}_{t+1}^{c}\mid \mathbf{h}_{t+1}^{c})
\Big)
$\;

Update critics by minimizing mean squared error:\;
$
\psi_m^{c} \leftarrow
\psi_m^{c}
-
\eta_Q^{c}\nabla_{\psi_m^{c}}
\Big(Q_{\psi_m^{c}}(\mathbf{h}_t^{c},\mathbf{A}_t^{c})-y_t^{c}\Big)^2,
\quad \forall m\in\{1,2\}
$\;
\Indm

\BlankLine
\textbf{Actor update:}\;
\Indp
Sample current action with reparameterization:
$\tilde{\mathbf{A}}_{t}^{c} \sim \pi_{\theta^{c}}(\cdot \mid \mathbf{h}_{t}^{c})$\;

Update actor by maximizing soft value:\;
$
\theta^{c} \leftarrow
\theta^{c}
-
\eta_\pi^{c}\nabla_{\theta^{c}}
\Big(
\alpha^{c}\log \pi_{\theta^{c}}(\tilde{\mathbf{A}}_{t}^{c}\mid \mathbf{h}_{t}^{c})
-
\min_{m\in\{1,2\}} Q_{\psi_m^{c}}(\mathbf{h}_{t}^{c},\tilde{\mathbf{A}}_{t}^{c})
\Big)
$\;
\Indm

\BlankLine
\textbf{Temperature update:}\;
\Indp
Adjust temperature to match target entropy:\;
$
\alpha^{c} \leftarrow
\alpha^{c}
-
\eta_\alpha^{c}\nabla_{\alpha^{c}}
\Big(
-\alpha^{c}\big(\log \pi_{\theta^{c}}(\tilde{\mathbf{A}}_{t}^{c}\mid \mathbf{h}_{t}^{c})+\bar{\mathcal{H}}^{c}\big)
\Big)
$\;
\Indm

\BlankLine
\textbf{Target network update:}\;
\Indp
Update target critics via Polyak averaging:\;
$
\bar{\psi}_m^{c} \leftarrow
\beta^{c}\psi_m^{c} + (1-\beta^{c})\bar{\psi}_m^{c},
\quad \forall m\in\{1,2\}
$\;
\Indm

\end{algorithm}

\subsection{Few-Shot Learning}

Upon deployment to an unseen infrastructure layout $C_{n}$, the system must bridge the gap between the generalized knowledge acquired during meta-training and the specific requirements of the local environment. 
Once effective transfer to unseen layouts has been jointly verified by meta-training and meta-validation, we checkpoint the complete training state, including the network parameters and the optimizer states accumulated during training. We then perform a large-scale few-shot evaluation on the held-out test set, where each unseen layout is adapted with only a small number of additional interactions. Our framework employs a hierarchical adaptation procedure that balances two competing objectives: the need for Area Agents to rapidly specialize to local physical heterogeneity, and the need for the Central Agent to maintain global coordination stability while harmonizing with the shifting dynamics of the subordinates. 

The Area Agents serve as the primary responders to the physical variations of charging stations, such as topology changes or local demand surges. Upon deployment to an unseen infrastructure layout $C_{n}$, the agent faces a cold start problem: while the context encoder is capable of inferring task-specific dynamics, it requires a history of interactions to function. Therefore, the few-shot learning process here serves a dual purpose. It acts primarily as a data acquisition phase to populate the context needed for the encoder by interacting with the environment and restoring contexts to the new replay buffer upon $C_{n}$, unlocking the inference mechanism. Simultaneously, it leverages the hybrid meta-training strategy, which ensures that the global initialization $\mathbf{w}_j$ is not arbitrary but is pre-positioned in the parameter space to offer both reasonable zero-shot behavior and a favorable landscape for rapid convergence.

Similar to meta-validation step, the adaptation process commences with a context and transitions collection phase. The agent interacts with the new environment for a few steps using the meta-learned initialization $\mathbf{w}_j = (\theta_j^a, \psi_j^a)$ and the prior latent estimate $p(\mathbf{z})$. This interaction yields a small support context batch $\mathcal{M}_{n}^j$ and a transition batch $\tilde{\mathcal{M}}_{n}^j$, which provides the necessary evidence to identify the unique dynamics of the current infrastructure. Once this support set is collected, the fixed context encoder $\phi_j$ processes the interaction history to infer a task-specific latent representation. We use a deterministic inference rule by taking the posterior mean rather than sampling:
\begin{equation}
    \mathbf{z}^j_{n} \leftarrow \mathbb{E}\!\left[\mathcal{Q}_{\phi_j}(\mathbf{Z}^j \mid \mathcal{M}_{n}^j)\right].
\end{equation}

While latent conditioning provides immediate coarse adaptation, we achieve fine-grained specialization through gradient-based tuning. Starting from the robust global initialization $\mathbf{w}_{j,0} = \mathbf{w}_j$, we perform $L_u$ steps of gradient descent on the actor-critic parameters. The resulting parameters $\mathbf{w}_{j,L_u}$ represent a specialized policy tailored to the specific infrastructure $C_{n}$. During few-shot adaptation and evaluation, action generation is also made deterministic, again mirroring meta-validation: the actor outputs the mean action to reduce variance and to measure adaptation progress consistently across test layouts.

While the Area Agents undergo substantial parameter tuning, the Central Agent employs a strategy of conservative fine-tuning. Although the Central Agent is less sensitive to local heterogeneity because it relies on aggregated state representations, the rapid adaptation of subordinate Area Agents inevitably introduces distributional shifts in the system-level demand and supply signals. Consequently, the previously learned value estimates may become misaligned with the updated system dynamics. To harmonize with these evolving conditions without destabilizing the global coordination logic, we permit the Central Agent to perform limited updates for both actor and critic networks.

However, performing unconstrained optimization in this context presents significant risks. First, aggressive updates on a single new deployment instance can lead to catastrophic forgetting, where the agent overfits to local idiosyncrasies and erases the generalized resource allocation logic acquired during meta-training. Second, since the Area Agents are simultaneously modifying their policies, the Central Agent faces a non-stationary environment. Rapidly updating the central policy against these shifting subordinate behaviors creates a moving target problem, which can induce system wide instability. To mitigate these risks, we adopt a conservative optimization strategy. First, regarding the magnitude of updates, we enforce a constraint on the global $L_2$-norm of the gradients. Any gradient vector exceeding a predefined threshold is rescaled to bound the magnitude of each update step. Second, regarding the frequency of updates, we explicitly regulate the central adaptation pace: the Central Agent performs a gradient update after every $K_{c}$ optimization steps of the Area Agents. 

When coupled with a reduced learning rate, this mechanism serves to restrain the adaptation dynamics. It mitigates the risk of catastrophic forgetting by preventing large deviations from the prior knowledge. Furthermore, the staggered update frequency promotes a hierarchical timescale separation: by ensuring the Central Agent adapts significantly slower than the Area Agents, we approximate a quasi-static coordination environment. This design facilitates the specialization of Area Agents against a relatively stable global signal, thereby alleviating the divergence issues commonly observed in simultaneous multi-agent adaptation. For consistency with the deterministic evaluation protocol, the Central Agent also operates in mean mode during few-shot deployment, using mean actions for coordination to reduce stochasticity in system-level assessments.

\subsection{Heuristic Operational Module}

The tactical directives $\mathbf{A}_t^{a,j}$ generated by the Area Agents provide macroscopic operational guidance rather than direct microscopic control. To translate these strategic and tactical signals into executable fleet level decisions, we design a deterministic heuristic dispatching algorithm that interfaces between the learning agents and the physical environment. This heuristic operates at a finer temporal resolution by subdividing each strategic decision period $t$ into $D$ discrete intervals indexed by $\eta$. Its primary role is to resolve vehicle assignments for order fulfillment, charging, and relocation while ensuring that the aggregate fleet behavior remains consistent with the budgets and incentives specified by the reinforcement learning policy. Throughout execution, all physical constraints, including battery limitations and charging station capacities, are strictly enforced. The cumulative operational outcome over the $D$ intervals is aggregated as the regional reward $R_t^{j}$ for the corresponding Area Agent.

The dispatching decisions are governed by a score maximization principle. For each candidate assignment, such as assigning vehicle $k$ to order $r$, a composite utility score $\tilde{U}_{r,k}$ is computed. This score consists of two components: a baseline economic value $U_{r,k}$ and a weighting term $\omega_r(\mathbf{A}_t^{a,j})$ induced by the agent’s action. The baseline value $U_{r,k}$ captures the immediate network profit of serving request $r$, defined as the expected fare revenue minus operating costs associated with travel and a nonlinear penalty for customer waiting time. The waiting penalty is modeled as a quadratic function of elapsed time to ensure that orders with long waiting time are progressively prioritized.

The weighting term $\omega_r(\mathbf{A}_t^{a,j})$ is derived from the Area Agent’s action vector and acts as a soft constraint that biases the heuristic toward assignments aligned with strategic objectives. When the agents allocate a higher budget to a specific activity or route, the corresponding augmentation term increases, effectively steering the dispatching decisions. This term is further scaled by the remaining demand and the number of remaining intervals $\bar{D}_{\eta}$ within the current period, ensuring that resource allocation is smoothed over time rather than concentrated at the beginning of the period, where $\bar{D}_{\eta} = D - \eta + 1$.

Within each operatinal interval $\eta$, the heuristic executes four sequential decision phases:
\begin{enumerate}
    \item \textbf{Safety protocol}. Prior to any economic optimization, a mandatory safety check is performed. For each vehicle $k$, the minimum energy required to reach the nearest charging station plus a safety buffer $\xi^0$ is computed. Vehicles whose SOC $\xi_k$ falls below this threshold are immediately removed from the dispatchable pool and routed to the closest charging station. This action incurs a cost equal to the corresponding travel expense plus the maximum charging cost $c_{e}^{\max}$.
    \item \textbf{High-SOC order fulfillment}. Vehicles with sufficiently high SOC are prioritized for passenger service. These vehicles evaluate all pending orders according to scores $\tilde{U}_{r,k}^{\mathrm{H}}$ derived from reinforcement learning, and assignments are chosen to maximize combined economic and strategic utility. Upon successful order fulfillment, the accumulated reward increases by the realized net revenue.
    \item \textbf{Dual decision for low-SOC vehicles}. Vehicles with low SOC face a tradeoff between immediate service and energy replenishment. Each such vehicle evaluates both the most profitable service option and the best available charging alternative. The charging score $\tilde{U}_{k,C}$ incorporates the travel cost to station $C$, the charging price $c_{e}$, and the expected queuing delay cost $c_w$. The charging incentive produced by the reinforcement learning model is further adjusted according to the energy recovery rate, thereby favoring fast charging options when urgency is high. The vehicle executes the action with the higher composite score.
    \item \textbf{Opportunistic charging}. Remaining high-SOC vehicles that are not matched to orders evaluate opportunistic charging opportunities. If the RL policy strongly incentivizes charging, yielding a positive net score despite the absence of immediate necessity, these vehicles proceed to charge to prepare for future demand.
\end{enumerate}

After completing all $D$ intervals within period $t$, a relocation phase is performed to restore spatial balance. Idle vehicles $\mathcal{K}_t^{0}$ are redistributed to neighboring regions according to the target relocation ratios $\mathbf{d}_t$ specified by the Area Agent. When the available idle fleet is insufficient to satisfy the total relocation demand, a proportional dispatching scheme is applied. Specifically, the number of vehicles assigned to each destination is scaled by the ratio of available supply to total demand, ensuring that the relative spatial distribution prescribed by the RL policy is preserved even under resource scarcity. Algorithm~\ref{alg:heuristic_main} shows the overall procedure of the heuristic operation module.

\begin{algorithm}[!t]
\SetAlgoLined
\DontPrintSemicolon
\caption{Heuristic Dispatching with Utility Derived from Reinforcement Learning}

\label{alg:heuristic_main}

\KwIn{
Passenger requests $\mathcal{R}_t$; 
vehicle set $\mathcal{K}_t$;
area agent action $\mathbf{A}_t^{a,j}(\forall j \in \mathcal{J})$;
number of intervals $D$.}
\KwOut{Aggregated period reward $R_t^{j}$ and updated system state.}

\KwInit{$R_t = 0$ and partition $\mathcal{K}_t$ into high- and low-SOC subsets.}

\BlankLine

\For{$\eta=1$ \KwTo $D$}{
    
    \tcp{\textbf{Phase I: Safety Charging}}
    Route vehicles $k$ with $\xi_k < \xi^0$ to the nearest feasible charging station;
    Update $R_t$\;

    \tcp{\textbf{Phase II: High-SOC Order Fulfillment}}
    \ForEach{idle vehicle $k$ with high-SOC}{
        Compute composite utility $\tilde{U}_{r,k}$ for all feasible $r \in \mathcal{R}_t$\;
        Assign $(r^\ast,k)$ with maximal $\tilde{U}_{r,k} > 0$;
        Update $R_t$\;
    }

    \tcp{\textbf{Phase III: Low-SOC Dual Decision}}
    \ForEach{idle low-SOC vehicle}{
        Compute best service utility $\max_r \tilde{U}_{r,k}$ and best charging utility $\tilde{U}_{k,C}$\;
        Execute the option with higher utility;
        Update $R_t$\;
    }

    \tcp{\textbf{Phase IV: Opportunistic Charging}}
    Assign remaining idle high-SOC vehicles to charge if $\tilde{U}_{k,C} > 0$\;

    Apply waiting and abandonment penalties for unserved requests;
    Update $R_t$\;
}

\tcp{\textbf{Phase V: Strategic Relocation}}
Redistribute idle vehicles according to target relocation ratios $\mathbf{d}_t$;
apply proportional scaling if supply is insufficient;
Update $R_t$\;

\KwRet{$R_t$}
\end{algorithm}

\section{Numerical Examples}\label{sec_numerical}

This section presents numerical study to evaluate the effectiveness of the proposed hierarchical GAT-PEARL framework. All experiments are conducted in a simulator calibrated using realistic ride-hailing data. Demand follows empirical temporal and spatial patterns, while vehicle operations and charging dynamics are simulated with parameterized models under dynamic charging network setting. First, we examine the proposed framework's generalization performance under previously unseen charging station layouts, to verify whether the learned meta policy provides a high-quality initialization that remains effective under different infrastructure layouts. Second, we evaluate adaptation efficiency by measuring how quickly the Area Agents adjust their policies to new charging layouts using limited interaction data. Third, we conduct a series of sensitivity analyses by varying the number of charging piles per station, charging station density, and overall demand intensity. In each setting, the proposed framework is compared against the hierarchical SAC baseline.

\subsection{Experimental Setup}\label{subsec_Experimental_Setup}

\begin{figure}[!htbp]
    \centering
    \includegraphics[width=0.8\linewidth]{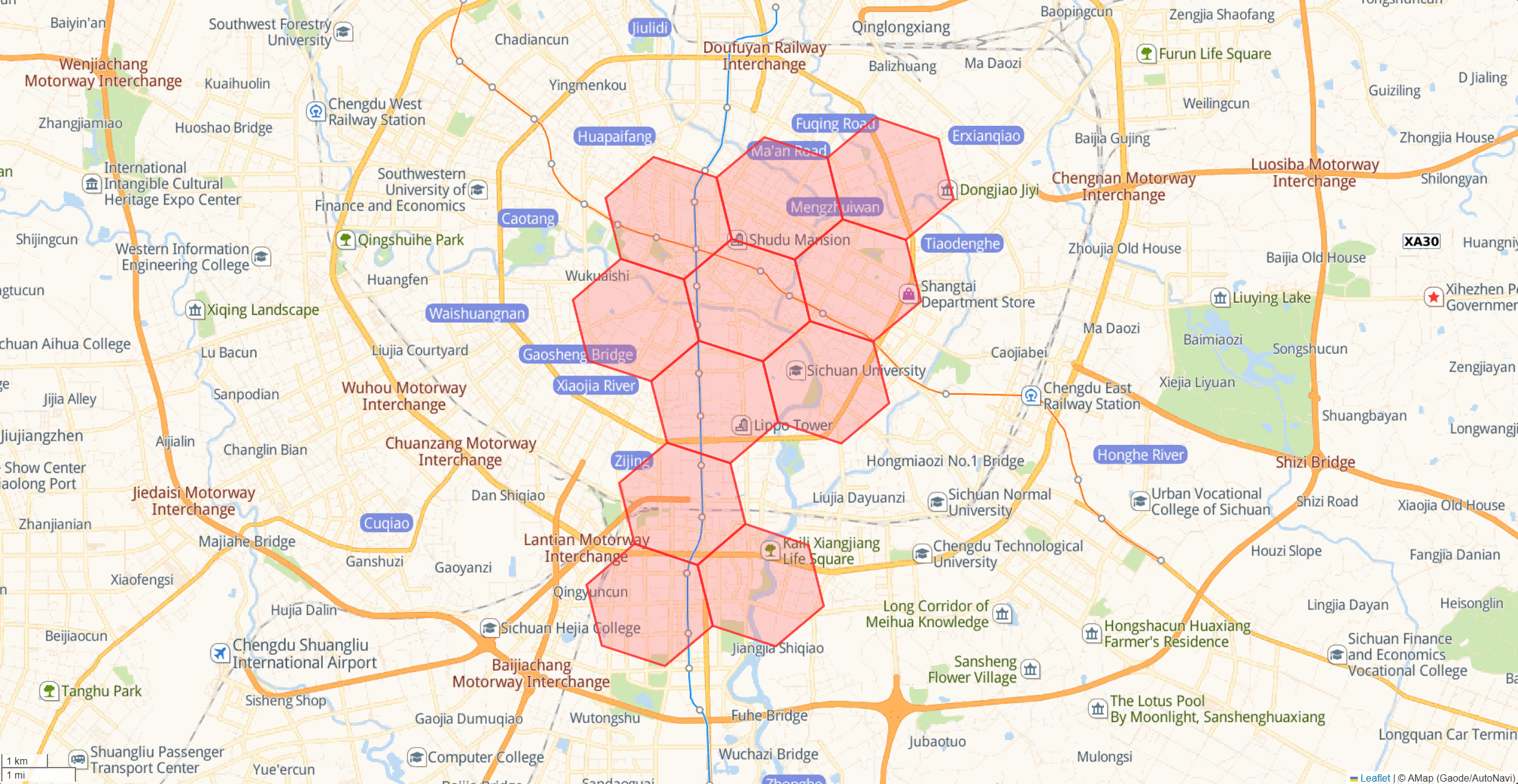} 
    \caption{The experimental areas in Chengdu, China.}
    \label{fig:chengdu_map}
\end{figure}

\begin{figure}[!htbp]
\centering
\subfloat[Spatial distribution of order density\label{fig:setup_spatial}]{%
  \includegraphics[width=0.48\textwidth,height=4.2cm,keepaspectratio]{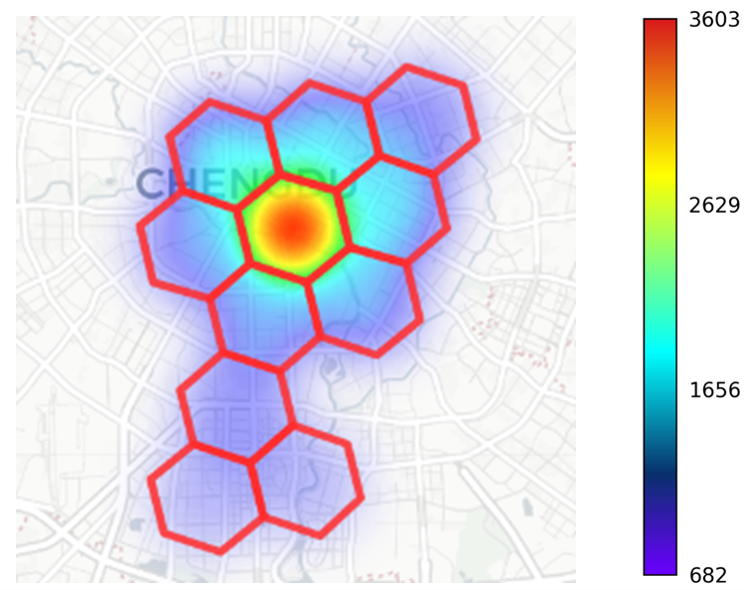}%
}\hspace{1em}
\subfloat[Temporal distribution of order arrivals\label{fig:setup_temporal}]{%
  \includegraphics[width=0.48\textwidth,height=4.2cm,keepaspectratio]{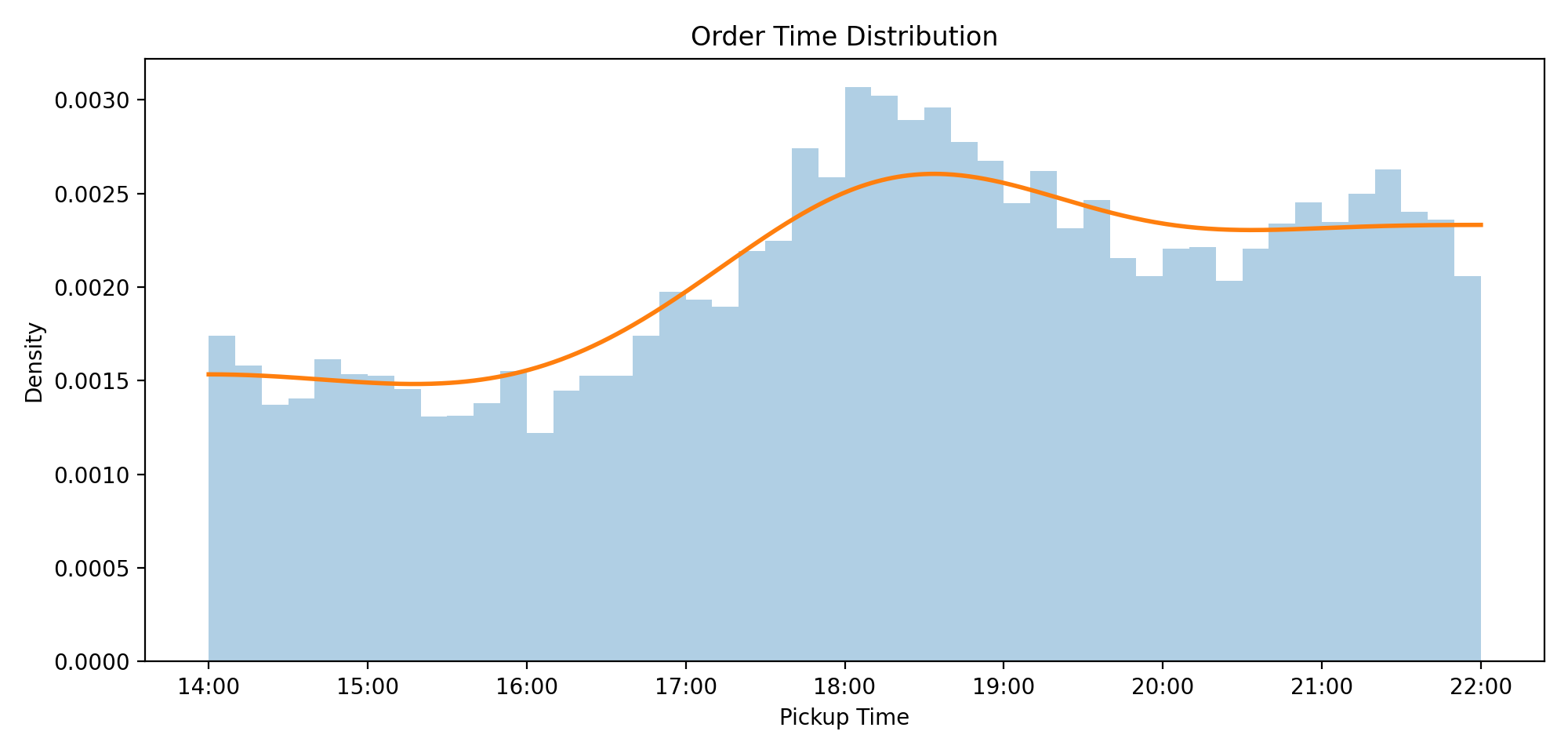}%
}
\caption{The spatio-temporal characteristics of the passenger requests.}
\label{fig:experimental_setup}
\end{figure}

We evaluate the proposed framework using a simulator calibrated with real-world operational data from Chengdu, China, covering the period from September to November 2020. To align with the spatial representation used in real-world operations, we discretize the study area into H3 Level-7 hexagonal cells, where each cell has an average area of 5.161~km$^2$ and an average edge length of 1.221 km \footnote{\url{https://h3geo.org/docs/core-library/restable/}}. Among these, we select the 11 regions with the highest historical order volumes as the experimental area. The geographical layout of these selected hexagonal zones is illustrated in Figure~\ref{fig:chengdu_map}, covering the core urban districts of Chengdu. The heatmap of the passenger requests' origins is visualized in Figure~\ref{fig:experimental_setup}(a), which highlights the spatial heterogeneity of order density across the network and indicates distinct demand hotspots and cooler peripheral zones. The simulation horizon spans 8 hours, from 2:00 PM to 10:00 PM. Following the hierarchical temporal structure introduced in Section~\ref{subsec_problem_statement}, the horizon is divided into decision periods of 15 minutes and operational intervals of one minute. Passenger demand is synthetically generated based on the temporal distribution at the minute level and spatial patterns between origins and destinations extracted from the historical dataset. The resulting demand set contains approximately 15,000 orders per episode. Figure~\ref{fig:experimental_setup}(b) shows the temporal pattern of the generated orders, indicating that the simulation horizon spans the evening rush-hour and includes a clear demand peak between 18:00 and 19:00. Each order is associated with a maximum waiting tolerance $t_r^{\max}$, independently sampled from a uniform distribution over the interval $[0, 15]$ minutes.

\begin{figure}[!htbp]
    \centering
    \includegraphics[width=0.45\linewidth]{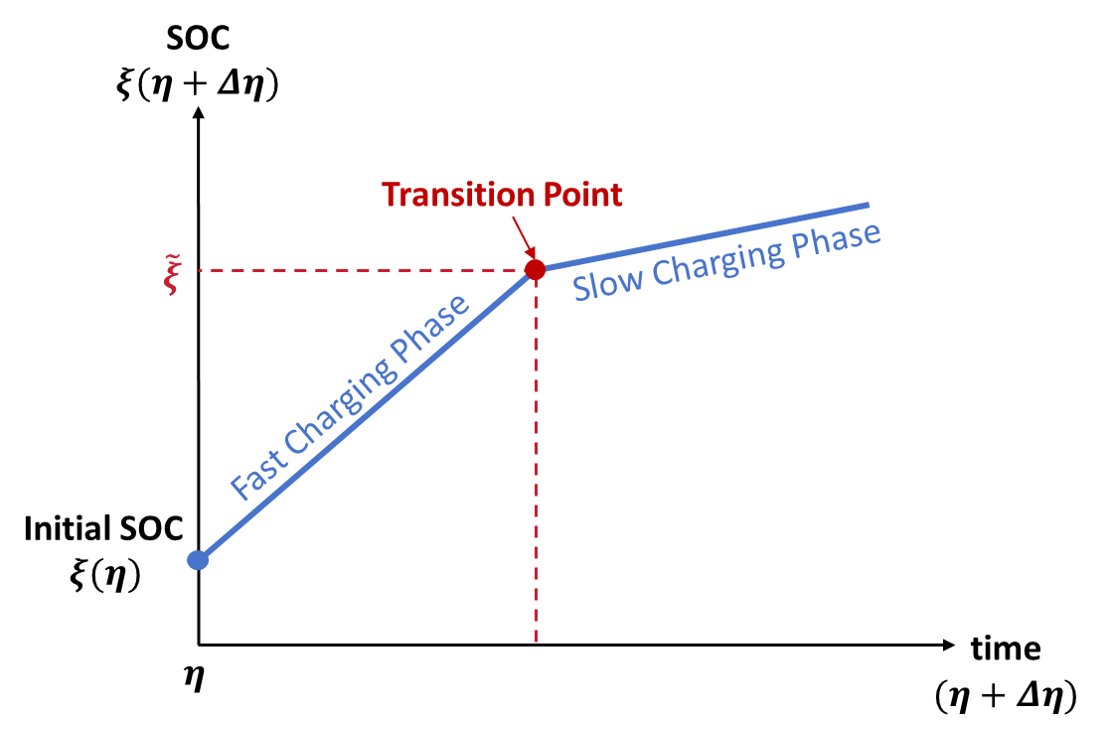}
    \caption{Illustration of the piecewise linear SOC evolution during charging.}
    \label{fig:soc_piecewise}
\end{figure}

On the supply side, the system is operated by a fleet of 1,100 electric vehicles. To reflect realistic spatial supply patterns, the initial fleet distribution across regions is set proportional to historical demand density rather than uniformly assigned. In addition, to capture realistic labor supply dynamics, vehicles do not enter service simultaneously. Instead, drivers follow stochastic shift schedules, with vehicles gradually entering the system during the first four decision periods, corresponding to the interval from 2:00 PM to 3:00 PM. At the beginning of each simulation episode, the initial SOC of each vehicle is independently drawn from a uniform distribution over $[\xi_{\max}/3, \xi_{\max}]$, where $\xi_{\max}$ denotes the maximum battery capacity, representing a fleet that starts operation with sufficiently charged batteries. Vehicle charging behavior is modeled using a state-dependent charging process that captures the nonlinear characteristics of lithium-ion batteries. Specifically, the evolution of SOC over consecutive charging durations is approximated by a piecewise linear function, as shown in equation \eqref{equ:SOC}:

\begin{equation}
    \xi(\eta + \Delta\eta)
    =
    \begin{cases}
    \xi(\eta) + p_1 \Delta\eta,
    & \text{if } \xi(\eta) + p_1 \Delta\eta < \tilde{\xi}, \\[6pt]
    \tilde{\xi}
    + p_2 \left(
    \Delta\eta
    -
    \dfrac{\tilde{\xi} - \xi(\eta)}{p_1}
    \right),
    & \text{otherwise}
    \end{cases}
    \label{equ:SOC}
\end{equation}

where $\xi(\eta)$ denotes the SOC at the beginning of charging, $\Delta\eta$ is the number of consecutive operational intervals allocated to charging, and $\tilde{\xi}$ represents the transition point between fast and slow charging phases. Figure~\ref{fig:soc_piecewise} provides an illustrative depiction of the resulting piecewise SOC trajectory.

To assess the generalization capability of the proposed policy, we define the meta-learning task distribution by varying the spatial layout of the charging infrastructure. Specifically, each task represents a unique topology where 5 charging stations are deployed among the 11 candidate zones. This specific density is selected to balance service coverage with infrastructure constraints, drawing on established empirical findings. Previous studies indicate that while overly sparse networks significantly hinder user accessibility and technology adoption \citep{amilia2022designing}, increasing density beyond a saturation point yields diminishing marginal returns in operational efficiency \citep{shahraki2015optimal}. Accordingly, the chosen parameter reflects a critical threshold that ensures adequate coverage without reaching saturation. The resulting combinatorial task space is randomly divided into training, validation, and testing subsets in an 8:1:1 proportion. A total of 47 charging network layouts are reserved strictly for the test set, where they serve as unseen environments for evaluating performance and conducting sensitivity analysis.

Table \ref{tab:hyperparameters} lists the detailed hyperparameter settings. For the Central Agent, the action space constraints are defined by a maximum net flow ratio of 0.3, a minimum charging quota of 0.8, and a maximum cost multiplier of 0.2. The network architecture consists of an Actor with 2 hidden layers and a Critic with 3 hidden layers, both using a hidden dimension of 256. Optimization is performed with a batch size of 32, using learning rates of $1 \times 10^{-5}$ for the Actor, $1 \times 10^{-4}$ for the Critic, and $3 \times 10^{-4}$ for the temperature parameter. The target entropy is set to the negative magnitude of the action dimension. The penalty weight $\lambda$ is set to $1 \times 10^2$. For the Area Agent, the GAT module is configured with a hidden dimension of 16, 8 attention heads, and a LeakyReLU slope of 0.2. The temporal embedding dimension is set to 8, and the context encoder produces a 32 dimensional latent variable. The subsequent MLP architecture uses a single hidden layer with 512 units. The batch sizes for the encoder and Actor-Critic updates are 32 and 16, respectively. Learning rates are set to $3 \times 10^{-5}$ for the Actor, $3 \times 10^{-4}$ for the Critic, and $1 \times 10^{-4}$ for both the encoder and temperature. The KL-divergence regularization coefficient $\beta^{\mathrm{KL}}$ is linearly annealed from 0 to 0.1 over the first 300 training epochs, while the PEARL representation weight $\beta^{\mathrm{R}}$ is fixed at 0.8 throughout training. Common settings shared across both agents include the use of ReLU activation functions and the Adam optimizer. The discount factor is fixed at 0.99, and the soft target update rate is set to 0.005. Regarding the training protocol, the meta-training phase is conducted for 5,000 epochs and few-shot learning process is conducted for 1,000 epochs. In this study, an epoch is defined as a complete cycle comprising both meta-training and meta-validation phases. In each epoch, the batch size of tasks $\{C_i\}$ is 15, with 10 tasks used for meta-training and 5 tasks reserved for meta-validation. To improve optimization stability, 3 meta-iterations are taken per epoch.

\begin{table}[t]
    \centering
    \caption{Hyperparameter settings of the experiment and proposed framework.}
    \label{tab:hyperparameters}
    \renewcommand{\arraystretch}{1.2} 
    \begin{tabular}{lllc}
        \toprule
        \textbf{Category} & \textbf{Hyperparameter} & \textbf{Symbol} & \textbf{Value} \\
        \midrule
        
        \multirow{12}{*}{Central Agent} 
          & Maximum net flow ratio & $f_{\max}$ & 0.3 \\
          & Minimum charging quota & $q_{\min}$ & 0.8 \\
          & Maximum cost multiplier & $p_{\max}$ & 0.2 \\
          & Actor MLP hidden layers & $-$ & 2 \\
          & Critic MLP hidden layers & $-$ & 3 \\
          & MLP hidden layer dimension & $-$ & 256 \\
          & Batch size & $B^{c}$ & 32 \\
          & Target entropy & $\bar{\mathcal{H}}^{c}$ & $-|\mathbf{A}_t^{c}|$ \\
          & Actor learning rate & $\eta_{\pi}^{c}$ & $1 \times 10^{-5}$ \\
          & Critic learning rate & $\eta_{Q}^{c}$ & $1 \times 10^{-4}$ \\
          & Temperature learning rate & $\eta_{\alpha}^{c}$ & $3 \times 10^{-4}$ \\
          & Penalty weighted & $\lambda$ & $1 \times 10^{2}$ \\

        \midrule 

        \multirow{17}{*}{Area Agent} 
          & GAT hidden dimension & $F'$ & 16 \\
          & GAT attention heads & $N_h$ & 8 \\
          & GAT LeakyReLU slope & $-$ & 0.2 \\
          & Temporal embedding dimension & $h_e$ & 8 \\ 
          & Encoder latent context dimension & $\dim(\tilde{\mathbf{z}}^j)$ & 32 \\
          & Number of MLP hidden layers & $-$ & 1 \\
          & MLP hidden layer dimension & $-$ & 512 \\ 
          & Encoder batch size & $N_{c}$ & 32 \\
          & Actor-Critic batch size & $-$ & 16 \\
          & Target entropy & $\bar{\mathcal{H}}^{a}$ & $-0.5 \cdot |\mathbf{A}_t^{a}|$ \\ 
          & Actor learning rate & $\eta_{\pi}^{a}$ & $3 \times 10^{-5}$ \\
          & Critic learning rate & $\eta_{Q}^{a}$ & $3 \times 10^{-4}$ \\
          & Encoder learning rate & $\eta_{\phi}$ & $1 \times 10^{-4}$ \\
          & Temperature learning rate & $\eta_{\alpha}^{a}$ & $1 \times 10^{-4}$ \\
          & Adaptation vs. Representation Gradient Ratio & $\beta^{\mathrm{R}}$ & 0.8 \\
          & KL weight (annealed) & $\beta^{\mathrm{KL}}(t)$ & $0 \rightarrow 0.1$ \\ 
          & Task adaptation step & $L$ & 10 \\ 

        \midrule 

        \multirow{4}{*}{Common Settings} 
          & MLP activation function & $-$ & ReLU \\
          & Optimizer & $-$ & Adam \\
          & Discount factor & $\gamma^{c},\gamma^{a}$ & 0.99 \\
          & Soft update rate & $\beta^{c},\beta^{a}$ & 0.005 \\
        
        \bottomrule
    \end{tabular}
\end{table}

The simulation environment is parameterized to align learning objectives with operational efficiency and economic realism. For each passenger request $r \in \mathcal{R}$, the service revenue $\varrho_r$ is defined based on its trip distance with a fixed unit rate of $4$ per unit travel distance. Conversely, vehicle movement incurs an operational travel cost of $1$ per unit travel distance, which is charged whenever a vehicle is in transit, regardless of whether it is serving a passenger, relocating, or traveling to a charging station. Passenger waiting is penalized to discourage excessive delays. Specifically, the waiting cost is proportional to the accumulated waiting time, with coefficient $c_w = 0.05$. If a request is ultimately abandoned, an abandonment penalty $\kappa_r = 10$ is incurred, reflecting the service quality loss associated with unmet demand. Vehicle charging costs are based on the time they choose to charge the battery. For each vehicle, the charging cost rate is set to $c_e = 2$ per unit charging time, accounting for electricity expenses and infrastructure usage. The battery dynamics of electric vehicles are characterized by their SOC $\xi$. The maximum battery capacity is $\xi_{\max} = 30$, while $\tilde{\xi} = 24$ denotes the SOC threshold separating fast and slow charging regimes in the piecewise linear charging model. To ensure operational safety, a minimum reserve level $\xi^0 = 3$ is enforced, below which vehicles are prohibited from participating in dispatching decisions and are routed to the nearest feasible charging station. Energy replenishment follows the piecewise linear charging process introduced earlier, with charging rates $p_1 = 0.6$ and $p_2 = 0.2$ corresponding to the fast and slow charging phases, respectively.

All computational experiments are implemented on a server running Ubuntu 22.04, equipped with dual Intel(R) Xeon(R) Platinum 8581C CPUs, 512 GB of RAM, and an NVIDIA RTX A6000 GPU. The algorithms are coded in Python 3.10, and the neural networks are constructed and trained using PyTorch 2.7.1 with CUDA 12.4.

\subsection{Results and Analysis}
\label{subsec_Results_and_Analysis}

\begin{figure}[!htbp]
    \centering
    \includegraphics[width=0.8\textwidth]{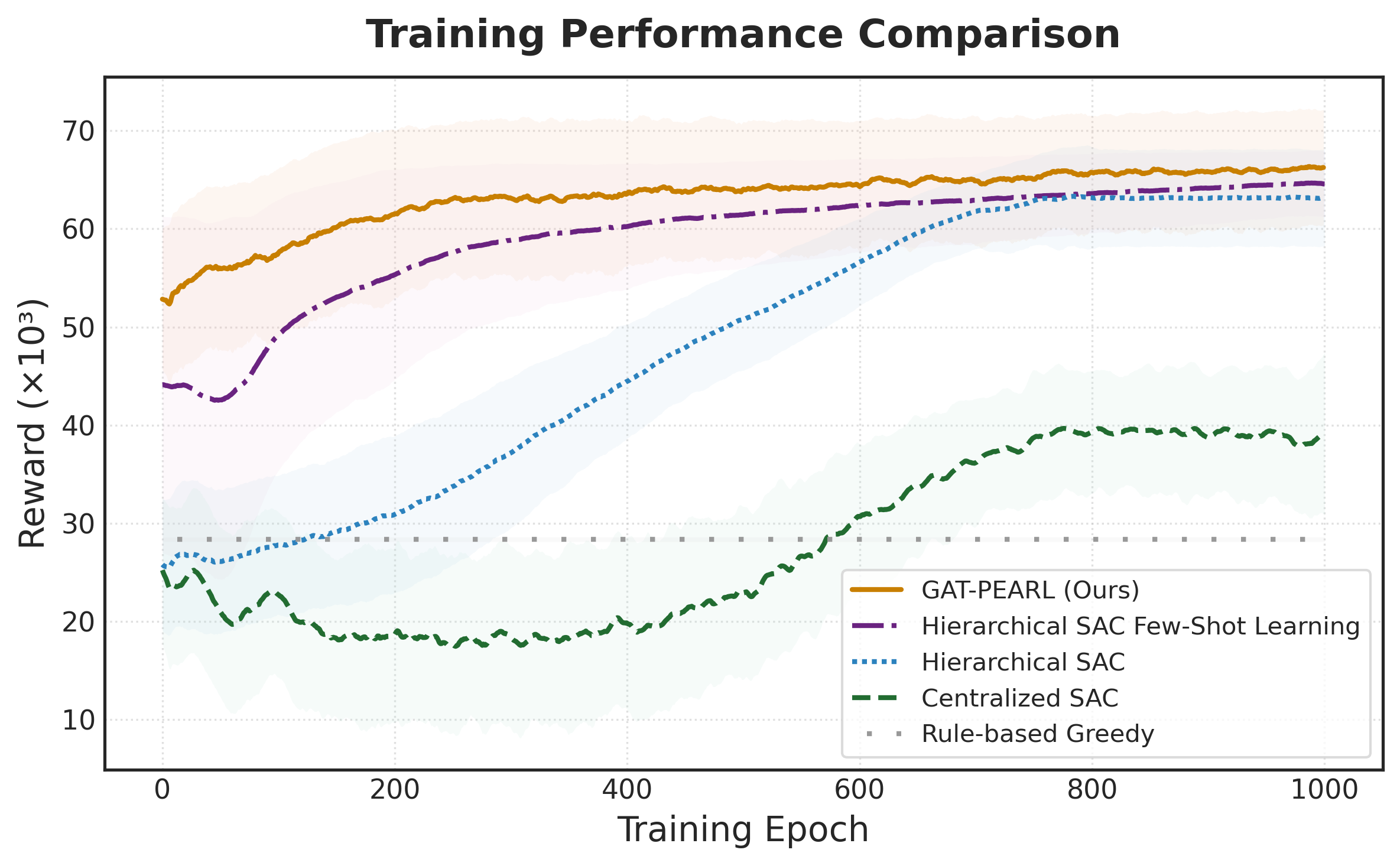}
    \caption{Training performance comparison over 1,000 epochs. Curves denote mean rewards, and shaded regions represent performance variance across test tasks.}
    \label{fig:learning_curves}
\end{figure}

This subsection evaluates the proposed framework by examining both training dynamics and asymptotic performance across the 47 unseen test tasks. The objective is to assess the effectiveness of the hierarchical decision structure and the proposed meta-RL mechanism under previously unseen charging station layouts. Figure~\ref{fig:learning_curves} reports the moving average rewards over 1,000 epochs of few-shot adaptation on the test tasks. Each curve represents the mean performance across test tasks, while the shaded bands indicate the corresponding variance.

We compare the proposed Meta-RL framework against four representative benchmarks:
\begin{itemize}
    \item \textbf{Rule-based Greedy (dotted grey line).}  
    This heuristic serves as a lower bound benchmark. Decisions are made myopically based on immediate feasibility and short-term profit, without considering future demand evolution or system-wide interactions. Vehicles do not proactively reposition toward emerging demand hotspots, and charging actions are triggered only when the battery level reaches a critical threshold.

    \item \textbf{Centralized SAC (dashed green line).}  
    This baseline employs a single centralized agent to jointly optimize dispatching and charging decisions for the entire fleet.

    \item \textbf{Hierarchical SAC (densely dotted blue line).}  
    This baseline adopts the same hierarchical decomposition as the proposed method but is trained independently for each task \textit{without} meta-learning. Accordingly, it follows a ``train-from-scratch'' paradigm for each new charging layout.

    \item \textbf{Hierarchical SAC Few-Shot Learning (dash-dot purple line).}  
    This approach evaluates direct parameter transfer. We first train the hierarchical SAC model on a single test task and save the final network parameters and optimizer states as a checkpoint. We then initialize from this checkpoint and re-run the full hierarchical SAC training procedure on the remaining test tasks.
\end{itemize}

As illustrated in Figure~\ref{fig:learning_curves}, in contrast to all benchmark methods, the proposed GAT-PEARL framework consistently achieves superior performance throughout the training process. It exhibits faster improvement in the early stages, reduced performance variance, and higher converged rewards, demonstrating strong robustness when deployed in previously unseen environments. By jointly learning a transferable policy initialization and a task embedding, our framework adapts quickly and converges reliably across different charging station layouts. A key distinction lies in the training objective. Unlike standard reinforcement learning approaches that optimize performance within a single task, our Meta-RL framework explicitly optimizes performance after adaptation through hybrid gradient updates. This design encourages the Actor-Critic networks to acquire parameters that are not only effective on average, but also highly adaptable with only a small amount of task-specific data, yielding a clear jump-start as reflected by the higher initial rewards of the orange curve. Moreover, the integration of a probabilistic context encoder fundamentally differentiates our approach from naive few-shot transfer. Instead of simple parameter reuse, the encoder infers a latent task embedding $z$ to characterize the structural variations of each charging layout. By conditioning the policy on this embedding, the framework explicitly accounts for task-specific infrastructure constraints, resulting in rapid adaptation and superior operational performance.

In contrast, the benchmark methods suffer from one or more key limitations. Although \textit{Rule-based Greedy} maintains operational feasibility, it yields consistently low rewards and exhibits no learning-driven improvement over time. \textit{Centralized SAC} is theoretically expressive, but it suffers from an excessively large state and action spaces. The resulting exploration burden significantly impedes effective learning, particularly in the early training stages. \textit{Hierarchical SAC} decentralizes decisions across Area Agents. This substantially reduces the effective action space and improves exploration relative to the centralized approach. However, because no prior knowledge is transferred across tasks, the agent must relearn charging and dispatching strategies from scratch for each new layout. This results in a slow learning curve, particularly during the initial training phase, highlighting the inefficiency of training separate models for individual tasks under infrastructure variability. \textit{Hierarchical SAC Few-Shot Learning} demonstrates faster initial improvement compared to training from scratch. However, its performance exhibits substantial variance across tasks. The effectiveness of transfer strongly depends on the similarity between the source and target charging layouts. When topological discrepancies are pronounced, the transferred parameters introduce bias, leading to unstable adaptation and degraded performance. This sensitivity underscores the limitations of naive few-shot transfer without explicit task inference.

\begin{figure}[!htbp]
    \centering
    \includegraphics[width=0.9\textwidth]{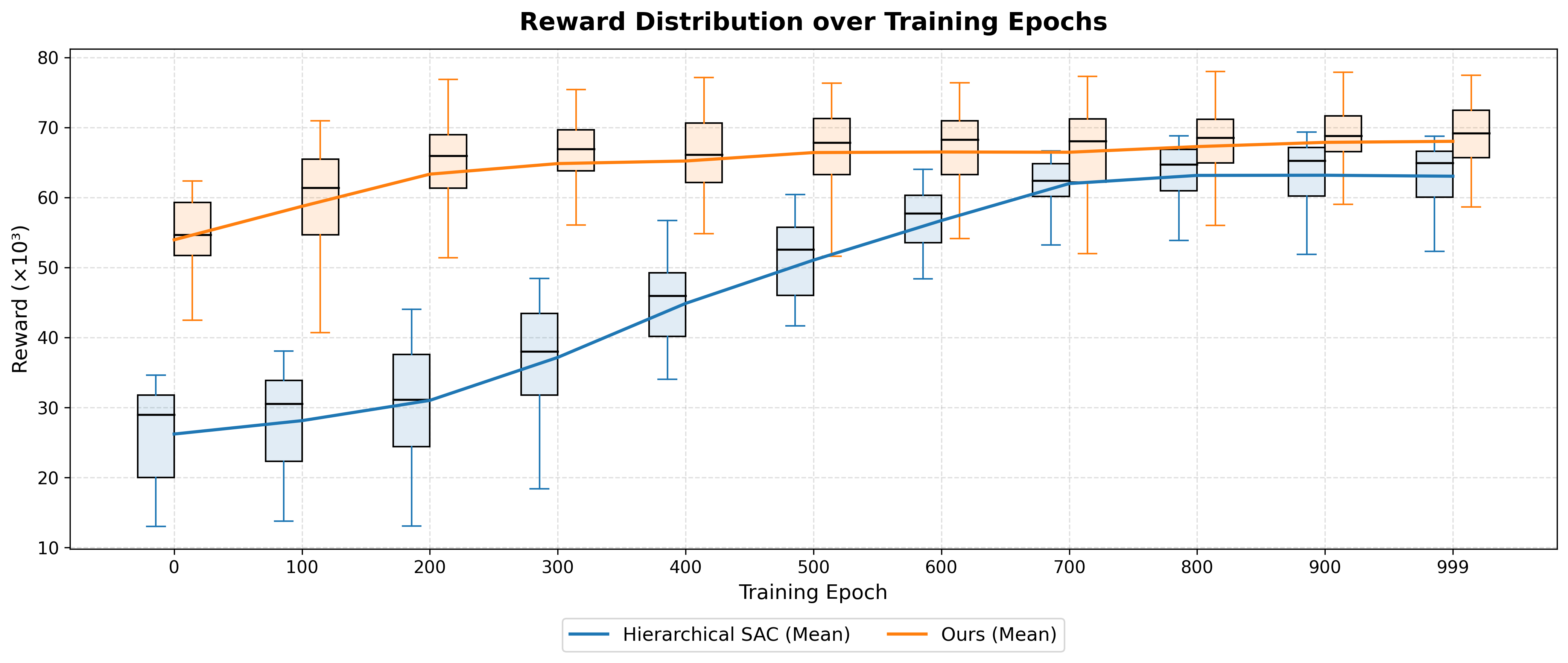}
    \caption{Reward distributions of Meta-RL Few-Shot Learning and Hierarchical SAC across 47 test tasks.}
    \label{fig:boxplot}
\end{figure}

To further examine the stability and generalization behavior, Figure~\ref{fig:boxplot} presents the reward distributions of Meta-RL Few-Shot Learning and Hierarchical SAC across all 47 unseen test tasks. The proposed Meta-RL framework exhibits a higher median reward and a tighter interquartile range, indicating superior stability and generalization. While both methods leverage few-shot adaptation, the boxplots reveal a critical contrast. The standard Hierarchical SAC baseline exhibits a wide interquartile range, indicating strong sensitivity to characteristics of individual tasks and reliance on favorable task similarity. In contrast, the proposed Meta-RL framework achieves a substantially tighter interquartile range together with a higher median reward. This reduced variability confirms that probabilistic context inference effectively decouples performance from layout-specific differences, enabling the agent to deliver consistently reliable operational outcomes across diverse infrastructure layouts. Collectively, these results demonstrate that the proposed framework not only accelerates learning, but also significantly improves robustness and generalization in AET management at a large scale.

\subsection{Sensitivity Analysis on Charging Infrastructure Capacity}
\label{subsec_sensitivity_piles}

\begin{figure}[!htbp]
    \centering
    \includegraphics[width=0.75\textwidth]{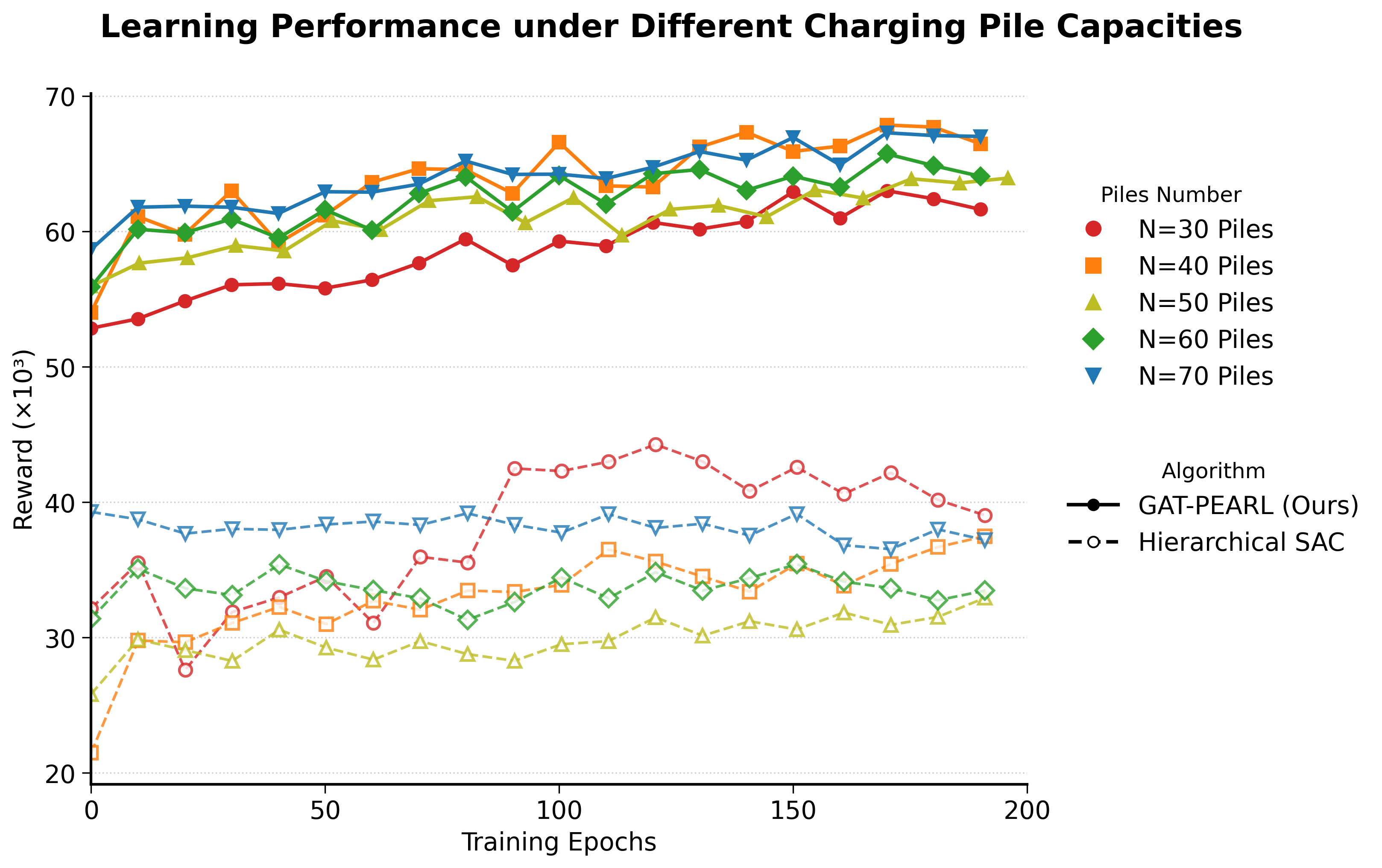}
    \caption{Learning performance comparison under different number of charging piles per station.}
    \label{fig:sens_reward}
\end{figure}

\begin{figure}[!htbp]
    \centering
    \includegraphics[width=0.8\textwidth]{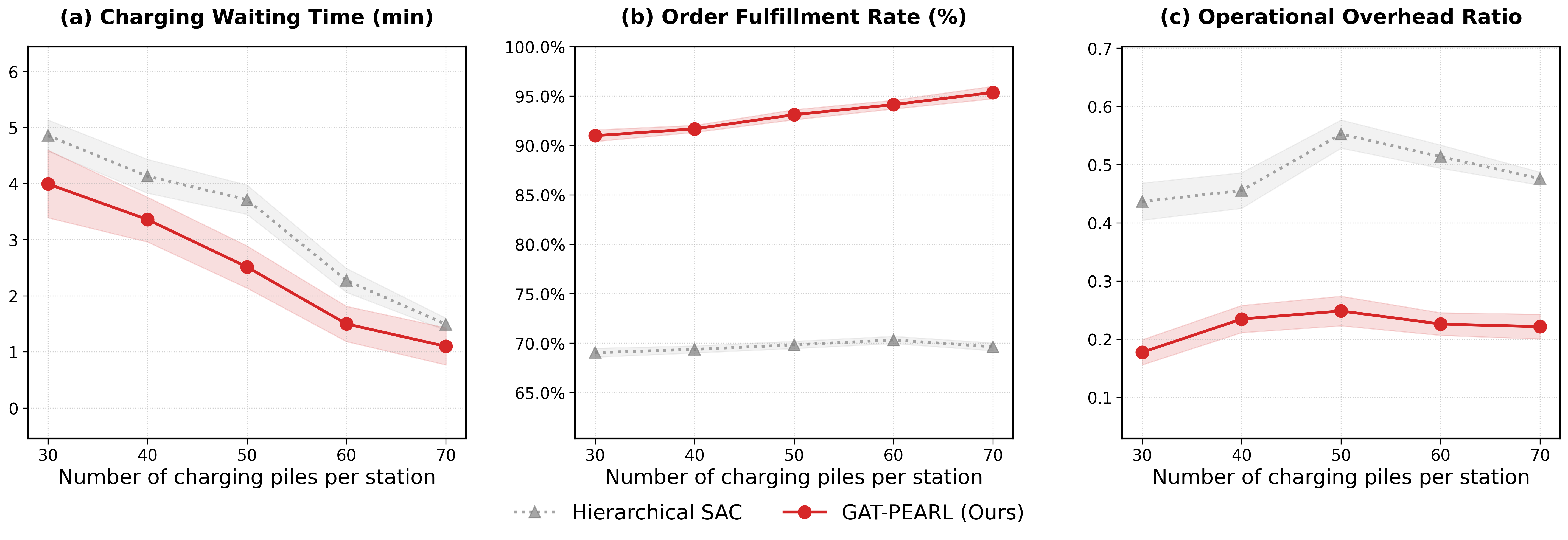}
    \caption{Impact of the number of charging piles per station on key operational metrics: (a) charging waiting time, (b) order fulfillment rate, and (c) operational overhead ratio. Shaded areas indicate the confidence interval.}
    \label{fig:sens_metrics}
\end{figure}

We further examine the robustness of the proposed framework against variations in infrastructure capacity. To represent operating environments ranging from resource scarcity to relative abundance, we vary the number of charging piles per station, denoted as $N_p$, across the set $\{30, 40, 50, 60, 70\}$. For each capacity setting, our framework is required to adapt its dispatching and charging strategy using limited interaction data. The proposed Meta-RL framework is compared against the Hierarchical SAC baseline, and the results are summarized in Figures~\ref{fig:sens_reward} and~\ref{fig:sens_metrics}.

Figure \ref{fig:sens_reward} illustrates the learning trajectories during the first 200 epochs. The solid lines represent the proposed Meta-RL framework, while the dashed lines denote the Hierarchical SAC benchmark. Across all charging capacity levels, the proposed Meta-RL framework consistently demonstrates rapid performance improvement and stable convergence. Regardless of whether the charging infrastructure is scarce or plentiful, the model reaches a high reward regime within a small number of training iterations. By contrast, the Hierarchical SAC exhibits slow learning progress accompanied by substantial variance, particularly when charging capacity is limited. An important observation concerns the magnitude of the performance gap between the two methods. Even under the most constrained configuration ($N=30$), the proposed framework achieves higher cumulative rewards than the benchmark operating under the most generous configuration ($N=70$) in the early stages of the training process. This result indicates that overall fleet performance is not determined solely by the quantity of available charging resources and the effectiveness of operational coordination and decision making plays a dominant role. By exploiting a transferable policy initialization together with latent task inference, the proposed framework learns to regulate charging demand over space and time, thereby alleviating congestion effects and compensating for limited infrastructure.

To further interpret these outcomes, Figure~\ref{fig:sens_metrics} reports three operational metrics that reveal system level behavior. Figure~\ref{fig:sens_metrics}(a) presents average charging waiting time. For all capacity levels, the proposed framework consistently reduces waiting time by approximately 1.0 to 1.5 minutes compared with the benchmark. This reduction reflects the agent's ability to anticipate future congestion and schedule charging activities in advance, rather than responding only after queues have formed. Figure~\ref{fig:sens_metrics}(b) presents the order fulfillment rate. As charging capacity increases, the proposed method shows a steady improvement, rising from approximately 91\% to more than 95\%. In contrast, the baseline remains nearly constant at around 70\% across all capacity settings. This suggests that without effective coordination mechanisms, increasing charging capacity alone does not improve service quality, since vehicles are not promptly or evenly directed to high demand areas or fully utilized charging resources. Figure~\ref{fig:sens_metrics}(c) presents the operational overhead ratio, defined as the total operational cost, including charging and repositioning, divided by total order revenue. The proposed framework maintains a low and stable ratio close to 0.2 across all configurations, indicating a balanced trade off between cost efficiency and service provision. In contrast, the baseline exhibits pronounced instability, with a clear variance increase around $N=50$, suggesting inefficient and inconsistent operational decisions.

Overall, this sensitivity analysis demonstrates that the proposed Meta-RL framework adapts effectively to changes in charging infrastructure capacity. By separating task inference from control execution and incorporating prior knowledge about infrastructure variations, the agent maintains high service quality even when charging resources are severely constrained. This robustness is particularly relevant for real world deployments, where charging capacity is heterogeneous, evolves over time, and is subject to long term planning limitations.

\subsection{Sensitivity Analysis on Charging Station Density}
\label{subsec_sensitivity_stations}

Beyond the number of charging piles per station, the spatial distribution of charging infrastructure constitutes another major source of operational uncertainty in urban electric taxi systems. To examine whether the proposed framework can accommodate such structural changes, we conduct a second sensitivity analysis by varying the number of charging stations in the network. Specifically, we consider five infrastructure configurations with the number of charging stations denoted by $N_s \in \{3,4,5,6,7\}$. All other system parameters are held fixed, allowing us to identify the impact of station density on learning behavior and operational outcomes.

\begin{figure}[!htbp]
    \centering
    \includegraphics[width=0.75\textwidth]{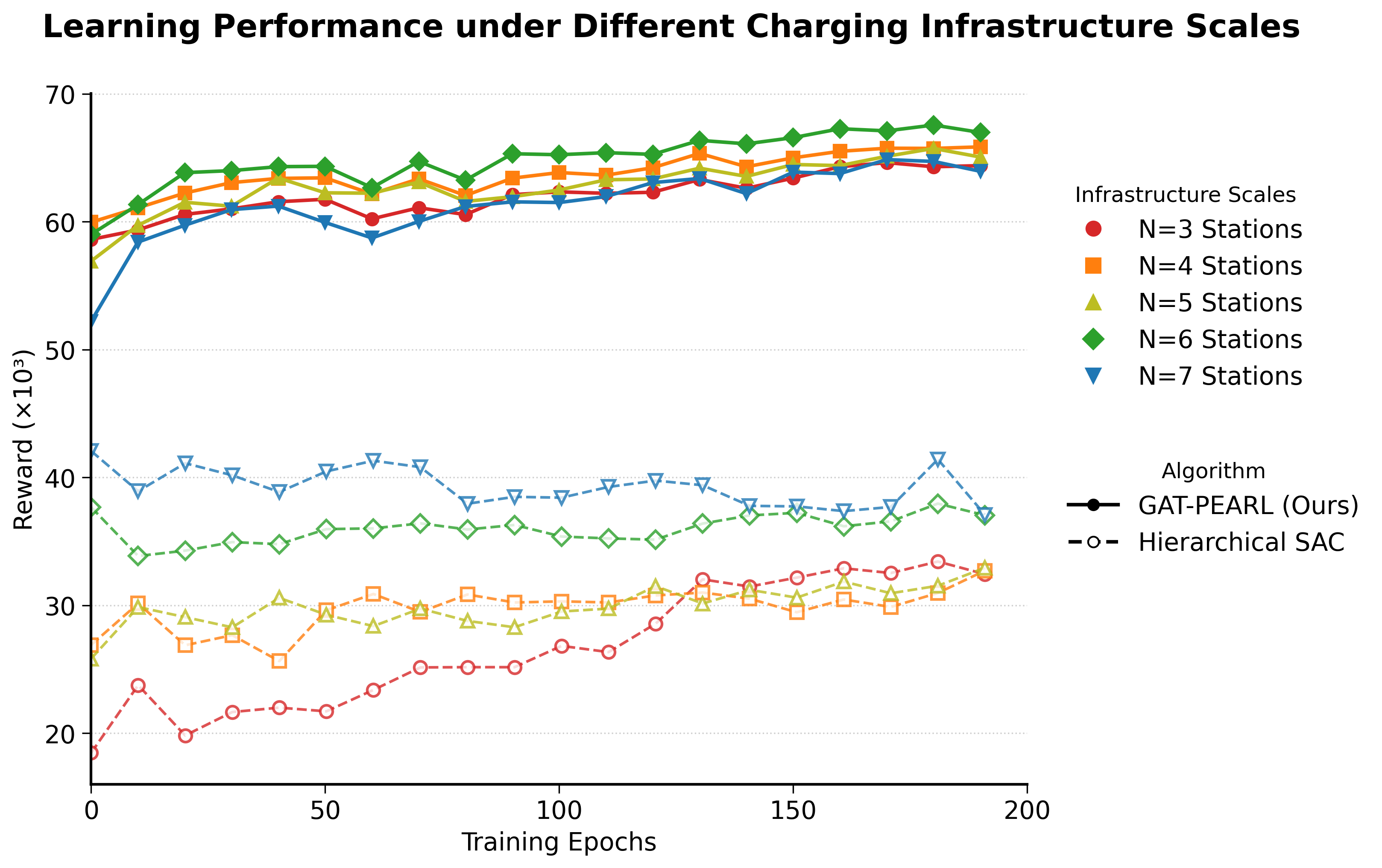}
    \caption{Learning performance comparison across different numbers of charging stations.}
    \label{fig:sens_station_reward}
\end{figure}

Figure \ref{fig:sens_station_reward} illustrates the learning trajectories during the first 200 training epochs. The solid lines represent the proposed Meta-RL framework, while the dashed lines denote the Hierarchical SAC baseline. As expected, increasing the number of charging stations improves overall performance for both methods. However, the proposed framework consistently achieves higher rewards and faster convergence across all numbers of charging stations. Even in environments with very limited spatial coverage ($N=3$), the learned policy rapidly approaches a high performance level, whereas the baseline exhibits substantial variance and slow progress. Performance gains depend critically on the ability to reason about spatial interactions and system wide dependencies. By combining graph based state encoding with latent task inference, the proposed framework effectively captures these spatial relationships and coordinates fleet behavior accordingly, even under constrained infrastructure.

\begin{figure}[!htbp]
    \centering
    \includegraphics[width=0.75\textwidth]{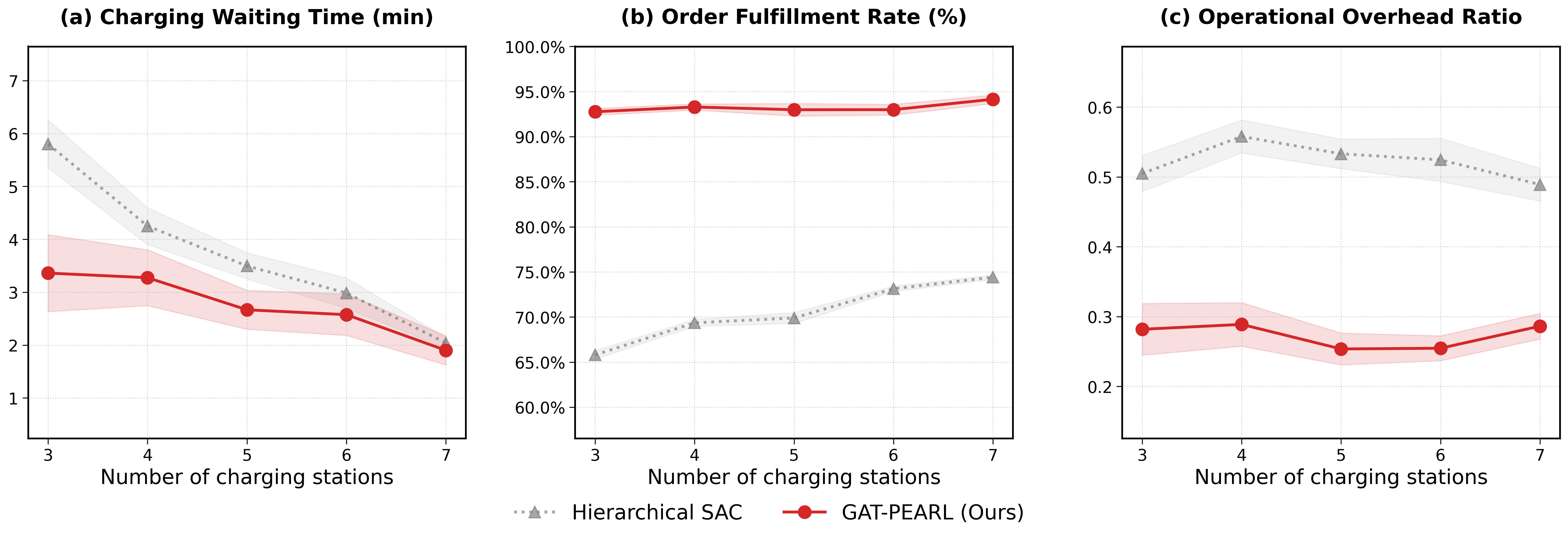}
    \caption{Impact of charging station count on key operational metrics: (a) charging waiting time, (b) order fulfillment rate, and (c) operational overhead ratio. Shaded areas indicate the confidence interval.}
    \label{fig:sens_station_metrics}
\end{figure}

Figure~\ref{fig:sens_station_metrics} reports three operational metrics. First, with respect to temporal efficiency (Figure~\ref{fig:sens_station_metrics}(a)), the proposed approach consistently yields shorter average charging waiting time. This advantage becomes particularly pronounced in sparse settings, where poor spatial coordination in the baseline leads to severe queuing and congestion effects. The learned policy proactively balances charging demand across regions and time. Second, in terms of service effectiveness (Figure~\ref{fig:sens_station_metrics}(b)), the proposed framework maintains an order fulfillment rate exceeding 92\% across all configurations, with steady improvements as the number of stations increases. In contrast, the baseline remains below 75\%, indicating limited ability to exploit additional spatial resources. Finally, Figure~\ref{fig:sens_station_metrics}(c) examines the operational overhead ratio. The proposed framework sustains a low and stable ratio around 0.25 regardless of station density. The baseline exhibits substantially higher values, often exceeding 0.45, which reflects inefficient empty driving and poorly timed charging decisions. These results confirm that the proposed method adapts its dispatching logic to changing spatial layouts and preserves economic efficiency even as infrastructure topology varies.

Overall, this sensitivity analysis demonstrates that the proposed framework exhibits strong robustness with respect to changes in charging station density. Performance improvements are not driven solely by increased infrastructure availability, but rather by the ability of the policy to interpret and exploit spatial structure effectively. By learning representations that encode regional interactions and adapting decisions based on inferred task characteristics, the framework maintains high service quality and stable economic performance across a wide range of spatial configurations. These results suggest that the proposed approach can accommodate gradual infrastructure expansion as well as temporary disruptions, making it well suited for deployment in evolving urban charging networks.

\subsection{Sensitivity Analysis on Order Demand Intensity}
\label{subsec_sensitivity_demand}

We next examine the robustness of the proposed framework under varying levels of passenger demand. Starting from the basic dataset containing approximately 15,000 orders, we scale the demand volume to construct three representative operating regimes: a low demand setting (0.8 times the basic), a balanced setting (basic), and a high demand setting (1.2 times the basic). This experiment evaluates whether the learned policy can maintain service quality and operational efficiency as market pressure intensifies.

\begin{figure}[!htbp]
    \centering
    \includegraphics[width=0.75\textwidth]{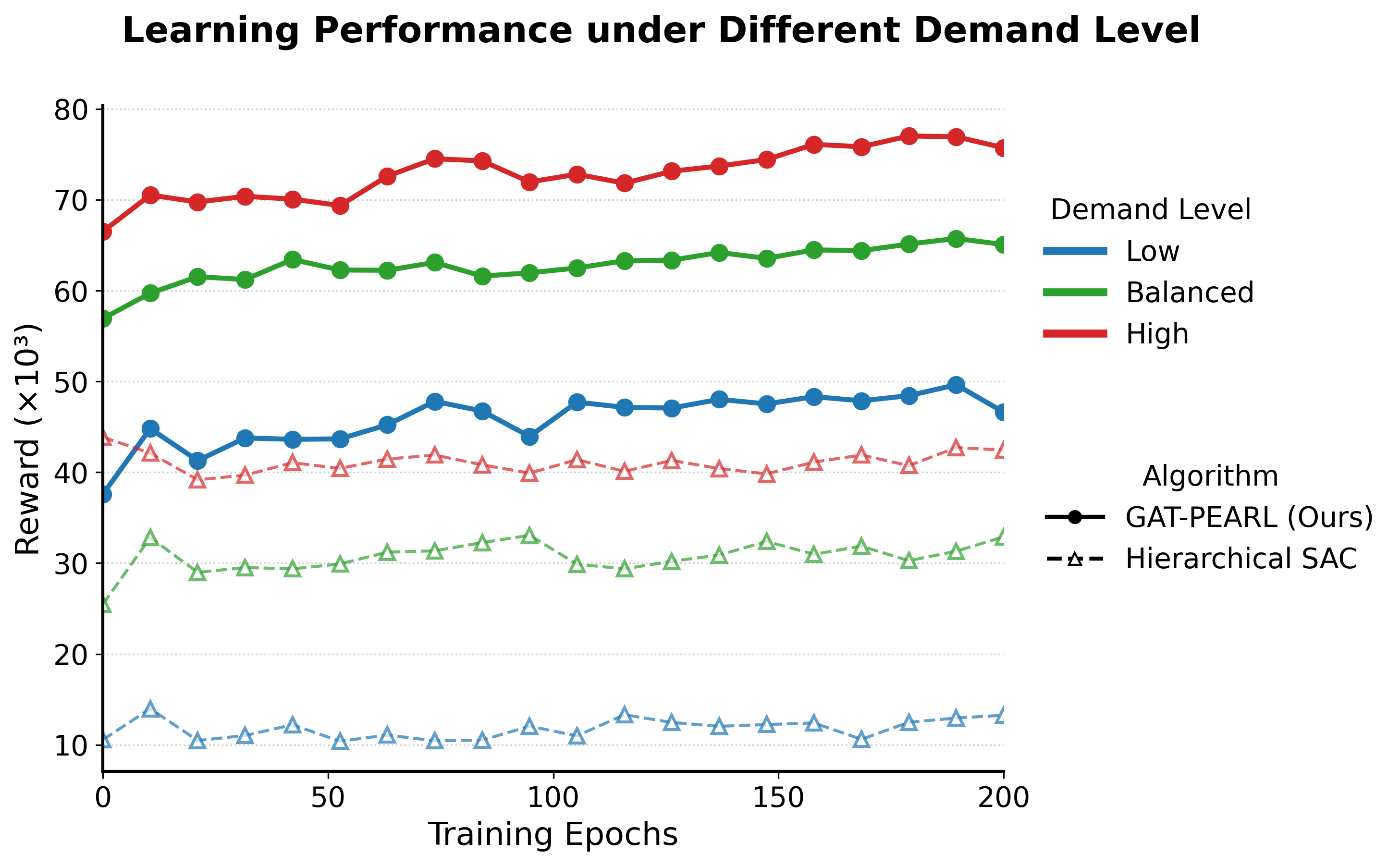}
    \caption{Learning performance comparison under different demand levels.}
    \label{fig:sens_demand_reward}
\end{figure}

Figure \ref{fig:sens_demand_reward} reports the learning trajectories during the first 200 epochs. Similarly, the solid lines represent the proposed Meta-RL framework, while the dashed lines denote the Hierarchical SAC baseline. Generally, higher demand levels create more revenue opportunities and lead to higher attainable rewards. However, the difference between the two methods is substantial. The proposed Meta-RL framework (solid lines) consistently achieves faster performance improvement and higher returns across all demand regimes. In contrast, the baseline method shows slow progress and unstable learning behavior, particularly when demand is high. Even under severe system congestion, the proposed approach converges quickly to an effective policy, demonstrating strong resilience to demand amplification.

\begin{figure}[!htbp]
    \centering
    \includegraphics[width=0.9\textwidth]{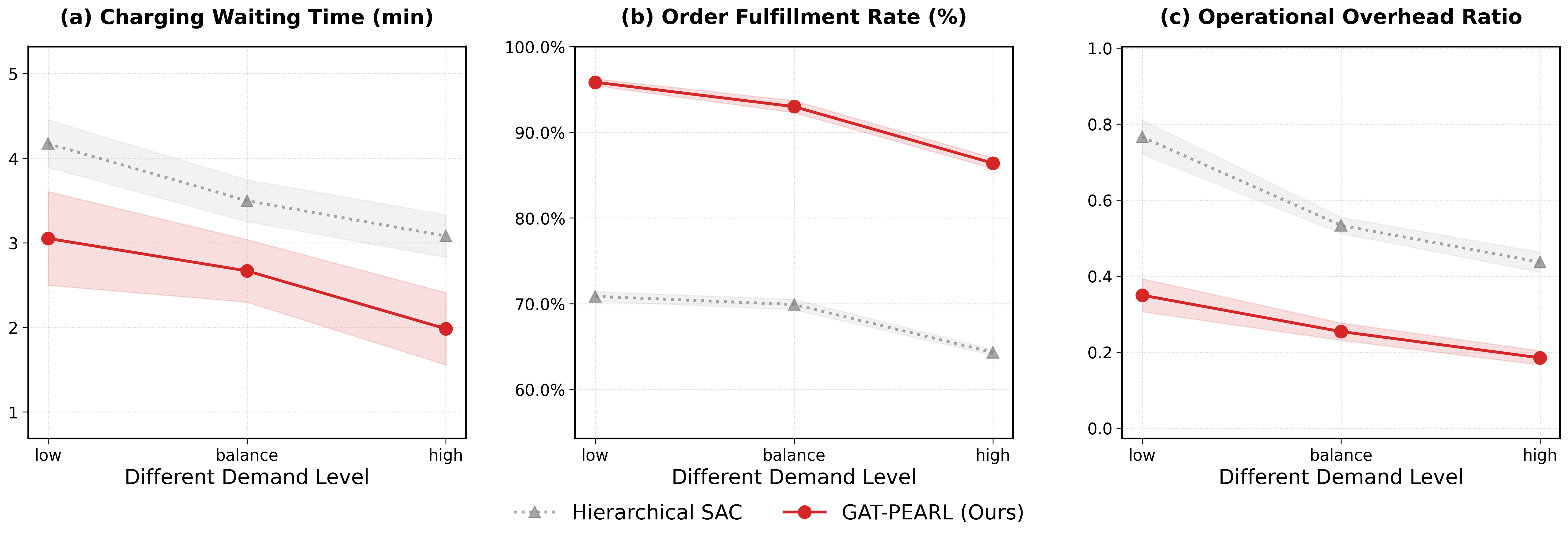}
    \caption{Impact of demand intensity on key operational metrics: (a) charging waiting time, (b) order fulfillment rate, and (c) operational overhead ratio. Shaded areas indicate the confidence interval.}
    \label{fig:sens_demand_metrics}
\end{figure}

To further interpret these results, Figure~\ref{fig:sens_demand_metrics} presents three operational indicators. A particularly informative pattern appears in Figure~\ref{fig:sens_demand_metrics}(a), which reports charging waiting time. While one might expect charging congestion to increase as demand rises, the opposite trend emerges for the proposed approach. Under high demand, charging waiting time decrease rather than increase. This behavior reflects deliberate strategic adjustment rather than an artifact of the simulation. When demand is high, the opportunity cost of charging rises significantly. The Meta RL agent responds by postponing non essential charging actions and prioritizing immediate order fulfillment, only sending vehicles to charge when energy constraints make it unavoidable. This adaptive tradeoff between energy replenishment and revenue generation highlights the ability of the proposed framework to reason about long term system value under varying demand pressure. Then we consider service effectiveness, shown in Figure~\ref{fig:sens_demand_metrics}(b). As demand increases, fleet saturation naturally reduces the fraction of orders that can be served. Nevertheless, the proposed method maintains a high fulfillment rate across all settings. Even in the high demand scenario, it serves more than 86\% of requests, whereas the baseline method deteriorates sharply and drops to approximately 64\%. This gap indicates that the proposed policy allocates vehicles more effectively when competition for service becomes intense. Finally we examine economic efficiency using the operational overhead ratio in Figure~\ref{fig:sens_demand_metrics}(c). Across all demand levels, the proposed framework consistently achieves a lower ratio than the baseline. This result confirms that the learned policy avoids unnecessary empty driving and excessive charging, thereby preserving profitability even under heavy load.

Overall, this sensitivity analysis demonstrates that the proposed framework remains effective across a wide range of demand intensities. By adjusting charging and dispatching behavior in response to changing market conditions, the learned policy delivers stable service quality and strong economic performance even when the operational environment becomes highly stressed.

\section{Conclusion} \label{sec_con}

This study examines a fundamental challenge in the operation of Autonomous Electric Taxi (AET) fleets, namely the fragility of dispatching and charging strategies that rely on a fixed charging infrastructure. In real-world urban environments, charging availability evolves continuously as the charging network expands. Policies optimized under a single infrastructure layout may therefore lose effectiveness when the network changes. 

To address this issue, we propose GAT-PEARL, a hierarchical meta-reinforcement learning framework that enables rapid and reliable adaptation to evolving charging layouts while preserving high operational performance. Meta-learning is suitable for this setting because it explicitly targets fast adaptation under distribution shifts, enabling the controller to achieve strong performance in previously unseen environments using only a small amount of new experience, rather than relying on repeated retraining. Specifically, we introduce two complementary components. First, a graph attention network provides robust spatial representations by representing inter regional interactions through connectivity driven attention, which captures the complex spatial heterogeneity and underlying topological structures of charging station layouts. Second, probabilistic context inference equips the policy with an explicit mechanism to identify the latent layout of the charging network from recent operational trajectories, allowing immediate policy modulation at inference time. Together, these components yield an integrated architecture that adapts efficiently to new charging layouts while maintaining stable dispatching, relocation, and charging performance.

From an operational perspective, our numerical experiments provide concrete verification that GAT-PEARL offers a practical pathway to resilient fleet management under evolving charging infrastructure. The proposed method consistently delivers faster adaptation, higher converged rewards, and markedly lower performance variance across unseen charging network layouts. The sensitivity analyses further yield actionable insights. First, performance is not determined solely by infrastructure quantity. Even under severely constrained pile capacity, the proposed policy achieves strong returns and systematically reduces charging congestion. Second, increasing station density improves performance only when the controller can exploit spatial structure. With graph-based coordination, the fleet maintains high service levels across sparse to dense layouts, whereas the baseline fails to translate added capacity into meaningful gains. Third, under demand surges, the policy adjusts its charging–service tradeoff by deferring nonessential charging and prioritizing revenue-generating trips, which preserves fulfillment and profitability under congestion pressure. Collectively, these results offer clear managerial guidance. Charging expansion should be paired with adaptive coordination algorithms. Operational intelligence can offset temporary capacity shortages, uneven rollouts, and station disruptions. It reduces reliance on manual rule reconfiguration and offline retraining. It also helps operators preserve service quality and economic efficiency during infrastructure upgrades and real-world contingencies.

Several directions remain open for future research. First, beyond adapting to a given charging layout, meta-reinforcement learning could be extended to the planning layer, where the objective is to design or stage charging network expansions that are robust to uncertain demand growth and operational constraints. This would connect infrastructure investment decisions with downstream fleet control, enabling a closed-loop from planning to operation adaptation. Second, the proposed adaptation mechanism can be broadened to address a wider range of real-world non-stationarities beyond charging layout changes, including stochastic queue formation at charging stations, and demand surges triggered by large public events, etc. Incorporating these factors through richer simulation modules or hybrid data-driven surrogates would further enhance realism and provide a more rigorous evaluation of meta-adaptive fleet control in complex urban environments.

\appendix
\section{Nomenclature} \label{sec_appen}
\setcounter{figure}{0}
\renewcommand\thefigure{A-\arabic{figure}}
\setcounter{equation}{0}
\renewcommand{\theequation}{A-\arabic{equation}}
\setcounter{table}{0}
\renewcommand{\thetable}{A-\arabic{table}}
\par

\section*{Acknowledgements}

The acknowledgements section will be completed after the peer-review process.


\section*{Declaration of generative AI and AI-assisted technologies in the writing process}

During the preparation of this work the authors used ChatGPT 5.2 in order to improve language and help write \LaTeX. After using this tool, the authors reviewed and edited the content as needed and take full responsibility for the content of the publication.

\printcredits

\bibliographystyle{cas-model2-names}

\bibliography{cas-refs}

\clearpage  

\begin{table*}[!t]
\centering
\caption{System Model and Environment}
\label{tab:system}
\small
\renewcommand{\arraystretch}{1.1}
\begin{tabular}{p{3cm} p{12.5cm}}
\hline
\textbf{Symbol} & \textbf{Description} \\
\hline

$\mathcal{G}=(\mathcal{J},\mathcal{E})$ 
& Transportation network represented as a graph of regions and adjacencies. \\
$\mathcal{J}$, $\mathcal{E}$ 
& Set of regions and adjacency edges. \\
$\mathcal{K}$ 
& Set of homogeneous AET fleet. \\
$\mathcal{R}$ 
& Set of passenger requests. \\
$\mathcal{C}$ 
& Set of admissible charging infrastructure layouts. \\

$\tau_{j,\bar{j}}$ 
&Travel time between region $j$ and $\bar{j}$. \\
$\varepsilon_{j,\bar{j}}$ 
&Energy consumption between region $j$ and $\bar{j}$. \\
$[b_r,a_r]$ 
& Pickup time window of request $r$. \\
$t_r^{\max}$ 
& Maximum additional waiting time tolerated after the pickup window of request $r$. \\
$\varrho_r$ 
& Revenue for serving request $r$. \\
$\kappa_r$ 
& abandonment penalty of request $r$ . \\

$\xi_k$ 
& State of charge (SOC) of vehicle $k$. \\
$\xi^0$ 
& Minimum safety SOC buffer. \\

$t \in \{1,\ldots,T\}$ 
& Strategic periods. \\
$\eta \in \{1,\ldots,D\}$ 
& Operational intervals within a period. \\

$\mathbf{S}_t$ 
& Global system state at period $t$. \\
$\mathbf{A}_t$ 
& Composite system action at period $t$. \\
$\mathbf{x}_t$ 
& Idle vehicle distribution at period $t$. \\
$g(\mathbf{S}_t, \mathbf{A}_t)$
& Operational cost incurred under system state and actions. \\
$\Theta_t = (\mathbf{O}_t, \mathbf{c})$
& Exogenous stochastic process. \\
$\mathbf{O}_t$
& Time varying demand process. \\
$\mathbf{c}=\Gamma(C)$
& Time-invariant infrastructure context. \\
$\Omega_t(\mathbf{c})$
& Feasible action space. \\
$\mathbf{Q}_t$
& Aggregated origin destination demand matrix. \\

\hline
\end{tabular}
\end{table*}

\begin{table*}[!t]
\centering
\caption{Central and Area Agents}
\label{tab:agents}
\small
\renewcommand{\arraystretch}{1.1}
\begin{tabular}{p{3cm} p{12.5cm}}
\hline
\textbf{Symbol} & \textbf{Description} \\
\hline

$(\mathbf{S}_t^{c}, \mathbf{A}_t^{c}, \mathbf{P}_t^{c}, R_t^{c}, d_t)$
& MDP formulation of Central Agent. \\
$\mathbf{F}_t$ 
& Vehicle availability across regions. \\
$\mathbf{H}_t$ 
& Charging infrastructure operational status. \\
$\mathbf{f}_t \in [-f_{\max}, f_{\max}]^{|\mathcal{J}|}$ 
& Normalized target net vehicle flow ratio. \\
$\mathbf{q}_t \in [q_{\min}, 1]^{|\mathcal{J}|}$ 
& Maximum fraction of charging capacity. \\
$\mathbf{p}_t \in [-p_{\max}, p_{\max}]^{|\mathcal{J}|}$ 
& Dynamic price multipliers. \\
$\bar{\mathbf{A}}_t$
& Aggregate actions actually realized by the system. \\

$(\mathbf{S}_t^{a,j}, \mathbf{A}_t^{a,j}, \mathbf{P}_t^{a}, R_t^{j}, d_t)$
& Local CMDP formulation of Area Agent in region $j$. \\
$\mathbf{h}_t^c$
& Integration of $\mathbf{S}_t^{c}$. \\
$d_{\mathrm{in}}$
& Dimension of $\mathbf{h}_t^c$. \\
$\mathbf{E}$
& Embedding matrix of discrete decision period. \\
$h_e$ 
& Dimension of $\mathbf{E}$. \\
$\mathbf{u}_t^{\cdot,j},\mathbf{v}_t^{\cdot,j},\mathbf{w}_t^{\cdot,j}$
& Flow thresholds for passenger matching, vehicle repositioning, and charging decisions. \\
$c_v^{r}, c_v^{c}$
& Operational cost incurred by repositioning and charging activities. \\

\hline
\end{tabular}
\end{table*}

\begin{table*}[!t]
\centering
\caption{Agent network and algorithms}
\label{tab:learning}
\small
\renewcommand{\arraystretch}{1.1}
\begin{tabular}{p{3cm} p{12.5cm}}
\hline
\textbf{Symbol} & \textbf{Description} \\
\hline

$\mathbf{h}_{t}^{a,i}$
& Initial node feature vector for region $i$. \\
$\mathcal{N}_i$
& Neighborhood set of region $i$. \\
$\mathbf{W}_\nu \in \mathbb{R}^{F_{\mathrm{in}} \times F'}$
& Learnable weight matrix in attention head. \\
$\mathbf{a}_\nu \in \mathbb{R}^{2F'}$
& Learnable attention vector in attention head. \\
$e_{ij}^\nu$
& Unnormalized attention score between regions $i$ and $j$. \\
$\alpha_{ij}^\nu$
& Normalized attention score between regions $i$ and $j$. \\
$\hat{\mathbf{h}}_{t}^{a,i}$
& Output of multi-head attention. \\
$\tilde{\mathbf{h}}_{t}^{a,i}$
& Output of GAT state encoder. \\

$\tilde{\mathbf{z}}^j$
& Latent task representation. \\
$\mathcal{M}^{j}$
& Context set for the $j$-th Area Agent. \\
$\Psi_{\phi_j}$ 
& Local probabilistic belief derived from a single context item. \\
$\mathcal{Q}_{\phi_j}$
& Posterior distribution of $\mathcal{M}^{j}$. \\
$\mathbf{b}_t^{a,j}$
& Area agent policy input representation. \\
$\hat{\mathbf{a}}_t^{a,j}$
& Raw action sample of area agent actor network. \\ 

$\boldsymbol{\Lambda} = (\phi_j, \theta_j^a, \psi_j^a,\bar{\psi}_j^a, \alpha^a)$
& Meta-parameters of area agent $j$. \\
$\theta^c, \psi^c, \alpha^c,\bar{\psi}^{c}$
& Parameters of central agent. \\
$y_t^j, y_t^c$
& Target value calculated by target networks. \\
$\gamma^a,\gamma^c$
& The coefficient preceding the Q-value. \\
$\alpha_j^a,\alpha_j^c$ 
& The temperature parameter. \\
$\bar{\mathcal{H}}^{a},\bar{\mathcal{H}}^{c}$
& Policy entropy. \\
$\mathcal{M}^{j}_{S,i},\tilde{\mathcal{M}}^{j}_{S,i},\mathcal{M}^{j}_{Q,i},\tilde{\mathcal{M}}^{j}_{Q,i}$
& Support/Query set of transition/context batch. \\
$\mathbf{w}_j$
& Actor-critic parameters. \\
$L,L_u$
& Task adaptation steps. \\
$\mathcal{L}_Q,\mathcal{L}_\pi,\mathcal{L}_\alpha,\mathcal{L}_e$
& Loss functions. \\
$\eta_{\pi}^a,\eta_{Q}^a,\eta_{\alpha}^a,\eta_{e}$
& Learning rates. \\
$\beta^{\mathrm{R}},\beta^{\mathrm{KL}}$
& The coefficients in hybrid gradient. \\
$\beta^{c},\beta^{a}$
& Soft update coefficients. \\
$\tilde{U}_{r,k}$
& composite utility score assigning vehicle $k$ to order $r$. \\

\hline
\end{tabular}
\end{table*}

\begin{table*}[!t]
\centering
\caption{List of acronyms used in the paper.}
\label{tab:acronyms}
\small
\renewcommand{\arraystretch}{1.1}
\begin{tabular}{p{3cm} p{12.5cm}}
\hline
\textbf{Acronym} & \textbf{Full name} \\
\hline

EV  &  Electric vehicle \\
AET    & Autonomous Electric Taxi \\
SOC  & State of Charge \\
Meta-RL   & Meta-Reinforcement Learning \\
GAT    & Graph Attention Network \\
ADP    & Approximate Dynamic Programming \\
AMoD    & Autonomous Mobility-on-Demand \\
DRL    & Deep Reinforcement Learning \\
MARL    & Multi-Agent Reinforcement Learning \\
MAML    & Model-Agnostic Meta-Learning \\
PEARL    & Probabilistic Embeddings for Actor-Critic Reinforcement Learning \\
MPMA-MRL    & Multi-Personality Multi-Agent Meta-Reinforcement Learning \\
MDP    & Markov Decision Process \\
CMDP    & Contextual Markov Decision Process \\
SDRAP    & Stochastic Dynamic Resource Allocation Problem \\
SAC    & Soft Actor-Critic \\

\hline
\end{tabular}
\end{table*}

\end{document}